\documentclass[twocolumn]{svjour3}          

\makeatletter
\expandafter\let\csname opt@amsmath.sty\endcsname\relax
\AtBeginDocument{
	\mathindent=15pt 
	\@mathmargin\@centering} 

\renewcommand{\subsection}{%
	\@startsection{subsection}
	{2}
	{\z@}
	{-21dd plus-8pt minus-4pt}
	{10.5dd}
	{\normalsize\bfseries\boldmath}%
}

\renewcommand{\subsubsection}{%
	\@startsection{subsubsection}
	{3}
	{\z@}
	{-13dd plus-8pt minus-4pt}
	{10.5dd}
	{\normalsize\bfseries\upshape}%
}

\makeatother

\smartqed  
\usepackage[usenames]{color}
\usepackage[numbers]{natbib}
\usepackage[table]{xcolor}
\usepackage{makecell}
\usepackage{multirow}
\usepackage{enumitem}
\usepackage{rotating}
\usepackage{array}
\usepackage{times}
\usepackage{graphicx}
\usepackage{psfrag}
\usepackage{subfigure}
\usepackage{rotating}
\usepackage{url}
\usepackage{rotating, graphicx}
\usepackage{booktabs}
\usepackage{lscape}
\usepackage{verbatim}
\usepackage{geometry}
\usepackage[misc]{ifsym}
\papertype{generic article}
\usepackage{amssymb} 
\usepackage{multirow}
\usepackage{longtable}
\usepackage{caption}
\usepackage{xcolor}
\usepackage{mathtools}
\usepackage{bbm}
\usepackage{float}

\usepackage{hyperref}
\hypersetup{colorlinks,linkcolor={red},citecolor={green},urlcolor={blue}}

\newcommand{\tabincell}[2]{\begin{tabular}{@{}#1@{}}#2\end{tabular}}

\begin{document}

\title{3D Object Detection for Autonomous Driving: A Comprehensive Survey}

\subtitle{}

\author{Jiageng Mao $^{1}$  \and
        Shaoshuai Shi $^{3}$ \and
        Xiaogang Wang $^{1,2}$ \and
        Hongsheng Li $^{1,2}$
}

\authorrunning{Jiageng Mao \emph{et al.}} 

\institute{Jiageng Mao (maojiageng@gmail.com) \\
        Shaoshuai Shi (shaoshuaics@gmail.com) \\
        Xiaogang Wang (xgwang@ee.cuhk.edu.hk) \\
        Hongsheng Li (hsli@ee.cuhk.edu.hk) \\
       1 The Chinese University of Hong Kong, China \\
       2 Centre for Perceptual and Interactive Intelligence \\
       3 Max Planck Institute for Informatics, Germany \\
 }

 \date{Received: 7 February 2023 }

\maketitle

\begin{abstract}
Autonomous driving, in recent years, has been receiving increasing attention for its potential to relieve drivers' burdens and improve the safety of driving. In modern autonomous driving pipelines, the perception system is an indispensable component, aiming to accurately estimate the status of surrounding environments and provide reliable observations for prediction and planning. 3D object detection, which aims to predict the locations, sizes, and categories of the 3D objects near an autonomous vehicle, is an important part of a perception system. This paper reviews the advances in 3D object detection for autonomous driving. First, we introduce the background of 3D object detection and discuss the challenges in this task. Second, we conduct a comprehensive survey of the progress in 3D object detection from the aspects of models and sensory inputs, including LiDAR-based, camera-based, and multi-modal detection approaches. We also provide an in-depth analysis of the potentials and challenges in each category of methods. Additionally, we systematically investigate the applications of 3D object detection in driving systems. Finally, we conduct a performance analysis of the 3D object detection approaches, and we further summarize the research trends over the years and prospect the future directions of this area.

\keywords{3D object detection \and perception \and autonomous driving \and deep learning \and computer vision \and robotics}

\end{abstract}

\section{Introduction}\label{sec:introduction}

%
%
%
%
Autonomous driving, which aims to enable vehicles to perceive the surrounding environments intelligently and move safely with little or no human effort, has attained rapid progress in recent years. Autonomous driving techniques have been broadly applied in many scenarios, including self-driving trucks, robotaxis, delivery robots, \textit{etc.}, and are capable of reducing human error and enhancing road safety. As a core component of autonomous driving systems, automotive perception helps autonomous vehicles understand the surrounding environments with sensory input. Perception systems generally take multi-modality data (images from cameras, point clouds from LiDAR scanners, high-definition maps \textit{etc.}) as input, and predict the geometric and semantic information of critical elements on a road. High-quality perception results serve as reliable observations for the following steps such as object tracking, trajectory prediction, and path planning. 

To obtain a comprehensive understanding of driving environments, many vision tasks can be involved in a perception system, \textit{e.g.} object detection and tracking, lane detection, and semantic and instance segmentation. Among these perception tasks, 3D object detection is one of the most indispensable tasks in an automotive perception system. 3D object detection aims to predict the locations, sizes, and classes of critical objects, \textit{e.g.} cars, pedestrians, cyclists, in the 3D space. In contrast to 2D object detection which only generates 2D bounding boxes on images and ignores the actual distance information of objects from the ego-vehicle, 3D object detection focuses on the localization and recognition of objects in the real-world 3D coordinate system. The geometric information predicted by 3D object detection in real-world coordinates can be directly utilized to measure the distances between the ego-vehicle and critical objects, and to further help plan driving routes and avoid collisions.


3D object detection methods have evolved rapidly with the advances of deep learning techniques in computer vision and robotics. These methods have been trying to address the 3D object detection problem from a particular aspect, \textit{e.g.} detection from a particular sensory type or data representation, and lack a systematic comparison with the methods of other categories. Hence a comprehensive analysis of the strengths and weaknesses of all types of 3D object detection methods is desirable and can provide some valuable findings to the research community.

\begin{figure*}[t]
	\centering	\includegraphics[width=\textwidth]{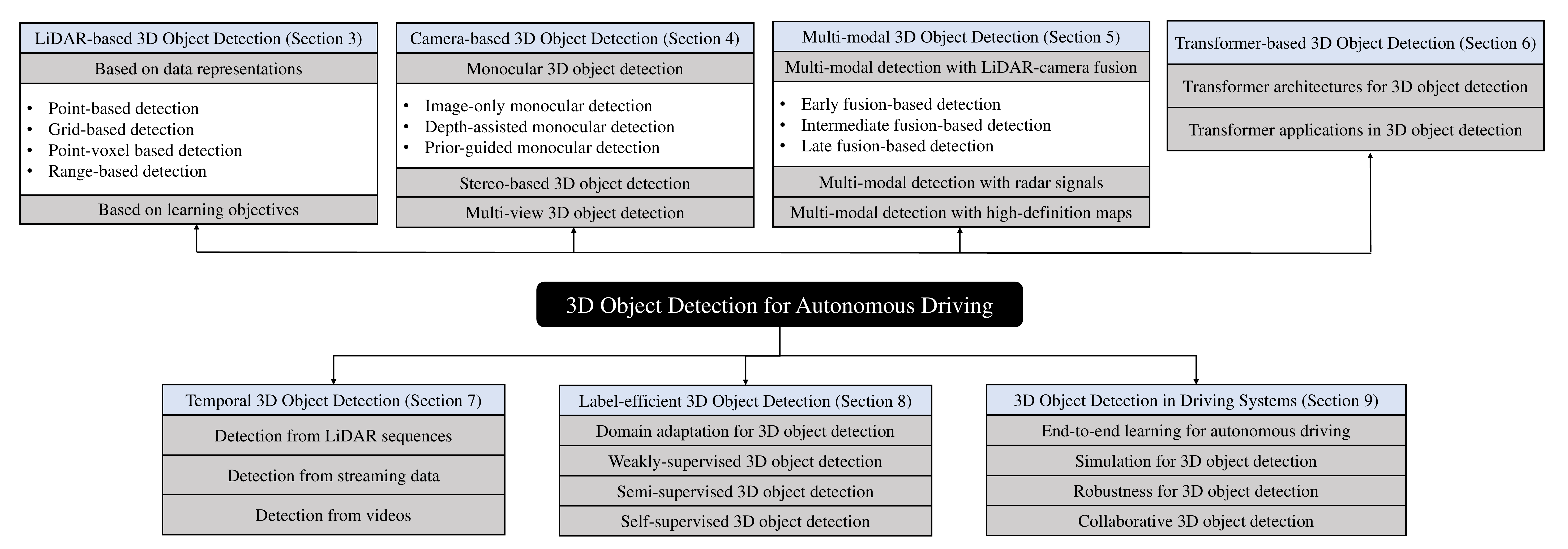}
	\caption{Hierarchically-structured taxonomy of 3D object detection for autonomous driving.}
	\label{fig0-taxonomy}
	\vspace{-2mm}
\end{figure*}

To this end, we propose to comprehensively review the 3D object detection methods for autonomous driving applications and provide in-depth analysis and a systematic comparison on different categories of approaches. Compared to the existing surveys~\cite{surveytits19, survey3d21, survey3darxiv21}, our paper broadly covers the recent advances in this area, \textit{e.g.} 3D object detection from range images, self-/semi-/weakly-supervised 3D object detection, 3D detection in end-to-end driving systems. In contrast to the previous surveys that only focus on detection from point cloud~\cite{surveypointpami19,surveypoint21, surveylidar21}, from monocular images~\cite{surveymono20, survey-mono}, and from multi-modal inputs~\cite{surveymultimodal21}, our paper systematically investigate the 3D object detection methods from all sensory types and in most application scenarios. The major contributions of this work can be summarized as follows:
\begin{itemize}
	\item We provide a comprehensive review of the 3D object detection methods from different perspectives, including detection from different sensory inputs (LiDAR-based, camera-based, and multi-modal detection), detection from temporal sequences, label-efficient detection, as well as the applications of 3D object detection in driving systems. 
	
	\item We summarize 3D object detection approaches structurally and hierarchically, conduct a systematic analysis of these methods, and provide valuable insights for the potentials and challenges of different categories of methods.
	
	\item We conduct a comprehensive performance and speed analysis on the 3D object detection approaches, identify the research trends over years, and provide insightful views on the future directions of 3D object detection.  
	
\end{itemize}

The structure of this paper is organized as follows. First, we introduce the problem definition, datasets, and evaluation metrics of 3D object detection in Section~\ref{sec:background}. Then, we review and analyze the 3D object detection methods based on LiDAR sensors (Section~\ref{sec:lidar}), cameras (Section~\ref{sec:camera}), multi-sensor fusion (Section~\ref{sec:multisensor}), and Transformer-based architectures (Section~\ref{sec:transfomer}). Next, we introduce the detection methods that leverage temporal data in Section~\ref{sec:temporal} and utilize fewer labels in Section~\ref{sec:xlreaning}. We subsequently discuss some critical problems of 3D object detection in driving systems in Section~\ref{sec:system}. Finally, we conduct a speed and performance analysis, investigate the research trends, and prospect the future directions of 3D object detection in Section~\ref{sec:outlook}. A hierarchically-structured taxonomy is shown in Figure~\ref{fig0-taxonomy}. We also provide a constantly updated project page~\href{https://github.com/PointsCoder/Awesome-3D-Object-Detection-for-Autonomous-Driving}{here}.

\begin{figure*}[t]
	\centering
	\includegraphics[width=0.9\textwidth]{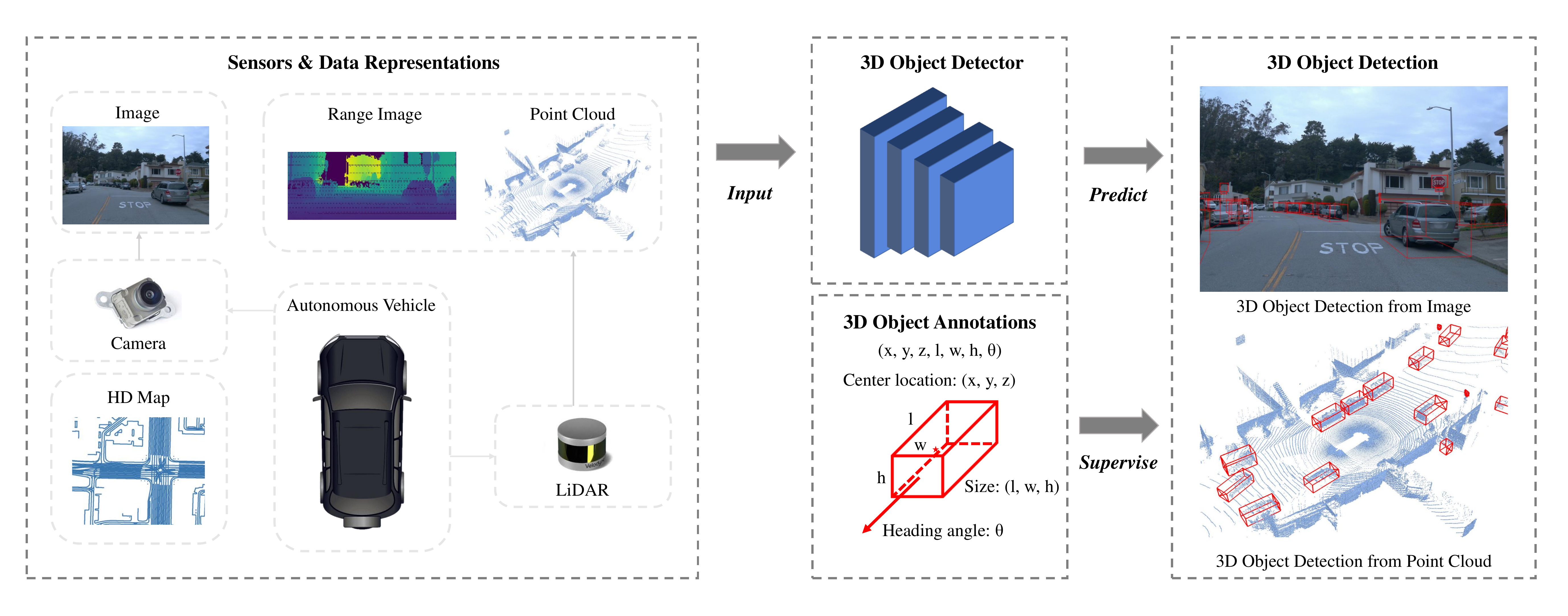}
	\caption{An illustration of 3D object detection in autonomous driving scenarios.}
	\label{fig1-what}
	\vspace{-2mm}
\end{figure*}

\section{Background} \label{sec:background}
\subsection{What is 3D object detection?} \label{sec:bg_task}
\textbf{Problem definition.} 3D object detection aims to predict bounding boxes of 3D objects in driving scenarios from sensory inputs. A general formula of 3D object detection can be represented as
\begin{equation} \label{eq:detection3d}
	\mathcal{B} = f_{det}(\mathcal{I}_{sensor}),
\end{equation}
where $\mathcal{B} = \{B_{1}, \cdots, B_{N}\}$ is a set of $N$ 3D objects in a scene, $f_{det}$ is a 3D object detection model, and $\mathcal{I}_{sensor}$ is one or more sensory inputs. How to represent a 3D object $B_{i}$ is a crucial problem in this task, since it determines what 3D information should be provided for the following prediction and planning steps. In most cases, a 3D object is represented as a 3D cuboid that includes this object, that is
\begin{equation} \label{eq:box}
	B = [x_{c}, y_{c}, z_{c}, l, w, h, \theta, class],
\end{equation}
where $(x_{c}, y_{c}, z_{c})$ is the 3D center coordinate of a cuboid, $l$, $w$, $h$ is the length, width, and height of a cuboid respectively, $\theta$ is the heading angle, \textit{i.e.} the yaw angle, of a cuboid on the ground plane, and $class$ denotes the category of a 3D object, \textit{e.g.} cars, trucks, pedestrians, cyclists. In~\cite{nuscenes20}, additional parameters $v_{x}$ and $v_{y}$ that describe the speed of a 3D object along x and y axes on the ground are employed.

\noindent\textbf{Sensory inputs.} There are many types of sensors that can provide raw data for 3D object detection. Among the sensors, radars, cameras, and LiDAR (Light Detection And Ranging) sensors are the three most widely adopted sensory types. Radars have long detection range and are robust to different weather conditions. Due to the Doppler effect, radars could provide additional velocity measurements. Cameras are cheap and easily accessible, and can be crucial for understanding semantics, \textit{e.g.} the type of traffic sign. Cameras produce images $\mathcal{I}_{cam} \in R^{W \times H \times 3}$ for 3D object detection, where $W$, $H$ are the width and height of an image, and each pixel has $3$ RGB channels. Albeit cheap, cameras have intrinsic limitations to be utilized for 3D object detection. First, cameras only capture appearance information, and are not capable of directly obtaining 3D structural information about a scene. On the other hand, 3D object detection normally requires accurate localization in the 3D space, while the 3D information, \textit{e.g.} depth, estimated from images normally has large errors. In addition, detection from images is generally vulnerable to extreme weather and time conditions. Detecting objects from images at night or on foggy days is much harder than detection on sunny days, which leads to the challenge of attaining sufficient robustness for autonomous driving. 

As an alternative solution, LiDAR sensors can obtain fine-grained 3D structures of a scene by emitting laser beams and then measuring their reflective information. A LiDAR sensor that emits $m$ beams and conducts measurements for $n$ times in one scan cycle can produce a range image $\mathcal{I}_{range} \in R^{m \times n \times 3}$, where each pixel of a range image contains range $r$, azimuth $\alpha$, and inclination $\phi$ in the spherical coordinate system as well as the reflective intensity. Range images are the raw data format obtained by LiDAR sensors, and can be further converted into point clouds by transforming spherical coordinates into Cartesian coordinates. A point cloud can be represented as $\mathcal{I}_{point} \in R^{N \times 3}$, where $N$ denotes the number of points in a scene, and each point has $3$ channels of xyz coordinates. Both range images and point clouds contain accurate 3D information directly acquired by LiDAR sensors. Hence in contrast to cameras, LiDAR sensors are more suitable for detecting objects in the 3D space, and LiDAR sensors are also less vulnerable to time and weather changes. However, LiDAR sensors are much more expensive than cameras, which may limit the applications in driving scenarios. An illustration of 3D object detection is shown in Figure~\ref{fig1-what}.

\begin{table*}[t!]
	\caption{Datasets for 3D object detection in driving scenarios.}
	\centering
	\begin{tabular}{|c|c|c|c|c|c|c|c|c|c|c|}
		\hline
		Dataset & Year & \tabincell{c}{Size \\ (hr.)} & \tabincell{c}{Real-\\world} & \tabincell{c}{LiDAR \\ scans} & Images & \tabincell{c}{3D \\ annotations} & Classes & night/rain & Locations & Other data  \\
		\hline
		\hline
		KITTI~\cite{kitti12conf, kitti13journal} & 2012 & 1.5 & Yes & 15k & 15k & 200k & 8 & No/No & Germany & - \\
		\hline
		KAIST~\cite{kaist18} & 2018 & - & Yes & 8.9k & 8.9k & Yes & 3 & Yes/No & Korea &  thermal images \\
		\hline
		ApolloScape~\cite{apolloscape19, trafficpredict19} & 2019 & 100 & Yes & 20k & 144k & 475k & 6 & -/- & China & - \\
		\hline
		H3D~\cite{h3d19} & 2019 & 0.77 & Yes & 27k & 83k & 1.1M & 8 & No/No & USA & - \\ 
		\hline
		Lyft L5~\cite{lyft19} & 2019 & 2.5 & Yes & 46k & 323k & 1.3M & 9 & No/No & USA & maps \\
		\hline
		Argoverse~\cite{argoversev119} & 2019 & 0.6 & Yes & 44k & 490k & 993k & 15 & Yes/Yes & USA & maps \\
		\hline
            WoodScape~\cite{woodscape} & 2019 & - & Yes & 10k & 10k & - & 3 & Yes/Yes & - & fish-eye camera \\ 
            \hline
		AIODrive~\cite{aiodrive} & 2020 & 6.9 & No & 250k & 250k & 26M & - & Yes/Yes & - & long-range data \\ 
		\hline
		A*3D~\cite{astar3d20} & 2020 & 55 & Yes & 39k & 39k & 230k & 7 & Yes/Yes & SG & - \\
		\hline
		A2D2~\cite{a2d220} & 2020 & - & Yes & 12.5k & 41.3k & - & 14 & -/- & Germany & - \\
		\hline
		Cityscapes 3D~\cite{cityscapes3d20} & 2020 & - & Yes & 0 & 5k & - & 8 & No/No & Germany & - \\
		\hline
		nuScenes~\cite{nuscenes20} & 2020 & 5.5 & Yes & 400k & 1.4M & 1.4M & 23 & Yes/Yes & SG, USA & maps, radar data \\
		\hline
		Waymo Open~\cite{waymo20} & 2020 & 6.4 & Yes & 230k & 1M & 12M & 4 & Yes/Yes & USA & maps \\
		\hline
		Cirrus~\cite{cirrus21} & 2021 & - & Yes & 6.2k & 6.2k & - & 8 & -/- & USA & long-range data \\
		\hline
		PandaSet~\cite{pandaset21} & 2021 & 0.22 & Yes & 8.2k & 49k & 1.3M & 28 & Yes/Yes & USA & - \\ 
		\hline
		KITTI-360~\cite{kitti36021} & 2021 & - & Yes & 80k & 300k & 68k & 37 & -/- & Germany & - \\
		\hline
		Argoverse v2~\cite{argoversev221} & 2021 & - & Yes & - & - & - & 30 & Yes/Yes & USA & maps\\
		\hline
		ONCE~\cite{once21} & 2021 & 144 & Yes & 1M & 7M & 417K & 5 & Yes/Yes & China & - \\
		\hline
	\end{tabular}
	\label{tab:datasets}
\end{table*}

\noindent\textbf{Analysis: comparisons with 2D object detection.} 2D object detection, which aims to generate 2D bounding boxes on images, is a fundamental problem in computer vision. 3D object detection methods have borrowed many design paradigms from the 2D counterparts: proposals generation and refinement, anchors, non maximum suppression, \textit{etc.} However, from many aspects, 3D object detection is not a naive adaptation of 2D object detection methods to the 3D space. (1) 3D object detection methods have to deal with heterogeneous data representations. Detection from point clouds requires novel operators and networks to handle irregular point data, and detection from both point clouds and images needs special fusion mechanisms. (2) 3D object detection methods normally leverage distinct projected views to generate object predictions. As opposed to 2D object detection methods that detect objects from the perspective view, 3D methods have to consider different views to detect 3D objects, \textit{e.g.} from the bird's-eye view, point view, and cylindrical view. (3) 3D object detection has a high demand for accurate localization of objects in the 3D space. A decimeter-level localization error can lead to a detection failure of small objects such as pedestrians and cyclists, while in 2D object detection, a localization error of several pixels may still maintain a high Intersection over Union (IoU) between predicted and ground truth bounding boxes. Hence accurate 3D geometric information is indispensable for 3D object detection from either point clouds or images.

\noindent\textbf{Analysis: comparisons with indoor 3D object detection.} There is also a branch of works~\cite{fpointnet18, votenet19, imvotenet20, groupfree21} on 3D object detection in indoor scenarios. Indoor datasets, e.g. ScanNet~\cite{scannet17}, SUN RGB-D~\cite{sunrgbd15}, provide 3D structures of rooms reconstructed from RGB-D sensors and 3D annotations including doors, windows, beds, chairs, \textit{etc.} 3D object detection in indoor scenes is also based on point clouds or images. However, compared to indoor 3D object detection, there are unique challenges of detection in driving scenarios. (1) Point cloud distributions from LiDAR and RGB-D sensors are different. In indoor scenes, points are relatively uniformly distributed on the scanned surfaces and most 3D objects receive a sufficient number of points on their surfaces. However, in driving scenes most points fall in a near neighborhood of the LiDAR sensor, and those 3D objects that are far away from the sensor will receive only a few points. Thus methods in driving scenarios are specially required to handle various point cloud densities of 3D objects and accurately detect those faraway and sparse objects. (2) Detection in driving scenarios has a special demand for inference latency. Perception in driving scenes has to be real-time to avoid accidents. Hence those methods are required to be computationally efficient, otherwise they will not be applied in real-world applications. 

\subsection{Datasets}
A large number of driving datasets have been built to provide multi-modal sensory data and 3D annotations for 3D object detection. Table~\ref{tab:datasets} lists the datasets that collect data in driving scenarios and provide 3D cuboid annotations. KITTI~\cite{kitti12conf} is a pioneering work that proposes a standard data collection and annotation paradigm: equipping a vehicle with cameras and LiDAR sensors, driving the vehicle on roads for data collection, and annotating 3D objects from the collected data. The following works made improvements mainly from the $4$ aspects. (1) Increasing the scale of data. Compared to~\cite{kitti12conf}, the recent large-scale datasets~\cite{waymo20, nuscenes20, once21} have more than 10x point clouds, images and annotations. (2) Improving the diversity of data. \cite{kitti12conf} only contains driving data obtained in the daytime and in good weather, while recent datasets~\cite{kaist18, argoversev119, astar3d20, nuscenes20, waymo20, pandaset21, once21, argoversev221} provide data captured at night or in rainy days. (3) Providing more annotated categories. Some datasets~\cite{kitti36021, pandaset21, a2d220, argoversev221, nuscenes20} can provide more fine-grained object classes, including animals, barriers, traffic cones, \textit{etc.} They also provide fine-grained sub-categories of existing classes, \textit{e.g.} the adult and child category of the existing pedestrian class in~\cite{nuscenes20}. (4) Providing data of more modalities. In addition to images and point clouds, recent datasets provide more data types, including high-definition maps~\cite{lyft19, argoversev119, waymo20, argoversev221}, radar data~\cite{nuscenes20}, long-range LiDAR data~\cite{aiodrive, cirrus21}, thermal images~\cite{kaist18}.

\noindent\textbf{Analysis: future prospects of driving datasets.} The research community has witnessed an explosion of datasets for 3D object detection in autonomous driving scenarios. A subsequent question may be asked: what will the next-generation autonomous driving datasets look like? Considering the fact that 3D object detection is not an independent task but a component in driving systems, we propose that future datasets will include all important tasks in autonomous driving: perception, prediction, planning, and mapping, as a whole and in an end-to-end manner, so that the development and evaluation of 3D object detection methods will be considered from an overall and systematic view. There are some datasets~\cite{waymo20, nuscenes20, woodscape} working towards this goal. 

\begin{figure*}[t]
	\centering
	\includegraphics[width=\textwidth]{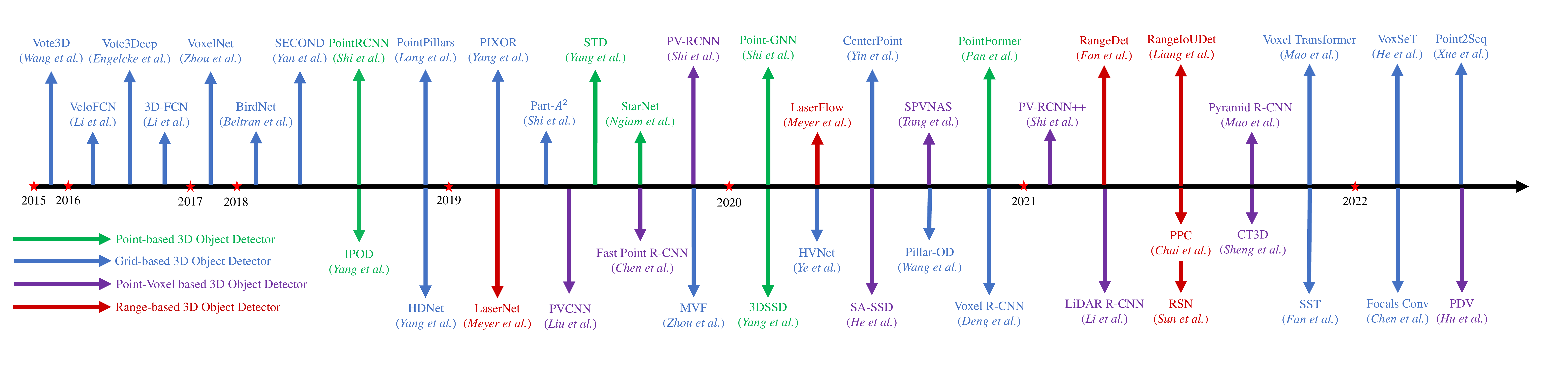}
	\caption{Chronological overview of the LiDAR-based 3D object detection methods.}
	\label{fig2-lidar-roadmap}
\end{figure*}

\subsection{Evaluation metrics}
Various evaluation metrics have been proposed to measure the performance of 3D object detection methods. Those evaluation metrics can be divided into two categories. The first category tries to extend the Average Precision (AP) metric~\cite{coco14} in 2D object detection to the 3D space:
\begin{equation}
	AP = \int_{0}^{1} max\{p(r^{\prime} | r^{\prime} \ge r)\}dr,
\end{equation}
where $p(r)$ is the precision-recall curve same as~\cite{coco14}. The major difference with the 2D AP metric lies in the matching criterion between ground truths and predictions when calculating precision and recall. KITTI~\cite{kitti12conf} proposes two widely-used AP metrics: $AP_{3D}$ and $AP_{BEV}$, where $AP_{3D}$ matches the predicted objects to the respective ground truths if the 3D Intersection over Union (3D IoU) of two cuboids is above a certain threshold, and $AP_{BEV}$ is based on the IoU of two cuboids from the bird's-eye view (BEV IoU). NuScenes~\cite{nuscenes20} proposes $AP_{center}$ where a predicted object is matched to a ground truth object if the distance of their center locations is below a certain threshold, and NuScenes Detection Score (NDS) is further proposed to take both $AP_{center}$ and the error of other parameters, \textit{i.e.} size, heading, velocity, into consideration. Waymo~\cite{waymo20} proposes $AP_{hungarian}$ that applies the Hungarian algorithm to match the ground truths and predictions, and AP weighted by Heading ($APH$) is proposed to incorporate heading errors as a coefficient into the AP calculation.

The other category of approaches tries to resolve the evaluation problem from a more practical perspective. The idea is that the quality of 3D object detection should be relevant to the downstream task, \textit{i.e.} motion planning, so that the best detection methods should be most helpful to planners to ensure the safety of driving in practical applications. Toward this goal, PKL~\cite{pkleval20} measures the detection quality using the KL-divergence of the ego vehicle's future planned states based on the predicted and ground truth detections respectively. SDE~\cite{sdeeval21} leverages the minimal distance from the object boundary to the ego vehicle as the support distance and measures the support distance error.

\noindent\textbf{Analysis: pros and cons of different evaluation metrics.} AP-based evaluation metrics~\cite{kitti12conf, nuscenes20, waymo20} can naturally inherit the advantages from 2D detection. However, those metrics overlook the influence of detection on safety issues, which are also critical in real-world applications. For instance, a misdetection of an object near the ego vehicle and far away from the ego vehicle may receive a similar level of punishment in AP calculation, but a misdetection of nearby objects is substantially more dangerous than a misdetection of faraway objects in practical applications. Thus AP-based metrics may not be the optimal solution from the perspective of safe driving. PKL~\cite{pkleval20} and SDE~\cite{sdeeval21} partly resolve the problem by considering the effects of detection in downstream tasks, but additional challenges will be introduced when modeling those effects. PKL~\cite{pkleval20} requires a pre-trained motion planner for evaluating the detection performance, but a pre-trained planner also has innate errors that could make the evaluation process inaccurate. SDE~\cite{sdeeval21} requires reconstructing object boundaries which is generally complicated and challenging. 

\begin{figure*}[t]
	\centering
	\includegraphics[width=\textwidth]{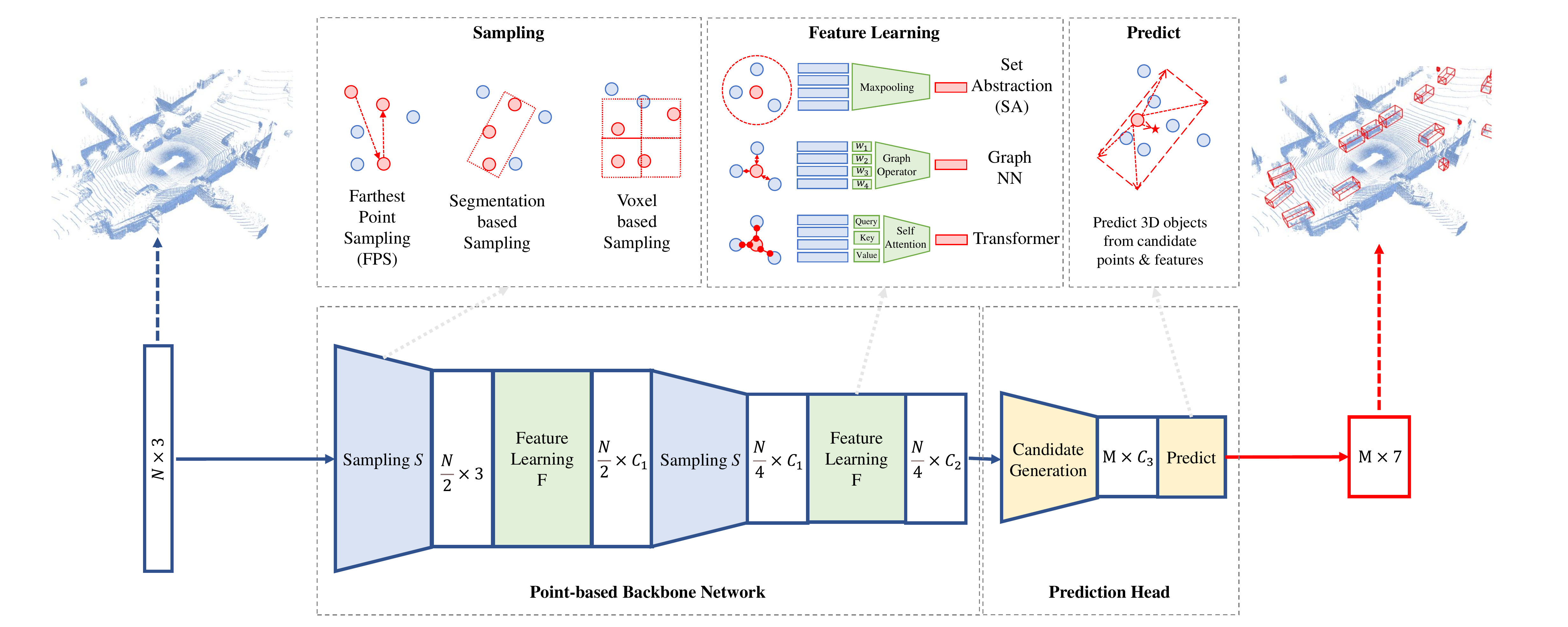}
	\caption{An illustration of point-based 3D object detection methods.} 
	\label{fig3-point-det}
	\vspace{-2mm}
\end{figure*}

\section{LiDAR-based 3D Object Detection} \label{sec:lidar}
In this section, we introduce the 3D object detection methods based on LiDAR data, \textit{i.e.} point clouds or range images. In Section~\ref{sec:lidar_rep}, we review and analyze the LiDAR-based 3D object detection models based on different data representations, including the point-based, grid-based, point-voxel based, and range-based methods. In Section~\ref{sec:lidar_learning}, we investigate the learning objectives for 3D object detectors, including the anchor-based and anchor-free frameworks, as well as the auxiliary tasks adopted in LiDAR-based 3D object detection. A chronological overview of the LiDAR-based 3D detection methods is shown in Figure~\ref{fig2-lidar-roadmap}.      

\subsection{Data representations for 3D object detection} \label{sec:lidar_rep}
\noindent\textbf{Problem and Challenge.} In contrast to images where pixels are regularly distributed on an image plane, point cloud is a sparse and irregular 3D representation that requires specially designed models for feature extraction. Range image is a dense and compact representation, but range pixels contain 3D information instead of RGB values. Hence directly applying conventional convolutional networks on range images may not be an optimal solution. On the other hand, detection in autonomous driving scenarios generally has a requirement for real-time inference. Therefore, how to develop a model that could effectively handle point cloud or range image data while maintaining a high efficiency remains an open challenge to the research community.

\subsubsection{Point-based 3D object detection}

\noindent\textbf{General Framework.} Point-based 3D object detection methods generally inherit the success of deep learning techniques on point cloud~\cite{pointnet17, pointnet++17, dgcnn19, interpconv19} and propose diverse architectures to detect 3D objects directly from raw points. Point clouds are first passed through a point-based backbone network, in which the points are gradually sampled and features are learned by point cloud operators. 3D bounding boxes are then predicted based on the downsampled points and features. A general point-based detection framework is shown in Figure~\ref{fig3-point-det} and a taxonomy of point-based detectors is in Table~\ref{tab:point_sample_feature}. There are two basic components of a point-based 3D object detector: point cloud sampling and feature learning.

\noindent\textbf{Point Cloud Sampling.} Farthest Point Sampling (FPS) in PointNet++~\cite{pointnet++17} has been broadly adopted in point-based detectors, in which the farthest points are sequentially selected from the original point set. PointRCNN~\cite{pointrcnn19} is a pioneering work that adopts FPS to progressively downsample input point cloud and generate 3D proposals from the downsampled points. Similar design paradigm has also been adopted in many following works with improvements like segmentation guided filtering~\cite{ipod}, feature space sampling~\cite{3dssd}, random sampling~\cite{starnet}, voxel-based sampling~\cite{pointgnn20}, and coordinate refinement~\cite{pointformer}.

\begin{table}[tbp]
	\caption{A taxonomy of point-based detection methods based on point cloud sampling and feature learning.}
	\centering
	\begin{tabular}{|c|c|c|c|}
		\hline
		Method & Context $\Omega$ & Sampling $S$ & Feature $F$ \\
		\hline
		\hline
		PointRCNN~\cite{pointrcnn19} & Ball Query & FPS & Set Abstraction \\
		\hline
		IPOD~\cite{ipod} & Ball Query & Seg. & Set Abstraction \\
		\hline
		STD~\cite{std} & Ball Query & FPS & Set Abstraction \\
		\hline
		3DSSD~\cite{3dssd} & Ball Query & Fusion-FPS & Set Abstraction \\
		\hline
		Point-GNN~\cite{pointgnn20} & Ball Query & Voxel & Graph \\
		\hline
		StarNet~\cite{starnet} & Ball Query & Targeted-FPS & Graph \\
		\hline
		Pointformer~\cite{pointformer} & Ball Query & FPS + Refine & Transformer \\
		\hline
	\end{tabular}
	\label{tab:point_sample_feature}
	\vspace{-2mm}
\end{table}

\begin{figure*}[t]
	\centering
	\includegraphics[width=\textwidth]{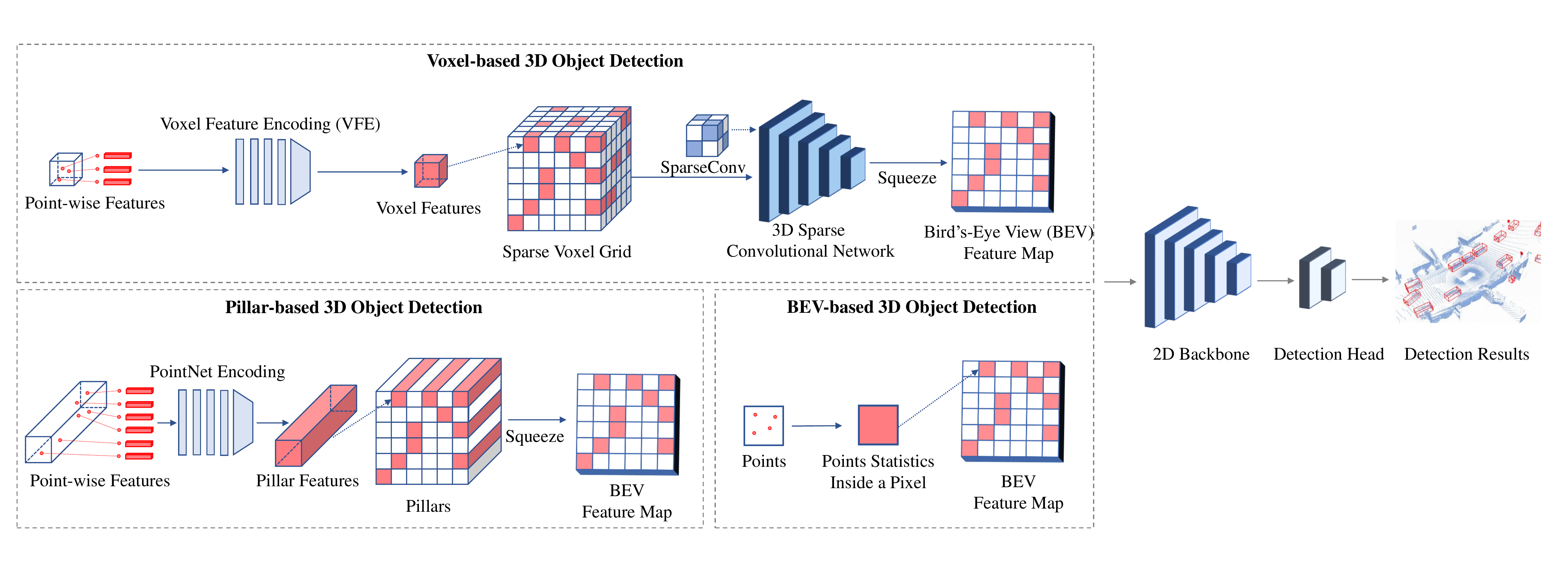}
	\caption{An illustration of grid-based 3D object detection methods.}
	\label{fig4-grid-det}
	\vspace{-2mm}
\end{figure*}

\noindent\textbf{Point Cloud Feature Learning.} A series of works~\cite{pointrcnn19, std, joint3dbaidu, 3dcenternet} leverage set abstraction in \cite{pointnet17} to learn features from point cloud. Specifically, context points are first collected within a pre-defined radius by ball query. Then, the context points and features are aggregated through multi-layer perceptrons and maxpooling to obtain the new features. There are also other works resorting to different point cloud operators, including graph operators~\cite{pointgnn20, pointrgcn19, starnet, relationgraph20, svganet}, attentional operators~\cite{attentionalpointnet}, and Transformer~\cite{pointformer}.

\noindent\textbf{Analysis: potentials and challenges on point cloud feature learning and sampling.} The representation power of point-based detectors is mainly restricted by two factors: the number of context points and the context radius adopted in feature learning. Increasing the number of context points will gain more representation power but at the cost of increasing much memory consumption. Suitable context radius in ball query is also an important factor: the context information may be insufficient if the radius is too small and the fine-grained 3D information may lose if the radius is too large. These two factors have to be determined carefully to balance the efficacy and efficiency of detection models.

Point cloud sampling is a bottleneck in inference time for most point-based methods. Random uniform sampling can be conducted in parallel with high efficiency. However, considering points in LiDAR sweeps are not uniformly distributed, random uniform sampling may tend to over-sample those regions of high point cloud density while under-sample those sparse regions, which normally leads to poor performance compared to farthest point sampling. Farthest point sampling and its variants can attain a more uniform sampling result by sequentially selecting the farthest point from the existing point set. Nevertheless, farthest point sampling is intrinsically a sequential algorithm and can not become highly parallel. Thus farthest point sampling is normally time-consuming and not ready for real-time detection.

\subsubsection{Grid-based 3D object detection}

\begin{table*}[t!]
	\caption{A taxonomy of grid-based detection methods based on models and data representations.}
	\centering
	\begin{tabular}{|c|c|c|c|c|c|c|c|c|c|c|}
		\hline
		\multirow{2}*{Method} & \multicolumn{3}{c|}{Representation} &  \multicolumn{3}{c|}{Encoder} & \multicolumn{4}{c|}{Neural Networks}\\
		\cline{2-11}
         & Voxels & BEV maps & Pillars & Voxelization & Projection & PointNet & 3D CNN & 2D CNN & Head & Transformer \\
		\hline
		Vote3D~\cite{vote3d} & \checkmark & & & \checkmark & & & \checkmark & & &\\
		\hline
		Vote3Deep~\cite{vote3deep} & \checkmark & & & \checkmark & & & \checkmark & & &\\
		\hline
		3D-FCN~\cite{3dfcn} & \checkmark & & & \checkmark & & & \checkmark & & &\\
		\hline
		VeloFCN~\cite{velofcn} & & \checkmark & & & \checkmark & & & \checkmark & &\\
		\hline
		BirdNet~\cite{birdnet} & & \checkmark & & & \checkmark & & & \checkmark & &\\
		\hline
		PIXOR~\cite{pixor} & & \checkmark & & & \checkmark & & & \checkmark & &\\
		\hline
		HDNet~\cite{hdnet} & & \checkmark & & & \checkmark & & & \checkmark & &\\
		\hline
		VoxelNet~\cite{voxelnet18} & \checkmark & \checkmark & & \checkmark & & \checkmark & \checkmark & \checkmark & &\\
		\hline
		SECOND~\cite{second} & \checkmark & \checkmark & & \checkmark & & & \checkmark & \checkmark & &\\
		\hline
		MVF~\cite{mvfwaymo} &  \checkmark & \checkmark & & \checkmark & & & & \checkmark & &\\
		\hline
		PointPillars~\cite{pointpillars} & & \checkmark & \checkmark & & & \checkmark & & \checkmark& & \\
		\hline
		Pillar-OD~\cite{pillar-od} & & \checkmark & \checkmark & & & \checkmark & & \checkmark & &\\
		\hline
		Part-A$^2$ Net~\cite{parta2} & \checkmark & \checkmark & & \checkmark & & & \checkmark & \checkmark & \checkmark &\\
		\hline
		Voxel R-CNN~\cite{voxelrcnn} & \checkmark & \checkmark & & \checkmark & & & \checkmark & \checkmark & \checkmark &\\
		\hline
		CenterPoint~\cite{centerpoint} & \checkmark & \checkmark & & \checkmark & & & \checkmark & \checkmark & &\\
		\hline
		Voxel Transformer~\cite{votr} & \checkmark & \checkmark & & \checkmark & & & & \checkmark & & \checkmark \\
		\hline
		SST~\cite{sst} & \checkmark & \checkmark & & \checkmark & & & & \checkmark & & \checkmark \\
		\hline
		SWFormer~\cite{swformer} & \checkmark & \checkmark & & \checkmark & & & & & & \checkmark \\
		\hline
	\end{tabular}
	\label{tab:grid_data_model}
	\vspace{-2mm}
\end{table*}

\textbf{General Framework.} Grid-based 3D object detectors first rasterize point clouds into discrete grid representations, \textit{i.e.} voxels, pillars, and bird's-eye view (BEV) feature maps. Then they apply conventional 2D convolutional neural networks or 3D sparse neural networks to extract features from the grids. Finally, 3D objects can be detected from the BEV grid cells. An illustration of grid-based 3D object detection is shown in Figure~\ref{fig4-grid-det} and a taxonomy of grid-based detectors is in Table~\ref{tab:grid_data_model}. There are two basic components in grid-based detectors: grid-based representations and grid-based neural networks.
%

\noindent\textbf{Grid-based representations.} There are $3$ major types of grid representations: voxels, pillars, and BEV feature maps.  

\textit{Voxels.} If we rasterize the detection space into a regular 3D grid, voxels are the grid cells. A voxel can be non-empty if point clouds fall into this grid cell. Since point clouds are sparsely distributed, most voxel cells in the 3D space are empty and contain no point. In practical applications, only those non-empty voxels are stored and utilized for feature extraction. VoxelNet~\cite{voxelnet18} is a pioneering work that utilizes sparse voxel grids and proposes a novel voxel feature encoding (VFE) layer to extract features from the points inside a voxel cell. A similar voxel encoding strategy has been adopted by a series of following works~\cite{cbgs, afdet, centerpoint, objectdgcnn, ciassd, psanetvoxel, voxelrcnn, parta2}. In addition, there are two categories of approaches trying to improve the voxel representation for 3D object detection: (1) Multi-view voxels. Some methods propose a dynamic voxelization and fusion scheme from diverse views, \textit{e.g.} from both the bird's-eye view and the perspective view~\cite{mvfwaymo}, from the cylindrical and spherical view~\cite{everyviewcounts}, from the range view~\cite{vista}. (2) Multi-scale voxels. Some papers generate voxels of different scales~\cite{hvnet} or use reconfigurable voxels~\cite{reconfigurablevoxels}.

\textit{Pillars.} Pillars can be viewed as special voxels in which the voxel size is unlimited in the vertical direction. Pillar features can be aggregated from points through a PointNet~\cite{pointnet17} and then scattered back to construct a 2D BEV image for feature extraction. PointPillars~\cite{pointpillars} is a seminal work that introduces the pillar representation and is followed by~\cite{pillar-od, sst}. 

\textit{BEV feature maps.} Bird's-eye view feature map is a dense 2D representation, where each pixel corresponds to a specific region and encodes the points information in this region. BEV feature maps can be obtained from voxels and pillars by projecting the 3D features into the bird's-eye view, or they can be directly obtained from raw point clouds by summarizing points statistics within the pixel region. The commonly-used statistics include binary occupancy~\cite{pixor, hdnet, rad} and the height and density of local point cloud~\cite{mv3d, birdnet, rt3d, yolo3d, complexyolo, zhangyi19, birdnet+, velofcn}.
 
\noindent\textbf{Grid-based neural networks.} There are $2$ major types of grid-based networks: 2D convolutional neural networks for BEV feature maps and pillars, and 3D sparse neural networks for voxels.

\textit{2D convolutional neural networks.} Conventional 2D convolutional neural networks can be applied to the BEV feature map to detect 3D objects from the bird's-eye view. In most works, the 2D network architectures are generally adapted from those successful designs in 2D object detection, \textit{e.g.} ResNet~\cite{resnet} adopted in~\cite{pixor}, Region Proposal Network (RPN)~\cite{rpn} and Feature Pyramid Network (FPN)~\cite{fpn} in~\cite{birdnet, birdnet+, pointpillars, voxel-fpn, complexyolo, psanetvoxel}, and spatial attention in~\cite{li2021anchor, tanet, sarpnet}.

\textit{3D sparse neural networks.} 3D sparse convolutional neural networks are based on two specialized 3D convolutional operators: sparse convolutions and submanifold convolutions~\cite{submanifold}, which can efficiently conduct 3D convolutions only on those non-empty voxels. Compared to~\cite{vote3d, vote3deep, dops, 3dfcn} that perform standard 3D convolutions on the whole voxel space, sparse convolutional operators are highly efficient and can obtain a real-time inference speed. SECOND~\cite{second} is a seminal work that implements these two sparse operators with GPU-based hash tables and builds a sparse convolutional network to extract 3D voxel features. This network architecture has been applied in numerous works~\cite{cbgs, centerpoint, segvoxelnet, objectdgcnn, afdet, hotspotnet, ssn, second, voxelrcnn, ciassd} and becomes the most widely-used backbone network in voxel-based detectors. There is also a series of works trying to improve the sparse operators~\cite{focalsconv3d}, extend~\cite{second} into a two-stage detector~\cite{parta2, voxelrcnn}, and introduce the Transformer~\cite{transformer} architecture into voxel-based detection~\cite{votr, sst}.  

\noindent\textbf{Analysis: pros and cons of different grid representations.} In contrast to the 2D representations like BEV feature maps and pillars, voxels contain more structured 3D information. In addition, deep voxel features can be learned through a 3D sparse network. However, a 3D neural network brings additional time and memory costs. BEV feature map is the most efficient grid representation that directly projects point cloud into a 2D pseudo image without specialized 3D operators like sparse convolutions or pillar encoding. 2D detection techniques can also be seamlessly applied to BEV feature maps without much modification. BEV-based detection methods generally can obtain high efficiency and a real-time inference speed. However, simply summarizing points statistics inside pixel regions loses too much 3D information, which leads to less accurate detection results compared to voxel-based detection. Pillar-based detection approaches leverage PointNet to encode 3D points information inside a pillar cell, and the features are then scattered back into a 2D pseudo image for efficient detection, which balances the effectiveness and efficiency of 3D object detection.

\noindent\textbf{Analysis: challenges of the grid-based detection methods.} A critical problem that all grid-based methods have to face is choosing the proper size of grid cells. Grid representations are essentially discrete formats of point clouds by converting the continuous point coordinates into discrete grid indices. The quantization process inevitably loses some 3D information and its efficacy largely depends on the size of grid cells: smaller grid size yields high resolution grids, and hence maintains more fine-grained details that are crucial to accurate 3D object detection. Nevertheless, reducing the size of grid cells leads to a quadratic increase in memory consumption for the 2D grid representations like BEV feature maps or pillars. As for the 3D grid representation like voxels, the problem can become more severe. Therefore, how to balance the efficacy brought by smaller grid sizes and the efficiency influenced by the memory increase remains an open challenge to all grid-based 3D object detection methods. 

\subsubsection{Point-voxel based 3D object detection}

\begin{figure*}[t]
	\centering
	\includegraphics[width=0.9\textwidth]{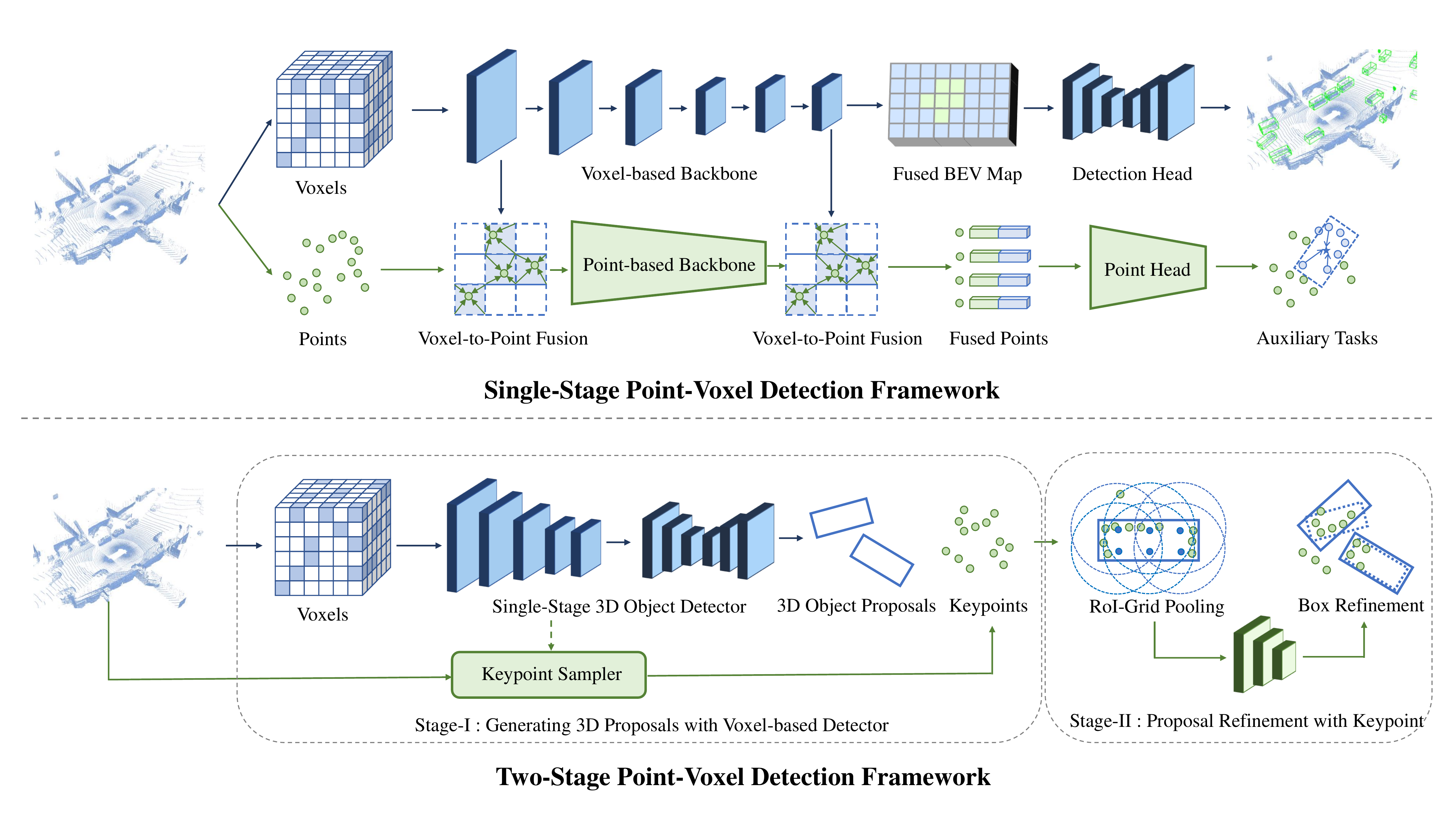}
	\caption{An illustration of point-voxel based 3D object detection methods.} 
	\label{fig5-pv-det}
	\vspace{-2mm}
\end{figure*}

Point-voxel based approaches resort to a hybrid architecture that leverages both points and voxels for 3D object detection. Those methods can be divided into two categories: the single-stage and two-stage detection frameworks. An illustration of the two categories is shown in Figure~\ref{fig5-pv-det} and a taxonomy is in Table~\ref{tab:pointvoxel_tax}. 

\noindent\textbf{Single-stage point-voxel detection frameworks.} Single-stage point-voxel based 3D object detectors try to bridge the features of points and voxels with the point-to-voxel and voxel-to-point transform in the backbone networks. Points contain fine-grained geometric information and voxels are efficient for computation, and combining them together in the feature extraction stage naturally benefits from both two representations. The idea that leverages point-voxel feature fusion in backbones has been explored by many works, with the contributions like point-voxel convolutions~\cite{pointvoxelcnn, spvnas}, auxiliary point-based networks~\cite{sassd, voxeltopoint, hollow3d}, and multi-scale feature fusion~\cite{pvgnet, hvpr, m3detr}.

\noindent\textbf{Two-stage point-voxel detection frameworks.} Two-stage point-voxel based 3D object detectors resort to different data representations for different detection stages. Specifically, at the first stage, they employ a voxel-based detection framework to generate a set of 3D object proposals. In the second stage, keypoints are first sampled from the input point cloud, and then the 3D proposals are further refined from the keypoints through novel point operators. PV-RCNN~\cite{pv-rcnn} is a seminal work that adopts~\cite{second} as the first-stage detector, and the RoI-grid pooling operator is proposed for the second-stage refinement. The following works try to improve the second-stage head with novel modules and operators, \textit{e.g.} RefinerNet~\cite{fastpointrcnn}, VectorPool~\cite{pv-rcnn++}, point-wise attention~\cite{infofocus}, scale-aware pooling~\cite{lidar-rcnn}, RoI-grid attention~\cite{pyramid-rcnn}, channel-wise Transformer~\cite{ct3d}, and point density-aware refinement module~\cite{pdv}. 

\begin{table}[t!]
	\caption{A taxonomy of point-voxel based detection methods.}
	\centering
	\begin{tabular}{|c|c|}
		\hline
		Method & Contribution\\
		\hline
		\hline
		\multicolumn{2}{|c|}{Single-Stage Detection Framework} \\
		\hline
		\hline
		PVCNN~\cite{pointvoxelcnn} & Point-Voxel Convolution\\
		\hline
		SPVNAS~\cite{spvnas} & Sparse Point-Voxel Convolution \\
		\hline
		SA-SSD~\cite{sassd} & Auxiliary Point Network\\
		\hline
		PVGNet~\cite{pvgnet} & Point-Voxel-Grid Fusion \\
		\hline
		\hline
		\multicolumn{2}{|c|}{Two-Stage Detection Framework} \\
		\hline
		\hline
		Fast Point R-CNN~\cite{fastpointrcnn} & RefinerNet \\
		\hline
		PV-RCNN~\cite{pv-rcnn} & RoI-grid Pooling \\
		\hline
		PV-RCNN++~\cite{pv-rcnn++} & VectorPool \\
		\hline
		Pyramid R-CNN~\cite{pyramid-rcnn} & RoI-grid Attention \\
		\hline
		LiDAR R-CNN~\cite{lidar-rcnn} & Scale-aware Pooling\\
		\hline
		CT3D~\cite{ct3d} & Channel-wise Transformer \\
		\hline
	\end{tabular}
	\label{tab:pointvoxel_tax}
	\vspace{-2mm}
\end{table}

\noindent\textbf{Analysis: potentials and challenges of the point-voxel based methods.} The point-voxel based methods can naturally benefit from both the fine-grained 3D shape and structure information obtained from points and the computational efficiency brought by voxels. However, some challenges still exist in these methods. For the hybrid point-voxel backbones, the fusion of point and voxel features generally relies on the voxel-to-point and point-to-voxel transform mechanisms, which can bring non-negligible time costs. For the two-stage point-voxel detection frameworks, a critical challenge is how to efficiently aggregate point features for 3D proposals, as the existing modules and operators are generally time-consuming. In conclusion, compared to the pure voxel-based detection approaches, the point-voxel based detection methods can obtain a better detection accuracy while at the cost of increasing the inference time.

\subsubsection{Range-based 3D object detection}

\begin{figure}[t]
	\centering
	\includegraphics[width=0.48\textwidth]{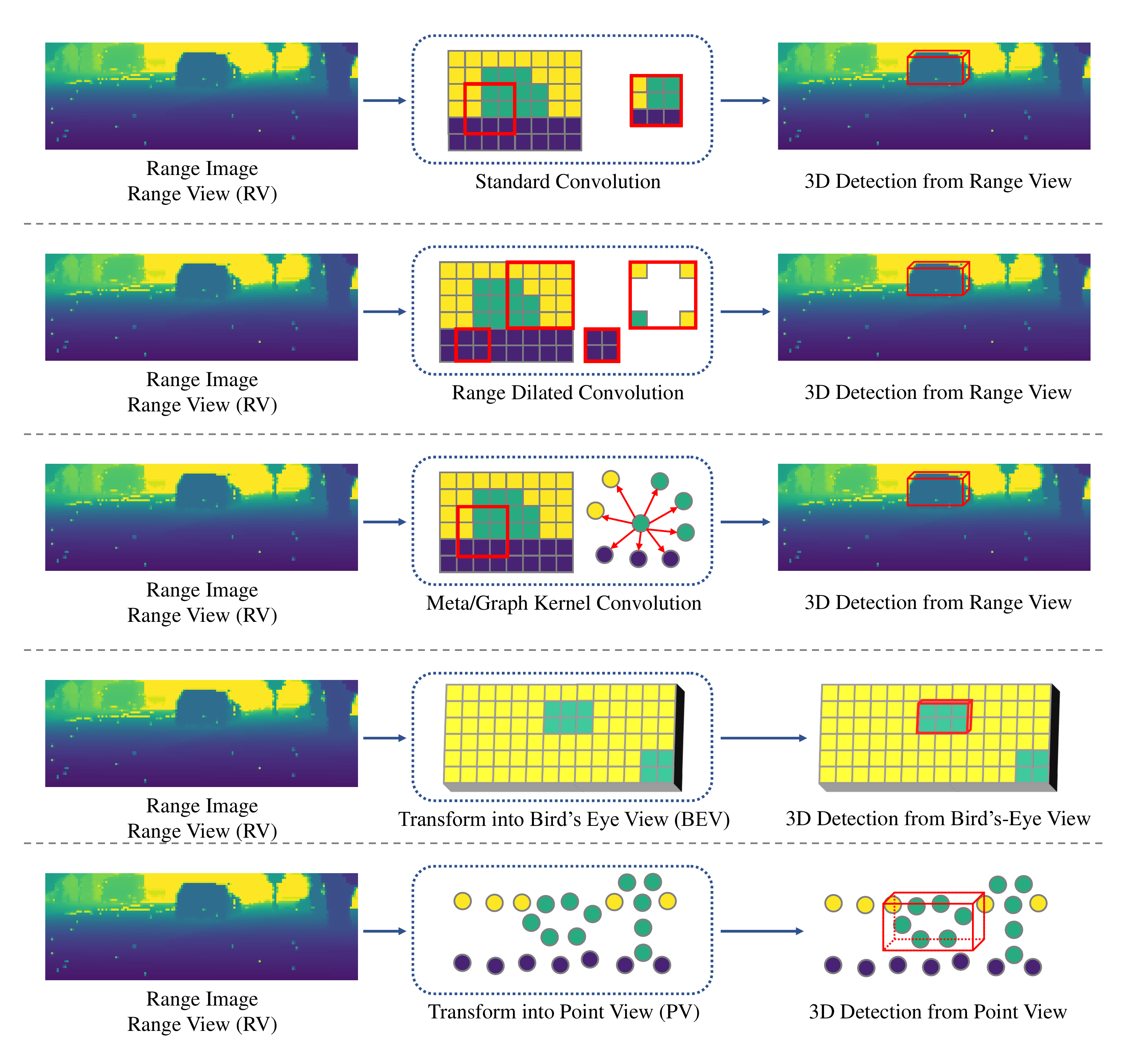}
	\caption{An illustration of range-based 3D object detection.} 
	\label{fig6-range-det}
	\vspace{-2mm}
\end{figure}

Range image is a dense and compact 2D representation in which each pixel contains 3D distance information instead of RGB values. Range-based methods address the detection problem from two aspects: designing new models and operators that are tailored for range images, and selecting suitable views for detection. An illustration of the range-based 3D object detection methods is shown in Figure~\ref{fig6-range-det} and a taxonomy is in Table~\ref{tab:range_tax}.   

\noindent\textbf{Range-based detection models.} Since range images are 2D representations like RGB images, range-based 3D object detectors can naturally borrow the models in 2D object detection to handle range images. LaserNet~\cite{lasernet} is a seminal work that leverages the deep layer aggregation network (DLA-Net)~\cite{dla} to obtain multi-scale features and detect 3D objects from range images. Some papers also adopt other 2D object detection architectures, \textit{e.g.} U-Net~\cite{u-net} is applied in~\cite{laserflow, rangercnn, rsn}, RPN~\cite{rpn} and R-CNN~\cite{faster-rcnn} are employed in~\cite{rangercnn, rcd}, FCN~\cite{fcn} is used in~\cite{rangeioudet}, and FPN~\cite{fpn} is leveraged in~\cite{rangedet}.

\noindent\textbf{Range-based operators.} Pixels of range images contain 3D distance information instead of color values, so the standard convolutional operator in conventional 2D network architectures is not optimal for range-based detection, as the pixels in a sliding window may be far away from each other in the 3D space. Some works resort to novel operators to effectively extract features from range pixels, including range dilated convolutions~\cite{rcd}, graph operators~\cite{tothepoint}, and meta-kernel convolutions~\cite{rangedet}.

\begin{table*}[t!]
	\caption{A taxonomy of range-based detection methods based on views, models, and operators.}
	\centering
	\begin{tabular}{|c|c|c|c|c|}
		\hline
		Method & View & Operator & Model & Note\\
		\hline
		\hline
		LaserNet~\cite{lasernet} & RV & Convolution & DLA-Net & -\\
		\hline
		Rapoport-Lavie \textit{et al.}~\cite{rangeguidednet} & RV, CYV & Convolution & Range-Guided Net & -\\
		\hline
		LaserFlow~\cite{laserflow} & RV, BEV & Convolution & U-Net & multi-sweep fusion\\
		\hline
		RangeRCNN~\cite{rangercnn} & RV, PV, BEV & Dilated Convolution & U-Net, RPN, RCNN & -\\
		\hline
		RangeIoUDet~\cite{rangeioudet} & RV, PV, BEV & Convolution & FCN, PointNet & point-wise segmentation\\
		\hline
		RCD~\cite{rcd} & RV & Conditioned Dilated Convolution & RPN, RCNN & -\\
		\hline
		RangeDet~\cite{rangedet} & RV & Meta Kernel Convolution & FPN & -\\
		\hline
		PPC~\cite{tothepoint} & RV & Graph Kernel Convolution & DLA-Net & -\\
		\hline
		RSN~\cite{rsn} & RV, BEV & Convolution & U-Net, VoxelNet & range-based segmentation\\
		\hline
	\end{tabular}
	\label{tab:range_tax}
	\vspace{-2mm}
\end{table*}

\noindent\textbf{Views for range-based detection.} Range images are captured from the range view (RV), and ideally, the range view is a spherical projection of a point cloud. It has been a natural solution for many range-based approaches~\cite{lasernet, rcd, rangedet, tothepoint} to detect 3D objects directly from the range view. Nevertheless, detection from the range view will inevitably suffer from the occlusion and scale-variation issues brought by the spherical projection. To circumvent these issues, many methods have been working on leveraging other views for predicting 3D objects, \textit{e.g.} the cylindrical view (CYV) leveraged in~\cite{rangeguidednet}, a combination of the range-view, bird's-eye view (BEV), and/or point-view (PV) adopted in~\cite{rangeioudet, rsn, laserflow, rangercnn}.

\noindent\textbf{Analysis: potentials and challenges of the range-based methods.} Range image is a dense and compact 2D representation, so the conventional or specialized 2D convolutions can be seamlessly applied on range images, which makes the feature extraction process quite efficient. Nevertheless, compared to bird's-eye view detection, detection from the range view is vulnerable to occlusion and scale variation. Hence, feature extraction from the range view and object detection from the bird's eye view becomes the most practical solution to range-based 3D object detection.   

\subsection{Learning objectives for 3D object detection} \label{sec:lidar_learning}

\noindent\textbf{Problem and Challenge.} Learning objectives are critical in object detection. Since 3D objects are quite small relative to the whole detection range, special mechanisms to enhance the localization of small objects are strongly required in 3D detection. On the other hand, considering point cloud is sparse and objects normally have incomplete shapes, accurately estimating the centers and sizes of 3D objects is a long-standing challenge. 

\subsubsection{Anchor-based 3D object detection} \label{sec:anchor-based}
\begin{figure}[t]
	\centering
	\includegraphics[width=0.45\textwidth]{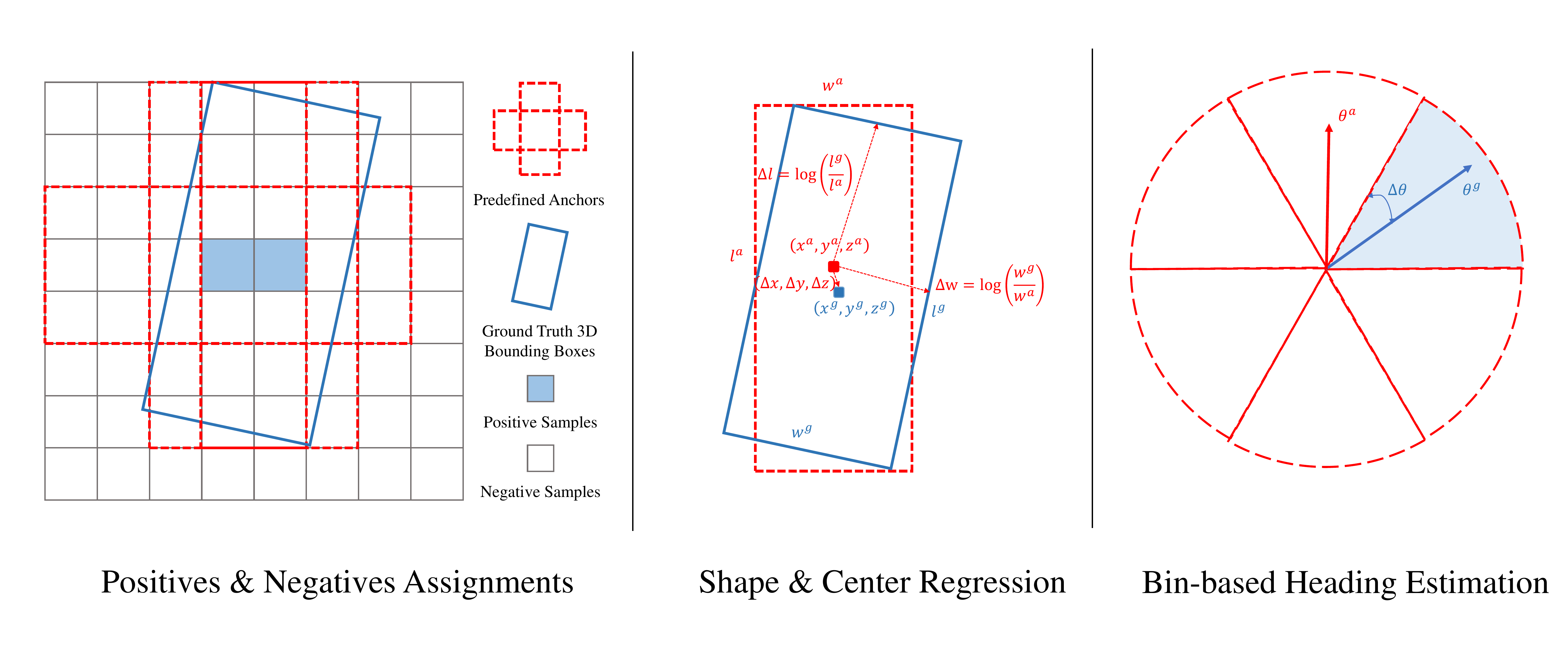}
	\caption{An illustration of anchor-based learning objectives.} 
	\label{fig7-anchor-based}
	\vspace{-2mm}
\end{figure}

Anchors are pre-defined cuboids with fixed shapes that can be placed in the 3D space. 3D objects can be predicted based on the positive anchors that have a high intersection over union (IoU) with ground truth. We will introduce the anchor-based 3D object detection methods from the aspect of anchor configurations and loss functions. An illustration of anchor-based learning objectives is shown in Figure~\ref{fig7-anchor-based} and a taxonomy is in Table~\ref{tab:anchor_tax}.

\noindent\textbf{Prerequisites.} The ground truth 3D objects can be represented as $[x^{g}, y^{g}, z^{g}, l^{g}, w^{g}, h^{g}, \theta^{g}]$ with the class $cls^{g}$. The anchors $[x^{a}, y^{a}, z^{a}, l^{a}, w^{a}, h^{a}, \theta^{a}]$ are used to generate predicted 3D objects $[x, y, z, l, w, h, \theta]$ with a predicted class probability $p$.

\noindent\textbf{Anchor configurations.} Anchor-based 3D object detection approaches generally detect 3D objects from the bird's-eye view, in which 3D anchor boxes are placed at each grid cell of a BEV feature map. 3D anchors normally have a fixed size for each category, since objects of the same category have similar sizes.

\noindent\textbf{Loss functions.} The anchor-based methods employ the classification loss $L_{cls}$ to learn the positive and negative anchors, and the regression loss $L_{reg}$ is utilized to learn the size and location of an object based on a positive anchor. Additionally, $L_{\theta}$ is applied to learn the object's heading angle. The loss function is
\begin{equation}\label{eq:det_loss_anchor}
	L_{det} = L_{cls} + L_{reg} + L_{\theta}.
\end{equation}

VoxelNet~\cite{voxelnet18} is a seminal work that leverages the anchors that have a high IoU with the ground truth 3D objects as positive anchors, and the other anchors are treated as negatives. To accurately classify those positive and negative anchors, for each category, the binary cross entropy loss can be applied to each anchor on the BEV feature map, which can be formulated as
\begin{equation} \label{eq:bce_cls_loss}
	L^{bce}_{cls} = - [q \cdot \log(p) + (1-q) \cdot \log(1 - p)],
\end{equation}
where $p$ is the predicted probability for each anchor and the target $q$ is $1$ if the anchor is positive and $0$ otherwise. In addition to the binary cross entropy loss, the focal loss~\cite{focalloss, focalloss3d} has also been employed to enhance the localization ability:
\begin{equation} \label{eq:focal_loss}
	L^{focal}_{cls} = - \alpha (1 - p)^{\gamma}\log(p),
\end{equation}
where $\alpha=0.25$ and $\gamma=2$ are adopted in most works.

The regression targets can be further applied to those positive anchors to learn the sizes and locations of 3D objects:
\begin{equation}
	\begin{split}
		\Delta x = \frac{x^{g} - x^{a}}{d^{a}}, \Delta y = \frac{y^{g} - y^{a}}{d^{a}},
		\Delta z = \frac{z^{g} - z^{a}}{h^{a}}, \\ \Delta l = \log(\frac{l^{g}}{l^{a}}), \Delta w = \log(\frac{w^{g}}{w^{a}}),
		\Delta h = \log(\frac{h^{g}}{h^{a}}),
	\end{split}
\end{equation}
where $d^{a} = \sqrt{(l^{a})^2+(w^{a})^2}$ is the diagonal length of an anchor from the bird's-eye view. Then the SmoothL1 loss~\cite{faster-rcnn} is adopted to regress the targets, which is represented as
\begin{equation} \label{eq:reg_loss}
	L_{reg} = \sum_{\substack{u \in \{x, y, z, l, w, h\}, \\ v \in \{\Delta x, \Delta y, \Delta z, \Delta l, \Delta w, \Delta h\}}} {\rm SmoothL1}(u - v).
\end{equation}

To learn the heading angle $\theta$, the radian orientation offset can be directly regressed with the SmoothL1 loss:
\begin{equation}\label{eq:reg1}
	\begin{split}
		&\Delta \theta = \theta^g - \theta^a, \\
		&L_{\theta} = {\rm SmoothL1}(\theta - \Delta \theta).
	\end{split}
\end{equation} 
However, directly regressing the radian offset is normally hard due to the large regression range. Alternatively, the bin-based heading estimation~\cite{fpointnet18} is a better solution to learn the heading angle, in which the angle space is first divided into bins, and bin-based classification $L_{dir}$ and residual regression are employed: 
\begin{equation} \label{eq:reg2}
	L_{\theta} = L_{dir} + {\rm SmoothL1}(\theta - \Delta \theta^{\prime}),
\end{equation}
where $\Delta \theta^{\prime}$ is the residual offset within a bin. The sine function can also be utilized to encode the radian offset:
\begin{equation} \label{eq:reg3}
	\begin{split}
		\Delta \theta = \sin(\theta^g - \theta^a),
	\end{split}
\end{equation}
and $L^{\theta}$ can be computed following Eqn.~\ref{eq:reg1} or Eqn.~\ref{eq:reg2}.

In addition to the loss functions that learn the objects' sizes, locations, and orientations separately, the intersection over union (IoU) loss~\cite{iouloss} that considers all object parameters as a whole can also be applied in 3D object detection:
\begin{equation} \label{eq:iou_loss}
	L_{IoU} = 1 - IoU(b^g, b),
\end{equation}
where $b^g$ and $b$ are the ground truth and predicted 3D bounding boxes, and $IoU(\cdot)$ calculates the 3D IoU in a differential manner. Apart from the IoU loss, the corner loss~\cite{fpointnet18} is also introduced to minimize the distances between the eight corners of the ground truth and predicted boxes, that is
\begin{equation} \label{eq:corner_loss}
	L_{corner} = \sum^{8}_{i=1} ||c^{g}_{i} - c_{i}||,
\end{equation}
where $c^{g}_{i}$ and $c_{i}$ are the $i$th corner of the ground truth and predicted cuboid respectively.

\noindent\textbf{Analysis: potentials and challenges of the anchor-based approaches.} The anchor-based methods can benefit from the prior knowledge that 3D objects of the same category should have similar shapes, so they can generate accurate object predictions with the help of 3D anchors. However, since 3D objects are relatively small with respect to the detection range, a large number of anchors are required to ensure complete coverage of the whole detection range, \textit{e.g.} around $70$k anchors are utilized in~\cite{second} on the KITTI~\cite{kitti12conf} dataset. Furthermore, for those extremely small objects such as pedestrians and cyclists, applying anchor-based assignments can be quite challenging. Considering the fact that anchors are generally placed at the center of each grid cell, if the grid cell is large and objects in the cell are small, the anchor of this cell may have a low IoU with the small objects, which may hamper the training process.  

\begin{table}[t!]
	\caption{A taxonomy of anchor-based methods based on loss functions.}
	\centering
	\begin{tabular}{|c|c|}
		\hline
		Loss Function & Methods\\
		\hline
		\hline
		$L^{bce}_{cls}$ (Eqn.~\ref{eq:bce_cls_loss}) & \tabincell{c}{\cite{complexyolo, yolo3d, birdnet, rt3d, voxelnet18, birdnet+, std, ipod, fastpointrcnn}\\ \cite{svganet}} \\
		\hline
		$L^{focal}_{cls}$ (Eqn.~\ref{eq:focal_loss}) & \tabincell{c}{\cite{second, pointpillars, mvfwaymo, iouloss, focalloss3d, hvnet, associate3ddet, ssn}\\ \cite{reconfigurablevoxels, segvoxelnet, votr, voxelrcnn, ciassd, ago-net, rangercnn, starnet}\\ \cite{pv-rcnn, sassd, infofocus, pvgnet, hvpr, pyramid-rcnn, pv-rcnn++}} \\
		\hline
		$L_{reg}$ (Eqn.~\ref{eq:reg_loss}) & \tabincell{c}{\cite{complexyolo, yolo3d, birdnet, rt3d, second, voxelnet18, pointpillars, mvfwaymo, focalloss3d} \\ \cite{hvnet, associate3ddet, ssn, reconfigurablevoxels, segvoxelnet, birdnet+, votr, voxelrcnn} \\ \cite{ciassd, ago-net, rangercnn, std, starnet, ipod, fastpointrcnn, pv-rcnn} \\ \cite{sassd, infofocus, svganet, pvgnet, hvpr, pyramid-rcnn, pv-rcnn++}} \\
		\hline
		$L^{1}_{\theta}$ (Eqn.~\ref{eq:reg1}) & \cite{yolo3d, voxelnet18, hvnet, associate3ddet, ago-net, fastpointrcnn} \\
		\hline
		$L^{2}_{\theta}$ (Eqn.~\ref{eq:reg2}) & \tabincell{c}{\cite{birdnet, reconfigurablevoxels, segvoxelnet, birdnet+, votr, ciassd, rangercnn, std, ipod} \\ \cite{pv-rcnn, sassd, infofocus, pyramid-rcnn, pv-rcnn++}}\\
		\hline
		$L^{3}_{\theta}$ (Eqn.~\ref{eq:reg3}) & \tabincell{c}{\cite{complexyolo, rt3d, second, pointpillars, mvfwaymo, ssn, voxelrcnn, starnet, svganet}\\ \cite{pvgnet, hvpr}} \\
		\hline
		$L_{IoU}$ (Eqn.~\ref{eq:iou_loss}) & \cite{iouloss} \\
		\hline
		$L_{corner}$ (Eqn.~\ref{eq:corner_loss}) & \cite{hvnet, rangercnn, std, ipod}\\
		\hline    
	\end{tabular}
	\label{tab:anchor_tax}
	\vspace{-2mm}
\end{table}

\subsubsection{Anchor-free 3D object detection}

\begin{figure}[t]
	\centering
	\includegraphics[width=0.45\textwidth]{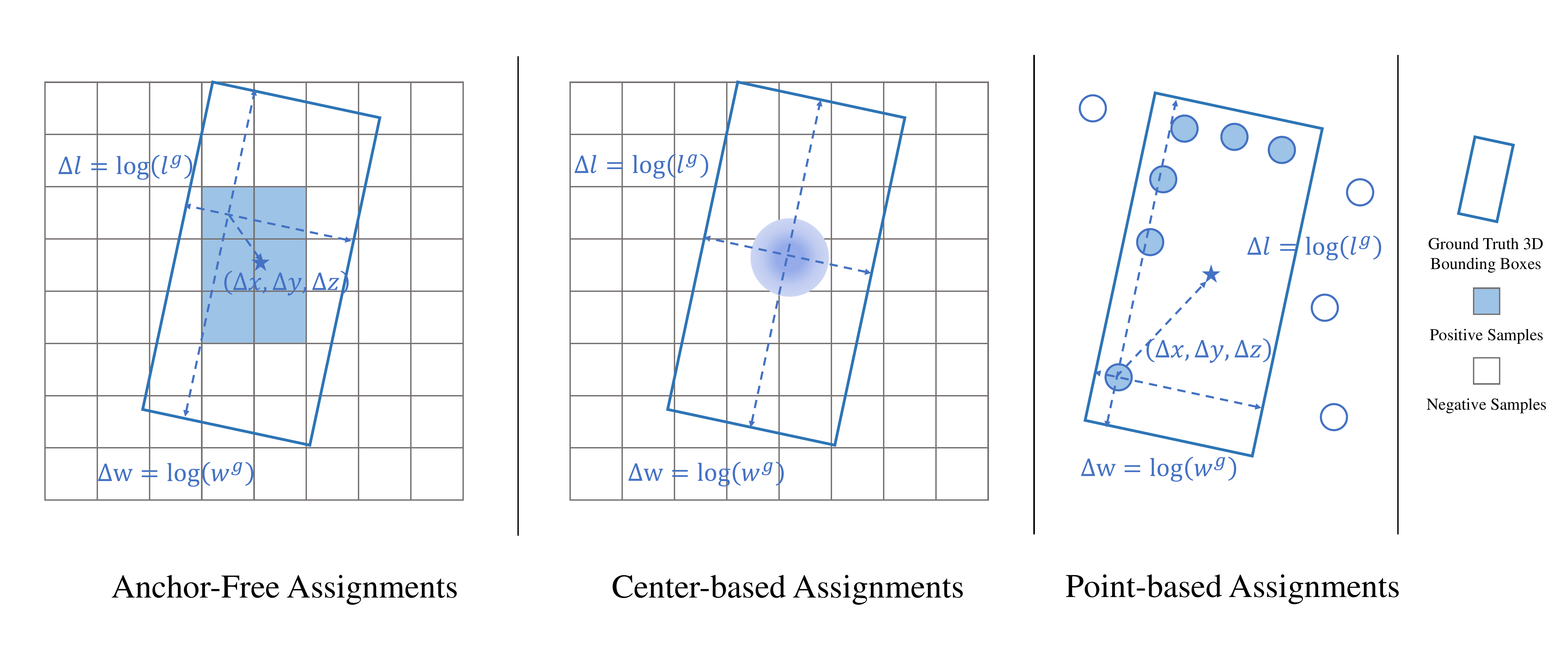}
	\caption{An illustration of anchor-free learning objectives.} 
	\label{fig8-anchor-free}
	\vspace{-2mm}
\end{figure}

Anchor-free approaches eliminate the complicated anchor designs and can be flexibly applied to diverse views, \textit{e.g.} the bird's-eye view, point view, and range view. An illustration of anchor-free learning objectives is shown in Figure~\ref{fig8-anchor-free} and a taxonomy is in Table~\ref{tab:anchor_free_tax}. The major difference between the anchor-based and anchor-free methods lies in the selection of positive and negative samples. We will introduce the anchor-free methods from the perspective of positive assignments, including grid-based, point-based, range-based, and set-to-set assignments. We still adopt the notations in Section~\ref{sec:anchor-based} for simplicity.

\noindent\textbf{Grid-based assignment.} In contrast to the anchor-based methods that rely on the IoUs with anchors to determine the positive and negative samples, the anchor-free methods leverage various grid-based assignment strategies for BEV grid cells, pillars, and voxels. PIXOR~\cite{pixor} is a pioneering work that leverages the grid cells inside the ground truth 3D objects as positives, and the others as negatives. This inside-object assignment strategy is adopted in~\cite{pillar-od}, and further improved in~\cite{afdet, afdetv2, hotspotnet} by selecting the grid cells nearest to the object center. CenterPoint~\cite{centerpoint} utilizes a Gaussian kernel at each object center to assign positive labels. These methods can still use Eqn.~\ref{eq:bce_cls_loss} or Eqn.~\ref{eq:focal_loss} as the classification loss, and the regression target is
\begin{equation} \label{eq:anchor_free_reg}
	\Delta = [dx, dy, z^g, \log(l^g), \log(w^g), \log(h^g), \sin(\theta^g), \cos(\theta^g)],
\end{equation}
where $dx$ and $dy$ are the offsets between positive grid cells and object centers. The SmoothL1 loss is leveraged to regress $\Delta$.

\noindent\textbf{Point-based assignment.} Most point-based detection approaches resort to the anchor-free and point-based assignment strategy, in which the points are first segmented and those foreground points inside or near 3D objects are selected as positive samples, and 3D bounding boxes are finally learned from those foreground points. This foreground point segmentation strategy has been adopted in most point-based detectors~\cite{pointrcnn19, 3dssd, ipod, pointformer}, with improvements such as adding centerness scores~\cite{3dssd}, \textit{etc.}

\noindent\textbf{Range-based assignment.} Anchor-free assignments can also be employed on range images. A common solution is to select the range pixels inside 3D objects as positive samples, which has been adopted in~\cite{lasernet, rangedet}. Different from other methods where the regression targets are based on the global 3D coordinate system, the range-based methods resort to an object-centric coordinate system for regression. Eqn.~\ref{eq:anchor_free_reg} can still be applied in these methods with an additional coordinate transform. 

\begin{table}[t!]
	\caption{A taxonomy of anchor-free detection methods based on the sample types for prediction and the assignment strategies.}
	\centering
	\begin{tabular}{|c|c|c|}
		\hline
		Method & \tabincell{c}{Samples \\ for Prediction} & \tabincell{c}{Positive Samples \\Selection} \\
		\hline
		\hline
		PIXOR~\cite{pixor} & BEV grid cells & inside objects\\
		\hline
		CenterPoint~\cite{centerpoint} & BEV grid cells & Gaussian radii on centers \\
		\hline
		ObjectDGCNN~\cite{objectdgcnn} & BEV grid cells & bipartite matching\\
		\hline
		Pillar-OD~\cite{pillar-od} & pillars & inside objects\\
		\hline
		HotSpotNet~\cite{hotspotnet} & voxels & K-nearest to object centers \\
		\hline
		PointRCNN~\cite{pointrcnn19} & points & foreground \\
		\hline
		3DSSD~\cite{3dssd} & points & foreground \& near centers \\
		\hline
		LaserNet~\cite{lasernet} & range pixels & inside objects\\
		\hline
		RangeDet~\cite{rangedet} & range pixels & inside objects\\
		\hline
	\end{tabular}
	\label{tab:anchor_free_tax}
	\vspace{-2mm}
\end{table}

\noindent\textbf{Set-to-set assignment.} DETR~\cite{detr} is an influential 2D detection method that introduces a set-to-set assignment strategy to automatically assign the predictions to the respective ground truths via the Hungarian algorithm~\cite{hungarian}:
\begin{equation} \label{eq:hunarian}
	\mathcal{M}^{*} = \mathop{\rm argmin}\limits_{\mathcal{M}} \sum_{(i \rightarrow j) \in \mathcal{M}} L_{det}(b^g_i, b_j),
\end{equation}
where $\mathcal{M}$ is a one-to-one mapping from each positive sample to a 3D object. The set-to-set assignments have also been explored in 3D object detection approaches~\cite{3detr, objectdgcnn, point2seq}, and~\cite{point2seq} further introduces a novel cost function for the Hungarian matching. 

\noindent\textbf{Analysis: potentials and challenges of the anchor-free approaches.} The anchor-free detection methods abandon the complicated anchor design and exhibit stronger flexibility in terms of the assignment strategies. With the anchor-free assignments, 3D objects can be predicted directly on various representations, including points, range pixels, voxels, pillars, and BEV grid cells. The learning process is also greatly simplified without introducing additional shape priors. Among those anchor-free methods, the center-based methods~\cite{centerpoint} have shown great potential in detecting small objects and have outperformed the anchor-based detection methods on the widely used benchmarks~\cite{nuscenes20, waymo20}.   

Despite these merits, a general challenge to the anchor-free methods is to properly select positive samples to generate 3D object predictions. In contrast to the anchor-based methods that only select those high IoU samples, the anchor-free methods may possibly select some bad positive samples that yield inaccurate object predictions. Hence, careful design to filter out those bad positives is important in most anchor-free methods.

\subsubsection{3D object detection with auxiliary tasks}

Numerous approaches resort to auxiliary tasks to enhance the spatial features and provide implicit guidance for accurate 3D object detection. The commonly used auxiliary tasks include semantic segmentation, intersection over union prediction, object shape completion, and object part estimation.

\noindent\textbf{Semantic segmentation.} Semantic segmentation can help 3D object detection in $3$ aspects: (1) Foreground segmentation could provide implicit information on objects' locations. Point-wise foreground segmentation has been broadly adopted in most point-based 3D object detectors~\cite{pointrcnn19, joint3dbaidu, std, sassd} for proposal generation. (2) Spatial features can be enhanced by segmentation. In~\cite{segvoxelnet}, a semantic context encoder is leveraged to enhance spatial features with semantic knowledge. (3) Semantic segmentation can be utilized as a pre-processing step to filter out background samples and make 3D object detection more efficient. \cite{ipod} and ~\cite{rsn} leverage semantic segmentation to remove those redundant points to speed up the subsequent detection model.  

\noindent\textbf{IoU prediction.} Intersection over union (IoU) can serve as a useful supervisory signal to rectify the object confidence scores. \cite{ciassd} proposes an auxiliary branch to predict an IoU score $S_{IoU}$ for each detected 3D object. During inference, the original confidence scores $S_{conf} = S_{cls}$ from the conventional classification branch are further rectified by the IoU scores $S_{IoU}$: 
\begin{equation}
	S_{conf} = S_{cls} \cdot (S_{IoU})^{\beta},
\end{equation}
where the hyper-parameter $\beta$ controls the degrees of suppressing the low-IoU predictions and enhancing the high-IoU predictions. With the IoU rectification, the high-quality 3D objects are easier to be selected as the final predictions. Similar designs have also been adopted in~\cite{sessd, rangeioudet, afdet, afdetv2}.

\begin{table}[t!]
	\caption{A taxonomy of detection methods based on auxiliary tasks.}
	\centering
	\begin{tabular}{|c|c|}
		\hline
		Auxiliary Task & Methods \\
		\hline
		\hline
		Semantic segmentation & \tabincell{c}{\cite{pointrcnn19, std, sassd, joint3dbaidu, segvoxelnet,ipod, rsn}} \\
		\hline
		IoU prediction &  \tabincell{c}{\cite{ciassd, sessd, rangeioudet, afdet, afdetv2}} \\
		\hline
		Object shape completion & \tabincell{c}{\cite{dops, ssn, spg, behind_the_curtain}} \\
		\hline
		Object part estimation & \tabincell{c}{\cite{hotspotnet, parta2}} \\
		\hline
	\end{tabular}
	\label{tab:aux_task_tax}
\end{table}

\noindent\textbf{Object shape completion.} Due to the nature of LiDAR sensors, faraway objects generally receive only a few points on their surfaces, so 3D objects are generally sparse and incomplete. A straightforward way of boosting the detection performance is to complete object shapes from sparse point clouds. Complete shapes could provide more useful information for accurate and robust detection. Many shape completion techniques have been proposed in 3D detection, including a shape decoder~\cite{dops}, shape signatures~\cite{ssn}, and a probabilistic occupancy grid~\cite{spg, behind_the_curtain}.

\noindent\textbf{Object part estimation.} Identifying the part information inside objects is helpful in 3D object detection, as it reveals more fine-grained 3D structure information of an object. Object part estimation has been explored in some works~\cite{hotspotnet, parta2}.      

\noindent\textbf{Analysis: future prospects of multitask learning for 3D object detection.} 3D object detection is innately correlated with many other 3D perception and generation tasks. Multitask learning of 3D detection and segmentation is more beneficial compared to training 3D object detectors independently, and shape completion can also help 3D object detection. There are also other tasks that can help boost the performance of 3D object detectors. For instance, scene flow estimation could identify static and moving objects, and tracking the same 3D object in a point cloud sequence yields a more accurate estimation of this object. Hence, it will be promising to integrate more perception tasks into the existing 3D object detection pipeline.  

\section{Camera-based 3D Object Detection} \label{sec:camera}

\begin{figure*}[t]
	\centering
	\includegraphics[width=\textwidth]{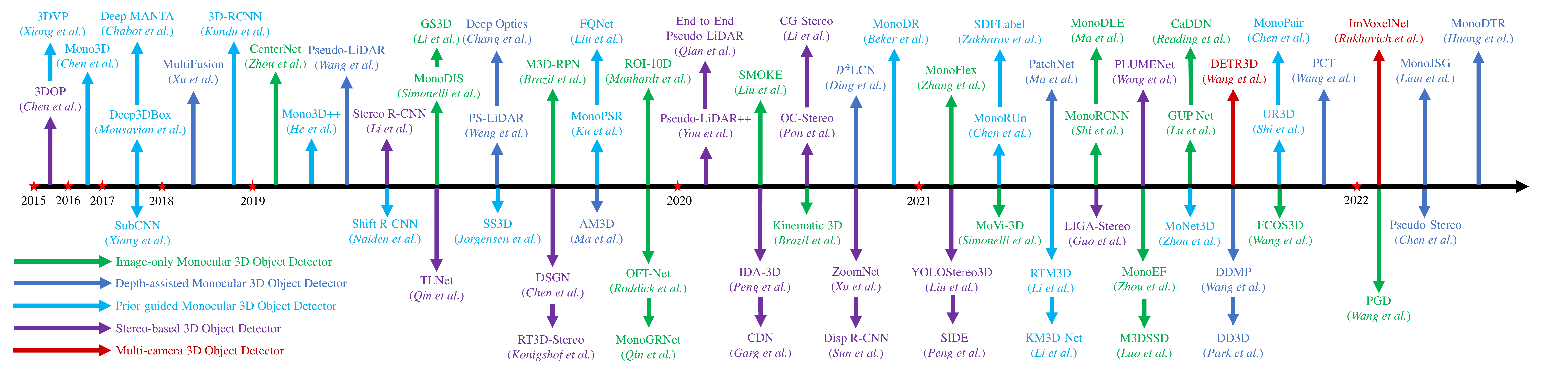}
	\caption{Chronological overview of the camera-based 3D object detection methods.}
	\label{fig9-camera-roadmap}
	\vspace{-2mm}
\end{figure*}

In this section, we introduce camera-based 3D object detection methods. In Section~\ref{sec:monocular}, we review and analyze the monocular 3D object detection methods, which can be further divided into the image-only, depth-assisted, and prior-guided approaches. In Section~\ref{sec:stereo}, we investigate the 3D object detection methods based on stereo images. In Section~\ref{sec:multi-camera}, we introduce the 3D object detection methods with multiple cameras. A chronological overview of the camera-based 3D object detection methods is shown in Figure~\ref{fig9-camera-roadmap}. 

\subsection{Monocular 3D object detection} \label{sec:monocular}

\noindent\textbf{Problem and Challenge.} Detecting objects in the 3D space from monocular images is an ill-posed problem since a single image cannot provide sufficient depth information. Accurately predicting the 3D locations of objects is the major challenge in monocular 3D object detection. Many endeavors have been made to tackle the object localization problem, \textit{e.g.} inferring depth from images, leveraging geometric constraints and shape priors. Nevertheless, the problem is far from being solved. Monocular 3D detection methods still perform much worse than the LiDAR-based methods due the poor 3D localization ability, which leaves an open challenge to the research community.

\subsubsection{Image-only monocular 3D object detection}

\begin{figure}[t]
	\centering
	\includegraphics[width=0.45\textwidth]{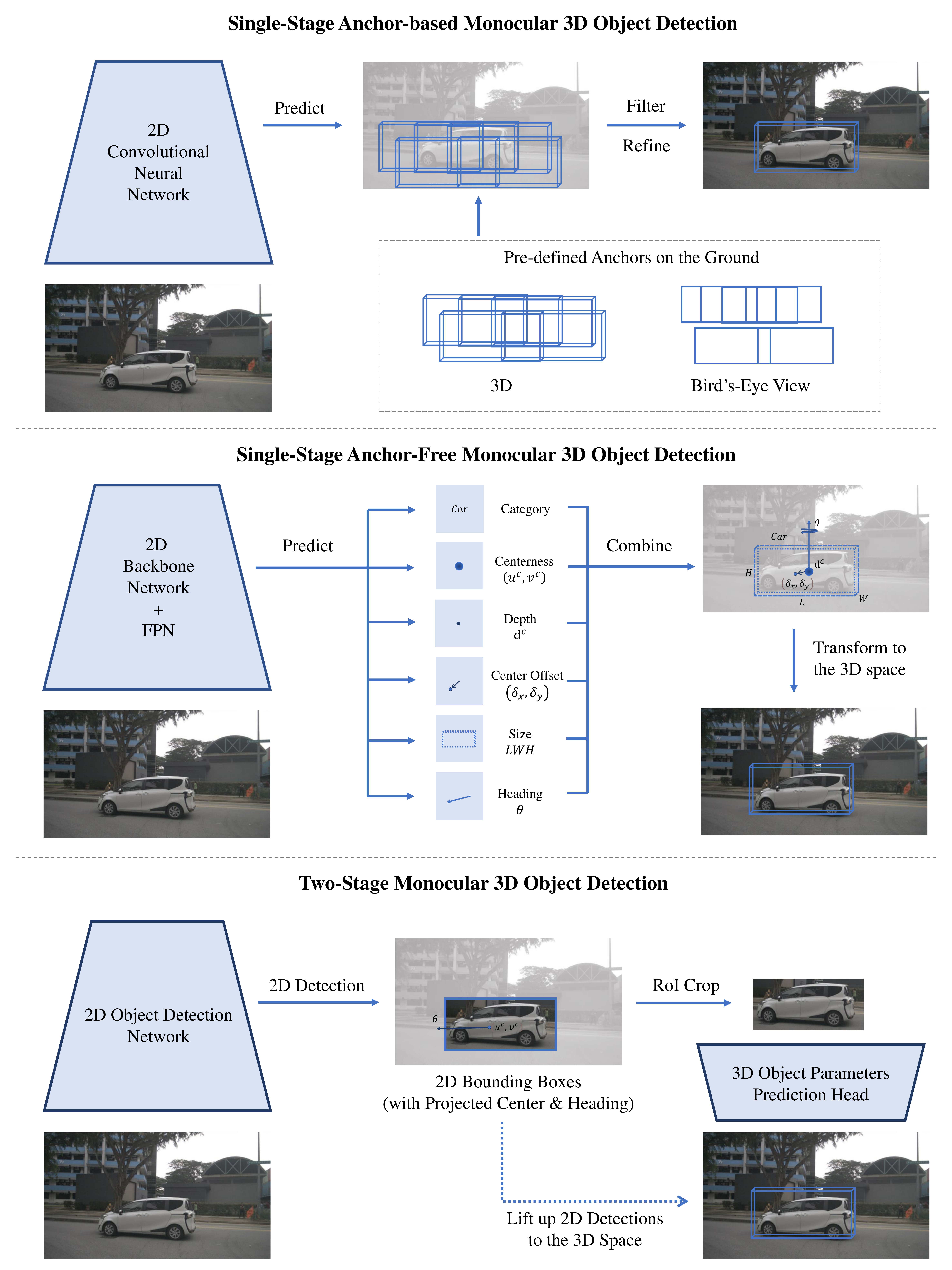}
	\caption{An illustration of image-only monocular 3D object detection methods. Image samples are from \cite{FCOS3D}.} 
	\label{fig10-image-det}
	\vspace{-2mm}
\end{figure}

Inspired by the 2D detection approaches, a straightforward solution to monocular 3D object detection is to directly regress the 3D box parameters from images via a convolutional neural network. The direct-regression methods naturally borrow designs from the 2D detection network architectures, and can be trained in an end-to-end manner. These approaches can be divided into the single-stage/two-stage, or anchor-based/anchor-free methods. An illustration of image-only 3D object detection is shown in Figure~\ref{fig10-image-det} and a taxonomy is in Table~\ref{tab:mono_image_tax}. 

\noindent\textbf{Single-stage anchor-based methods.} Anchor-based monocular detection approaches rely on a set of 2D-3D anchor boxes placed at each image pixel, and use a 2D convolutional neural network to regress object parameters from the anchors. Specifically, for each pixel $[u, v]$ on the image plane, a set of 3D anchors $[w^a, h^a, l^a, \theta^a]_{3D}$, 2D anchors $[w^a, h^a]_{2D}$, and depth anchors $d^a$ are pre-defined. An image is passed through a convolutional network to predict the 2D box offsets $\delta_{2D} = [\delta_x, \delta_y, \delta_w, \delta_h]_{2D}$ and the 3D box offsets $\delta_{3D} = [\delta_x, \delta_y, \delta_d, \delta_w, \delta_h, \delta_l, \delta_{\theta}]_{3D}$ based on each anchor. Then, the 2D bounding boxes $b_{2D} = [x, y, w, h]_{2D}$ can be decoded as
\begin{equation}
	\begin{split}
		&[x, y]_{2D} = [u, v] + [\delta_x, \delta_y]_{2D} \cdot [w^a, h^a]_{2D}, \\
		&[w, h]_{2D} = e^{[\delta_w, \delta_h]_{2D}} \cdot [w^a, h^a]_{2D}, 
	\end{split}
\end{equation}
and the 3D bounding boxes $b_{3D} = [x, y, z, l, w, h, \theta]_{3D}$ can be decoded from the anchors and $\delta_{3D}$:
\begin{equation}
	\begin{split}
		&[u^{c}, u^{c}] = [u, v] + [\delta_x, \delta_y]_{3D} \cdot [w^a, h^a]_{2D},\\
		&[w, h, l]_{3D} = e^{[\delta_w, \delta_h, \delta_l]_{3D}} \cdot [w^a, h^a, l^a]_{3D}, \\
		&d^{c} = d^a + \delta_{d_{3D}}, \theta_{3D} = \theta^a_{3D} + \delta_{\theta_{3D}}, 
	\end{split}
\end{equation}
where $[u^{c}, v^{c}]$ is the projected object center on the image plane. Finally, the projected center $[u^{c}, v^{c}]$ and its depth $d^{c}$ are transformed into the 3D object center $[x, y, z]_{3D}$:
\begin{equation}
	d^{c} \cdot
	\begin{bmatrix}
		u^{c} \\ v^{c} \\ 1
	\end{bmatrix} = K \ T \ \begin{bmatrix}
		x \\ y \\ z \\ 1
	\end{bmatrix}_{3D},
\end{equation}
where $K$ and $T$ are the camera intrinsics and extrinsics.

M3D-RPN~\cite{m3d-rpn} is a seminal paper that proposes the anchor-based framework, and many papers have tried to improve this framework, \textit{e.g.} extending it into video-based 3D detection~\cite{Kinematic3D}, introducing differential non-maximum suppression~\cite{Mono-NMS}, designing an asymmetric attention module~\cite{M3DSSD}.

\begin{table}[t]
	\caption{A taxonomy of image-only monocular detection methods based on frameworks.}
	\centering
	\begin{tabular}{|c|c|c|}
		\hline
		\multicolumn{2}{|c|}{Framework}  & Methods \\
		\hline
		\hline
		\multirow{2}*{Single-stage} & anchor-based & \tabincell{c}{\cite{m3d-rpn, Kinematic3D, Mono-NMS, M3DSSD, FQNet}} \\
		\cline{2-3}
		&  anchor-free &\tabincell{c}{\cite{centernet, deep3dbox, SMOKE, FCOS3D, PGD, MonoDLE} \\ \cite{RTM3D, MonoFlex, exfree-mono, oft-net, CaDDN, MoVi-3D}} \\
		\hline
		\multicolumn{2}{|c|}{Two-stage} & \tabincell{c}{\cite{ROI-10D, monodis, GS3D, monogrnet, MonoRCNN, GUPNet}} \\
		\hline
	\end{tabular}
	\label{tab:mono_image_tax}
	\vspace{-2mm}
\end{table}

\noindent\textbf{Single-stage anchor-free methods.} Anchor-free monocular detection approaches predict the attributes of 3D objects from images without the aid of anchors. Specifically, an image is passed through a 2D convolutional neural network and then multiple heads are applied to predict the object attributes separately. The prediction heads generally include a category head to predict the object's category, a keypoint head to predict the coarse object center $[u, v]$, an offset head to predict the center offset $[\delta_{x}, \delta_{y}]$ based on $[u, v]$, a depth head to predict the depth offset $\delta_d$, a size head to predict the object size $[w, h, l]$, and an orientation head to predict the observation angle $\alpha$. The 3D object center $[x, y, z]$ can be converted from the projected center $[u^c, v^c]$ and depth $d^c$:
\begin{equation}
	\begin{split}
		&d^c = \sigma^{-1}(\frac{1}{\delta_d+1}), u^c = u + \delta_x, v^c = v + \delta_y, \\
		&d^{c} \cdot
		\begin{bmatrix}
			u^{c} \\ v^{c} \\ 1
		\end{bmatrix} = K \ T \ \begin{bmatrix}
			x \\ y \\ z \\ 1
		\end{bmatrix}_{3D},
	\end{split}
\end{equation}
where $\sigma$ is the sigmoid function. The yaw angle $\theta$ of an object can be converted from the observation angle $\alpha$ using
\begin{equation}
	\theta = \alpha + {\rm arctan}(\frac{x}{z}).
\end{equation}

CenterNet~\cite{centernet} first introduces the single-stage anchor-free framework for monocular 3D object detection. Many following papers work on improving this framework, including novel depth estimation schemes~\cite{SMOKE, PGD, MonoFlex}, an FCOS~\cite{fcos}-like architecture~\cite{FCOS3D}, a new IoU-based loss function~\cite{MonoDLE}, keypoints~\cite{RTM3D}, pair-wise relationships~\cite{MonoPair}, camera extrinsics prediction~\cite{exfree-mono}, and view transforms~\cite{oft-net, CaDDN, MoVi-3D}.

\noindent\textbf{Two-stage methods.} Two-stage monocular detection approaches generally extend the conventional two-stage 2D detection architectures to 3D object detection. Specifically, they utilize a 2D detector in the first stage to generate 2D bounding boxes from an input image. Then in the second stage, the 2D boxes are lifted up to the 3D space by predicting the 3D object parameters from the 2D RoIs. ROI-10D~\cite{ROI-10D} extends the conventional Faster R-CNN~\cite{faster-rcnn} architecture with a novel head to predict the parameters of 3D objects in the second stage. A similar design paradigm has been adopted in many works with improvements like disentangling the 2D and 3D detection loss~\cite{monodis}, predicting heading angles in the first stage~\cite{GS3D}, learning more accurate depth information~\cite{monogrnet, MonoRCNN, GUPNet}.

\noindent\textbf{Analysis: potentials and challenges of the image-only methods.} The image-only methods aim to directly regress the 3D box parameters from images via a modified 2D object detection framework. Since these methods take inspiration from the 2D detection methods, they can naturally benefit from the advances in 2D object detection and image-based network architectures. Most methods can be trained end-to-end without pre-training or post-processing, which is quite simple and efficient.

A critical challenge of the image-only methods is to accurately predict depth $d^c$ for each 3D object. As shown in~\cite{PGD}, simply replacing the predicted depth with ground truth yields more than $20\%$ car AP gain on the KITTI~\cite{kitti12conf} dataset, while replacing other parameters only results in an incremental gain. This observation indicates that the depth error dominates the total errors and becomes the most critical factor hampering accurate monocular detection. Nevertheless, depth estimation from monocular images is an ill-posed problem, and the problem becomes severer with only box-level supervisory signals.

\subsubsection{Depth-assisted monocular 3D object detection}

\begin{table}[t!]
	\caption{A taxonomy of depth-assisted monocular detection methods based on data representations and detection networks (2D: convolutional networks; 3D: point cloud networks).}
	\centering
	\begin{tabular}{|c|c|c|c|c|c|c|}
		\hline
		\multirow{2}*{Method} & \multicolumn{4}{c|}{Representation} &  \multicolumn{2}{c|}{Network} \\
		\cline{2-7}
        & RGB & Depth & \tabincell{c}{Pseudo \\ LiDAR} & \tabincell{c}{Coord. \\ Map} & 2D & 3D \\ 
		\hline
		MultiFusion~\cite{multifusion} & \checkmark & \checkmark & & & \checkmark & \\
		\hline
		D$^4$LCN~\cite{D4LCN} &  \checkmark & \checkmark & & & \checkmark & \\
		\hline
		DDMP~\cite{DDMP} &  \checkmark & \checkmark & & & \checkmark & \\
		\hline
		\tabincell{c}{Pseudo-\\LiDAR~\cite{pslidar}} & & & \checkmark & & & \checkmark \\
		\hline
		Deep Optics~\cite{deep-optics} & & & \checkmark & & & \checkmark \\
		\hline
		AM3D~\cite{color-pslidar} & \checkmark & & \checkmark & & \checkmark & \checkmark \\
		\hline
		Weng \textit{et al.}~\cite{end-to-end-plidar} &  \checkmark & & \checkmark & & \checkmark & \checkmark \\
		\hline
		PatchNet~\cite{PatchNet} & & & & \checkmark & \checkmark & \\
		\hline
	\end{tabular}
	\label{tab:mono_depth_tax}
	\vspace{-2mm}
\end{table}

\begin{figure*}[t]
	\centering
	\includegraphics[width=0.9\textwidth]{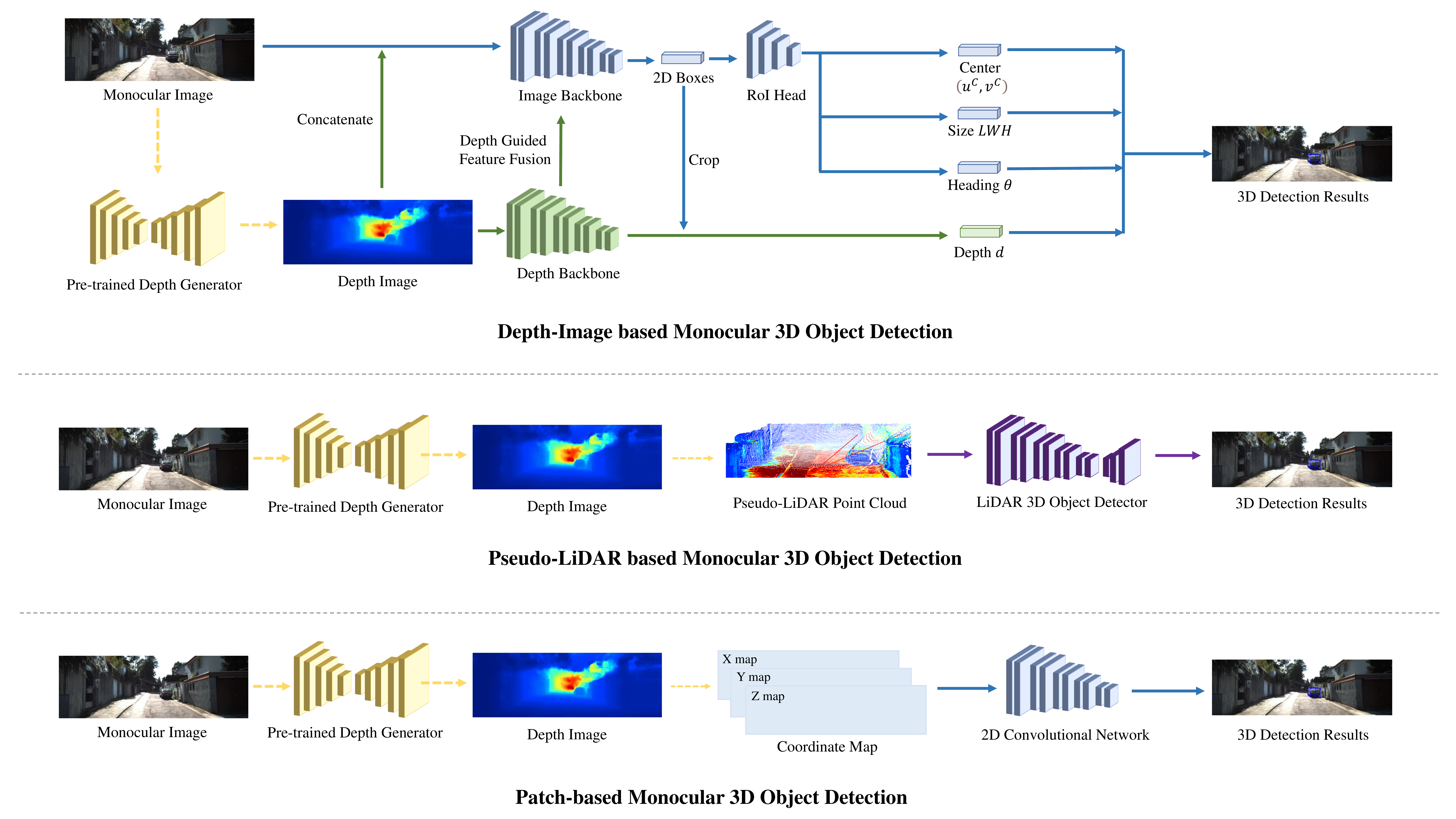}
	\caption{An illustration of depth-assisted monocular 3D object detection methods. Image and depth samples are from \cite{color-pslidar}.} 
	\label{fig11-depth-det}
	\vspace{-2mm}
\end{figure*}

Depth estimation is critical in monocular 3D object detection. To achieve more accurate monocular detection results, many papers resort to pre-training an auxiliary depth estimation network. Specifically, a monocular image is first passed through a pre-trained depth estimator, \textit{e.g.} MonoDepth~\cite{monodepth} or DORN~\cite{dorn}, to generate a depth image. Then, there are mainly two categories of methods to deal with depth images and monocular images. The depth-image based methods fuse images and depth maps with a specialized neural network to generate depth-aware features that could enhance the detection performance. The pseudo-LiDAR based methods convert a depth image into a pseudo-LiDAR point cloud, and LiDAR-based detectors can then be applied to the point cloud to predict 3D objects. An illustration of depth-assisted monocular 3D object detection is shown in Figure~\ref{fig11-depth-det} and a taxonomy of these methods is in Table~\ref{tab:mono_depth_tax}.

\noindent\textbf{Depth-image based methods.} Most depth-image based methods leverage two backbone networks for RGB and depth images respectively. They obtain depth-aware image features by fusing the information from the two backbones with specialized operators. More accurate 3D bounding boxes can be learned from the depth-ware features and can be further refined with depth images. MultiFusion~\cite{multifusion} is a pioneering work that introduces the depth-image based detection framework. Following papers adopt similar design paradigms with improvements in network architectures, operators, and training strategies, \textit{e.g.} a point-based attentional network~\cite{MonoFENet}, depth-guided convolutions~\cite{D4LCN}, depth-conditioned message passing~\cite{DDMP}, disentangling appearance and localization features~\cite{DFR-Net}, and a novel depth pre-training framework~\cite{DD3D}.

\noindent\textbf{Pseudo-LiDAR based methods.} Pseudo-LiDAR based methods transform a depth image into a pseudo-LiDAR point cloud, and LiDAR-based detectors can then be employed to detect 3D objects from the point cloud. Pseudo-LiDAR point cloud is a data representation first introduced in~\cite{pslidar}, where they convert a depth map $\mathcal{D} \in R^{H \times W}$ into a pseudo point cloud $\mathcal{P} \in R^{HW \times 3}$. Specifically, for each pixel $[u, v]$ and its depth value $d$ in a depth image, the corresponding 3D point coordinate $[x, y, z]$ in the camera coordinate system is computed as
\begin{equation}
	\begin{split}
		x = \frac{(u - c_{u}) \times z}{f_{u}},  y = \frac{(v - c_{v}) \times z}{f_{v}}, z = d,
	\end{split}
\end{equation}
where $[c_{u}, c_{v}]$ is the camera principal point, and $f_u$ and $f_v$ are the focal lengths along the horizontal and vertical axis respectively. Thus $\mathcal{P}$ can be obtained by back-projecting each pixel in $\mathcal{D}$ into the 3D space. $\mathcal{P}$ is referred as the pseudo-LiDAR representation: it is essentially a 3D point cloud but is extracted from a depth image instead of a real LiDAR sensor. Finally, LiDAR-based 3D object detectors can be directly applied on the pseudo-LiDAR point cloud $\mathcal{P}$ to predict 3D objects. Many papers have worked on improving the pseudo-LiDAR detection framework, including augmenting pseudo point cloud with color information~\cite{color-pslidar}, introducing instance segmentation~\cite{end-to-end-plidar}, designing a progressive coordinate transform scheme~\cite{PCT}, improving pixel-wise depth estimation with separate foreground and background prediction~\cite{ForeSeE}, domain adaptation from real LiDAR point cloud~\cite{da-3ddet}, and a new physical sensor design~\cite{deep-optics}.

PatchNet~\cite{PatchNet} challenges the conventional idea of leveraging the pseudo-LiDAR representation $\mathcal{P} \in R^{HW \times 3}$ for monocular 3D object detection. They conduct an in-depth investigation and provide an insightful observation that the power of pseudo-LiDAR representation comes from the coordinate transformation instead of the point cloud representation. Hence, a coordinate map $\mathcal{M} \in R^{H \times W \times 3}$ where each pixel encodes a 3D coordinate can attain a comparable monocular detection result with the pseudo-LiDAR point cloud representation. This observation enables us to directly apply a 2D neural network on the coordinate map to predict 3D objects, eliminating the need of leveraging the time-consuming LiDAR-based detectors on point clouds.

\noindent\textbf{Analysis: potentials and challenges of the depth-assisted approaches.} The depth-assisted approaches pursue more accurate depth estimation by leveraging a pre-trained depth prediction network. Both the depth image representation and the pseudo-LiDAR presentation could significantly boost the monocular detection performance. Nevertheless, compared to the image-only methods that only require 3D box annotations, pre-training a depth prediction network requires expensive ground truth depth maps, and it also hampers the end-to-end training of the whole framework. Furthermore, pre-trained depth estimation networks suffer from poor generalization ability. Pretrained depth maps are usually not well calibrated on the target dataset and typically the scale needs to be adapted to the target dataset. Thus there remains a non-negligible domain gap between the source domain leveraged for depth pre-training and the target domain for monocular detection. Given the fact that driving scenarios are normally diverse and complex, pre-training depth networks on a restricted domain may not work well in real-world applications.

\subsubsection{Prior-guided monocular 3D object detection}

\begin{figure}[t]
	\centering
	\includegraphics[width=0.48\textwidth]{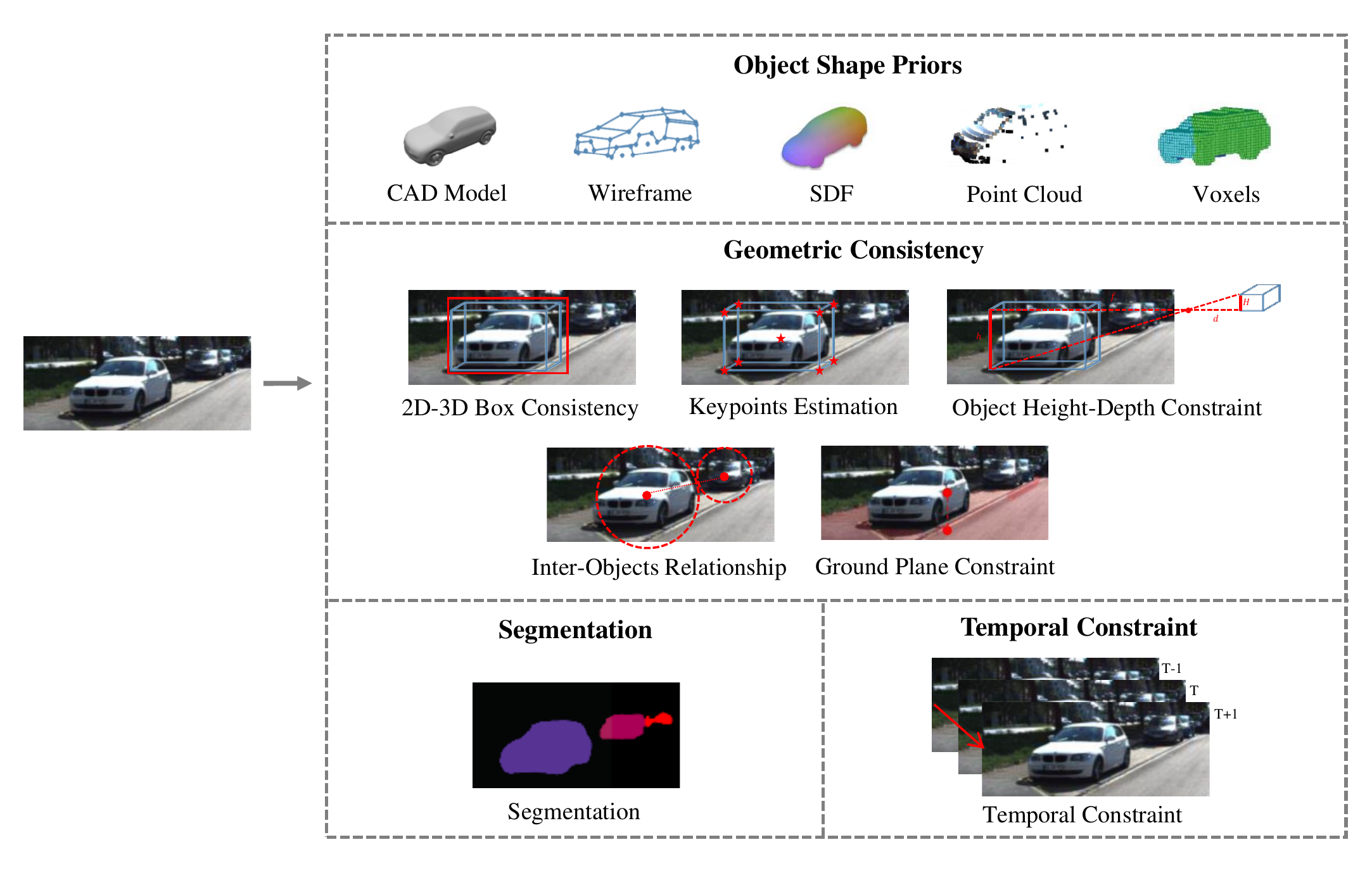}
	\caption{An illustration of the prior types in monocular 3D object detection methods. Samples are from \cite{mono3d, mono3d++, 3dvp, MonoDR, deepsdf}.}
	\label{fig12-prior-det}
\end{figure}

Numerous approaches try to tackle the ill-posed monocular 3D object detection problem by leveraging the hidden prior knowledge of object shapes and scene geometry from images. The prior knowledge can be learned by introducing pre-trained sub-networks or auxiliary tasks, and they can provide extra information or constraints to help accurately localize 3D objects. The broadly adopted prior knowledge includes object shapes, geometry consistency, temporal constraints, and segmentation information. An illustration of the prior types is shown in Figure~\ref{fig12-prior-det}.     

\noindent\textbf{Object shapes.} Many methods resort to shape reconstruction of 3D objects directly from images. The reconstructed shapes can be further leveraged to determine the locations and poses of the 3D objects. There are 5 types of reconstructed representations: computer-aided design (CAD) models, wireframe models, signed distance function (SDF), points, and voxels. 

Some papers \cite{mono2014, deepmanta, mono3d++} learn morphable wireframe models to represent 3D objects. Other works \cite{3drcnn, ROI-10D, SDFLabel, MonoDR} leverage DeepSDF~\cite{deepsdf} to learn implicit signed distance functions or low-dimensional shape parameters from CAD models, and they further propose a render-and-compare approach to learn the parameters of 3D objects. Some works~\cite{3dvp, subcnn} utilize voxel patterns to represent 3D objects. Other papers~\cite{MonoPSR, MonoRUn} resort to point cloud reconstruction from images and estimate the locations of 3D objects with 2D-3D correspondences. 

\noindent\textbf{Geometric consistency.} Given the extrinsics matrix $T \in SE(3)$ that transforms a 3D coordinate in the object frame to the camera frame, and the camera intrinsics matrix $K$ that project the 3D coordinate onto the image plane, the projection of a 3D point $[x, y, z]$ in the object frame into the image pixel coordinate $[u, v]$ can be represented as
\begin{equation} \label{eq:projection}
	d \cdot \begin{bmatrix}
		u \\ v \\ 1
	\end{bmatrix} = K \ T \ \begin{bmatrix}
		x \\ y \\ z \\ 1
	\end{bmatrix},
\end{equation}
where $d$ is the depth of transformed 3D coordinate in the camera frame. Eqn.~\ref{eq:projection} provides a geometric relationship between 3D points and 2D image pixel coordinates, which can be leveraged in various ways to encourage consistency between the predicted 3D objects and the 2D objects on images. There are mainly $5$ types of geometric constraints in monocular detection: 2D-3D boxes consistency, keypoints, object's height-depth relationship, inter-objects relationship, and ground plane constraints.

Some works~\cite{deep3dbox, m3d-rpn, ss3d, shiftrcnn} propose to encourage the consistency between 2D and 3D boxes by minimizing reprojection errors. These methods introduce a post-processing step to optimize the 3D object parameters by gradually fitting the projected 3D boxes to 2D bounding boxes on images. There is also a branch of papers~\cite{RTM3D, UR3D, KM3D-Net} that predict the object keypoints from images, and the keypoints can be leveraged to calibrate the sizes of locations of 3D objects. Object's height-depth relationship can also serve as a strong geometric prior. Specifically, given the physical height of an object $H$ in the 3D space, the visual height $h$ on images, and the corresponding depth of the object $d$, there exists a geometric constraint: $d = f \cdot H/h$, where $f$ is the camera focal length. This constraint can be leveraged to obtain more accurate depth estimation and has been broadly applied in a lot of works~\cite{cai-mono, MonoFlex, GUPNet, MonoRCNN}. There are also some papers~\cite{MoNet3D, MonoPair} trying to model the inter-objects relationships by exploiting new geometric relations among objects. Other papers~\cite{mono3d, m3d-rpn, FQNet, liu-mono} leverage the assumption that 3D objects are generally on the ground plane to better localize those objects.

\begin{table}[t!]
	\caption{A taxonomy of prior-guided monocular detection methods based on prior types.}
	\centering
	\begin{tabular}{|c|c|c|}
		\hline
		\multicolumn{2}{|c|}{Prior Types} & Methods \\
		\hline
		\hline
		\multirow{4}*{Object shape} & wireframe &\tabincell{c}{\cite{mono2014, deepmanta, mono3d++}} \\
		\cline{2-3}
		& SDF &\tabincell{c}{\cite{3drcnn, ROI-10D, MonoDR, SDFLabel}} \\
		\cline{2-3}
		& points &\tabincell{c}{\cite{MonoRUn, MonoPSR}} \\
		\cline{2-3}
		& voxels &\tabincell{c}{\cite{3dvp, subcnn}} \\
		\hline
		\multirow{5}*{Geometric consistency} & 2D-3D boxes &\tabincell{c}{\cite{deep3dbox, m3d-rpn, ss3d, shiftrcnn}} \\
		\cline{2-3}
		& keypoints &\tabincell{c}{\cite{RTM3D, KM3D-Net, UR3D}} \\
		\cline{2-3}
		& height-depth &\tabincell{c}{\cite{cai-mono, GUPNet, MonoFlex, MonoRCNN}} \\
		\cline{2-3}
		& inter-objects &\tabincell{c}{\cite{MoNet3D, MonoPair}} \\
		\cline{2-3}
		& ground plane &\tabincell{c}{\cite{mono3d, liu-mono, m3d-rpn, FQNet}} \\
		\hline
		\multicolumn{2}{|c|}{Temporal constraints} &\tabincell{c}{\cite{mono-video, Kinematic3D}} \\
		\hline
		\multicolumn{2}{|c|}{Segmentation} &\tabincell{c}{\cite{3dvp, MonoDR, mono3d, MonoCInIS}} \\
		\hline
	\end{tabular}
	\label{tab:prior_tax}
\end{table}

\noindent\textbf{Temporal constraints.} Temporal association of 3D objects can be leveraged as strong prior knowledge. The temporal object relationships have been exploited as depth-ordering~\cite{mono-video} and multi-frame object fusion with a 3D Kalman filter~\cite{Kinematic3D}.

\noindent\textbf{Segmentation} Image segmentation helps monocular 3D object detection mainly in two aspects. First, object segmentation masks are crucial for instance shape reconstruction in some works~\cite{3dvp, MonoDR}. Second, segmentation indicates whether an image pixel is inside a 3D object from the perspective view, and this information has been utilized in~\cite{mono3d, MonoCInIS} to help localize 3D objects.

\noindent\textbf{Analysis: potentials and challenges of leveraging prior knowledge in monocular 3D detection.} With shape reconstruction, we could obtain more detailed object shape information from images, which is beneficial to 3D object detection. We can also attain more accurate detection results through the projection or render-and-compare loss. However, there exist two challenges for shape reconstruction applied in monocular 3D object detection. First, shape reconstruction normally requires an additional step of pre-training a reconstruction network, which hampers end-to-end training of the monocular detection pipeline. Second, object shapes are generally learned from CAD models instead of real-world instances, which imposes the challenge of generalizing the reconstructed objects to real-world scenarios. 

Geometric consistencies are broadly adopted and can help improve detection accuracy. Nevertheless, some methods formulate the geometric consistency as an optimization problem and optimize object parameters in post-processing, which is quite time-consuming and hampers end-to-end training. 

Image segmentation is useful information in monocular 3D detection. However, training segmentation networks requires expensive pixel annotations. Pre-training segmentation models on external datasets will suffer from the generalization problem.

\subsection{Stereo-based 3D object detection} \label{sec:stereo}

\begin{figure*}[t]
	\centering
	\includegraphics[width=0.9\textwidth]{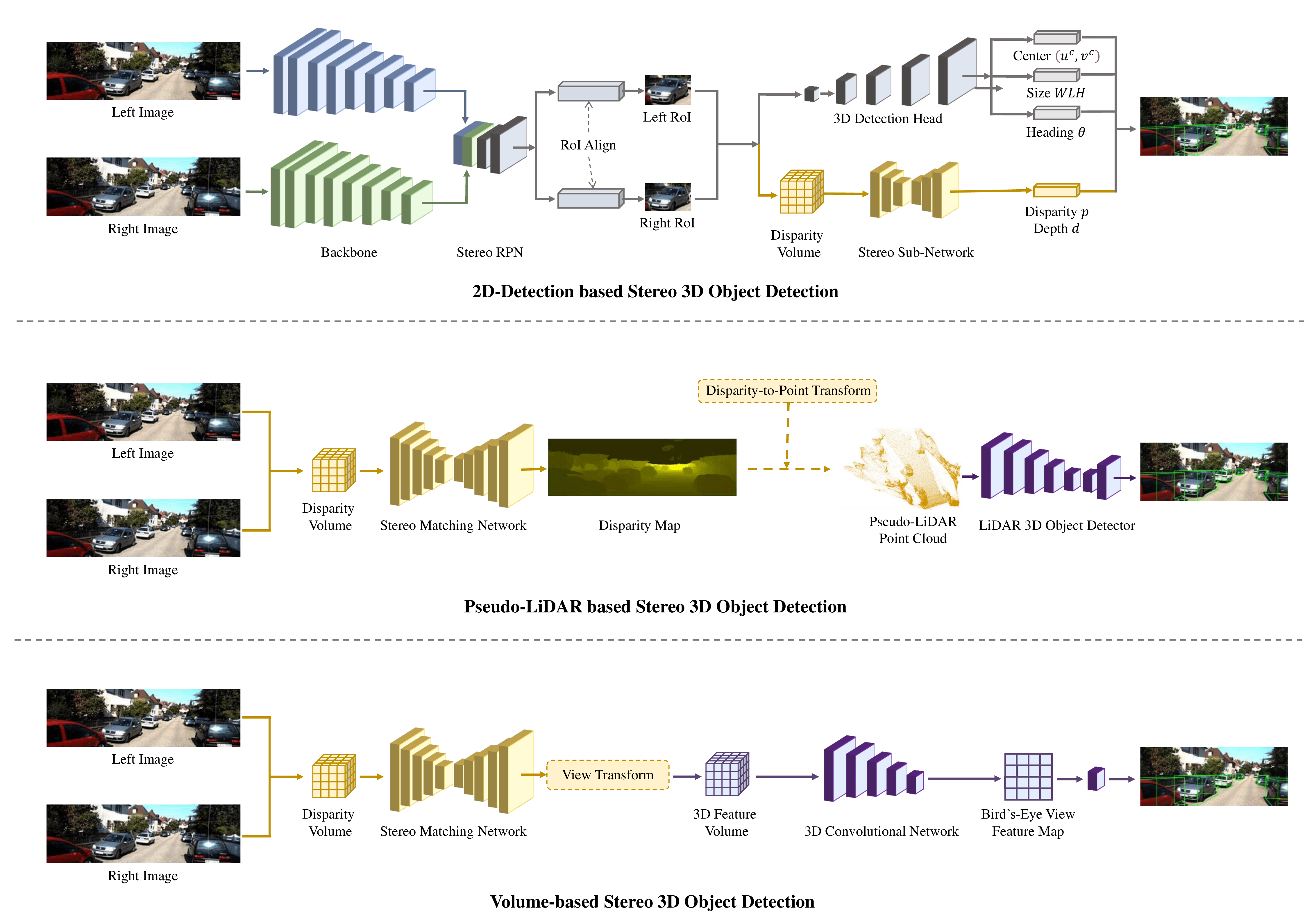}
	\caption{An illustration of stereo-based 3D object detection methods. Image and disparity samples are from \cite{end-to-end-pseudo-lidar}.} 
	\label{fig13-stereo-det}
	\vspace{-2mm}
\end{figure*}

\noindent\textbf{Problem and Challenge.} Stereo-based 3D object detection aims to detect 3D objects from a pair of images. Compared to monocular images, paired stereo images provide additional geometric constraints that can be utilized to infer more accurate depth information. Hence, the stereo-based methods generally obtain a better detection performance than the monocular-based methods. Nevertheless, stereo cameras typically require very accurate calibration and synchronization, which are normally difficult to achieve in real applications. An illustration of stereo-based 3D object detection approaches is shown in Figure~\ref{fig13-stereo-det} and a taxonomy is in Table~\ref{tab:stereo_tax}.  

\noindent\textbf{Stereo matching and depth estimation.} A stereo camera can produce a pair of images, \textit{i.e.} the left image $\mathcal{I}_{L}$ and the right image $\mathcal{I}_{R}$, in one shot. With the stereo matching techniques~\cite{Dispnet, PSMNet}, a disparity map can be estimated from the paired stereo images leveraging multi-view geometry~\cite{mvg}. Ideally, for each pixel on the left image $\mathcal{I}_{L}(u, v)$, there exists a pixel on the right image $\mathcal{I}_{R}(u, v+p)$ with the disparity value $p$ so that the two pixels picture the same 3D location. Finally, the disparity map can be transformed into a depth image with the following formula:
\begin{equation}
	d = \frac{f \times b}{p},
\end{equation}
where $d$ is the depth value, $f$ is the focal length, and $b$ is the baseline length of the stereo camera. The pixel-wise disparity constraints from stereo images enable more accurate depth estimation compared to monocular depth prediction.

\begin{table}[t!]
	\caption{A taxonomy of stereo-based detection methods based on auxiliary tasks and data representations.}
	\centering
	\begin{tabular}{|c|c|c|c|c|c|}
		\hline
		Method &  \tabincell{c}{2D \\ Det.} &  \tabincell{c}{2D \\ Seg.} & \tabincell{c}{Disp./ \\ Depth} &  \tabincell{c}{Pseudo \\ LiDAR} & \tabincell{c}{3D \\ Volume} \\
		\hline
		\hline
		3DOP~\cite{3dop-conf} & & & \checkmark & &\\
		\hline
		TLNet~\cite{TL-Net} & \checkmark & & & & \\
		\hline
		Stereo R-CNN~\cite{stereo-rcnn} & \checkmark & & & & \\
		\hline
		Disp R-CNN~\cite{disp-rcnn} & \checkmark & \checkmark & \checkmark & & \\
		\hline
		ZoomNet~\cite{zoomnet} & \checkmark & \checkmark & \checkmark & & \\
		\hline
		OC-Stereo~\cite{oc-stereo} & \checkmark & \checkmark & \checkmark & & \\
		\hline
		IDA-3D~\cite{IDA-3D} & \checkmark &  & \checkmark & & \\
		\hline
		YOLOStereo3D~\cite{yolostereo3d} & & & \checkmark & & \\
		\hline
		SIDE~\cite{side} & \checkmark & & \checkmark & & \\
		\hline
		P-LiDAR++~\cite{Pseudo-LiDAR++} & & & \checkmark & \checkmark & \\
		\hline
		Qian \textit{et al.}~\cite{end-to-end-pseudo-lidar} & & & \checkmark & \checkmark & \\
		\hline
		CDN~\cite{wasserstein-stereo} & & & \checkmark & \checkmark & \\
		\hline
		RT3D-Stereo~\cite{RT-Stereo} & & \checkmark & \checkmark & \checkmark & \\
		\hline
		CG-Stereo~\cite{CG-Stereo} & & \checkmark & \checkmark & \checkmark & \\
		\hline
		LIGA-Stereo~\cite{LIGA-Stereo} & & & \checkmark & & \checkmark \\
		\hline
		DSGN~\cite{DSGN} & & & \checkmark & & \checkmark \\
		\hline
		PLUMENet~\cite{PLUMENet} & & & \checkmark & & \checkmark \\
		\hline
	\end{tabular}
	\label{tab:stereo_tax}
	\vspace{-2mm}
\end{table}

\noindent\textbf{2D-detection based methods.} Conventional 2D object detection frameworks can be modified to resolve the stereo detection problem. Specifically, paired stereo images are passed through an image-based detector with Siamese backbone networks to generate left and right regions of interest (RoIs) for the left and right images respectively. Then in the second stage, the left and right RoIs are fused to estimate the parameters of 3D objects. Stereo R-CNN~\cite{stereo-rcnn} first proposes to extend 2D detection frameworks to stereo 3D detection. This design paradigm has been adopted in numerous papers. \cite{TL-Net} proposes a novel stereo triangulation learning sub-network at the second stage; \cite{zoomnet, oc-stereo, disp-rcnn, disp-rcnn-journal} learn instance-level disparity by object-centric stereo matching and instance segmentation;  \cite{IDA-3D} proposes adaptive instance disparity estimation; \cite{yolostereo3d, side} introduce single-stage stereo detection frameworks; \cite{3dop-conf, 3dop-journal} propose an energy-based framework for stereo-based 3D object detection.

\begin{figure*}[t]
	\centering
	\includegraphics[width=0.9\textwidth]{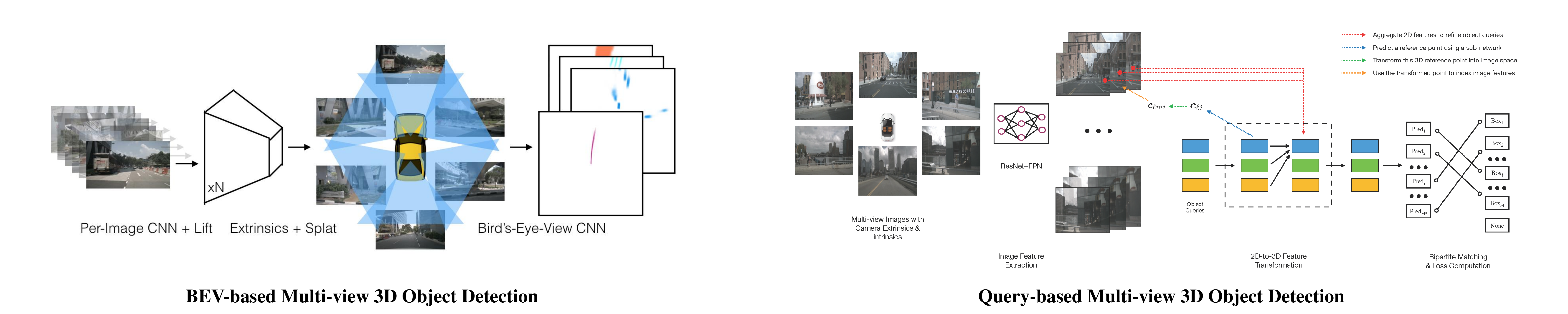}
	\caption{An illustration of multi-view 3D object detection methods. Figures are from~\cite{LSS} and \cite{detr3d}.}
	\label{fig-multiview}
	\vspace{-2mm}
\end{figure*}

\noindent\textbf{Pseudo-LiDAR based methods.} The disparity map predicted from stereo images can be transformed into the depth image and then converted into the pseudo-LiDAR point cloud. Hence, similar to the monocular detection methods, the pseudo-LiDAR representation can also be employed in stereo-based 3D object detection methods. Those methods try to improve the disparity estimation in stereo matching for more accurate depth prediction. \cite{Pseudo-LiDAR++} introduces a depth cost volume in stereo matching networks; \cite{end-to-end-pseudo-lidar} proposes an end-to-end stereo matching and detection framework; \cite{RT-Stereo, CG-Stereo} leverage semantic segmentation and predict disparity for foreground and background regions separately; \cite{wasserstein-stereo} proposes a Wasserstein loss for disparity estimation.

\noindent\textbf{Volume-based methods.} There exists a category of methods that skip the pseudo-LiDAR representation and perform 3D object detection directly on 3D stereo volumes. DSGN~\cite{DSGN} proposes a 3D geometric volume derived from stereo matching networks and applies a grid-based 3D detector on the volume to detect 3D objects. \cite{LIGA-Stereo} and \cite{PLUMENet} improve \cite{DSGN} by leveraging knowledge distillation and 3D feature volumes respectively.

\noindent\textbf{Potentials and challenges of the stereo-based methods.} Compared to the monocular detection methods, the stereo-based methods can obtain more accurate depth and disparity estimation with stereo matching techniques, which brings a stronger object localization ability and significantly boosts the 3D object detection performance. Nevertheless, an auxiliary stereo matching network brings additional time and memory consumption. Compared to LiDAR-based 3D object detection, detection from stereo images can serve as a much cheaper solution for 3D perception in autonomous driving scenarios. However, there still exists a non-negligible performance gap between the stereo-based and the LiDAR-based 3D object detection approaches. 

\subsection{Multi-view 3D object detection} \label{sec:multi-camera}

\noindent\textbf{Problem and Challenge.} Autonomous vehicles are generally equipped with multiple cameras to obtain complete environmental information from multiple viewpoints. Recently, multi-view 3D object detection has evolved rapidly. Some multi-view 3D detection approaches try to construct a unified BEV space by projecting multi-view images into the bird's-eye view, and then employ a BEV-based detector on top of the unified BEV feature map to detect 3D objects. The transformation from camera views to the bird's-eye view is ambiguous without accurate depth information, so image pixels and their BEV locations are not perfectly aligned. How to build reliable transformations from camera views to the bird's-eye view is a major challenge in these methods. Other methods resort to 3D object queries that are generated from the bird's-eye view and Transformers where cross-view attention is applied to object queries and multi-view image features. The major challenge is how to properly generate 3D object queries and design more effective attention mechanisms in Transformers.

\noindent\textbf{BEV-based multi-view 3D object detection.} LSS~\cite{LSS} is a pioneering work that proposes a lift-splat-shoot paradigm to solve the problem of BEV perception from multi-view cameras. There are three steps in LSS. \textit{Lift:} bin-based depth prediction is conducted on image pixels and multi-view image features are lifted to 3D frustums with depth bins. \textit{Splat:} 3D frustums are splatted into a unified bird's-eye view plane and image features are transformed into BEV features in an end-to-end manner. \textit{Shoot:} downstream perception tasks are performed on top of the BEV feature map. This paradigm has been successfully adopted by many following works. BEVDet~\cite{bevdet, bevdet4d} improves LSS~\cite{LSS} with a four-step multi-view detection pipeline, where the image view encoder encodes features from multi-view images, the view transformer transforms image features from camera views to the bird's-eye view, the BEV encoder further encodes the BEV features, and the detection head is employed on top of the BEV features for 3D detection. The major bottleneck in \cite{bevdet, LSS} is depth prediction, as it is normally inaccurate and will result in inaccurate feature transforms from camera views to the bird's-eye view. To obtain more accurate depth information, many papers resort to mining additional information from multi-view images and past frames, \textit{e.g.} \cite{bevdepth} leverages explicit depth supervision, \cite{sts} introduces surround-view temporal stereo, \cite{bevstereo} uses dynamic temporal stereo, \cite{solofusion} combines both short-term and long-term temporal stereo for depth prediction. In addition, there are also some papers~\cite{m2bev, fastbev} that completely abandon the design of depth bins and categorical depth prediction. They simply assume that the depth distribution along the ray is uniform, so the camera-to-BEV transformation can be conducted with higher efficiency.  

\noindent\textbf{Query-based multi-view 3D object detection.} In addition to the BEV-based approaches, there is also a category of methods where object queries are generated from the bird's-eye view and interact with camera view features. Inspired by the advances in Transformers for object detection~\cite{detr}, DETR3D~\cite{detr3d} introduces a sparse set of 3D object queries, and each query corresponds to a 3D reference point. The 3D reference points can collect image features by projecting their 3D locations onto the multi-view image planes and then object queries interact with image features through Transformer layers. Finally, each object query will decode a 3D bounding box. Many following papers try to improve this design paradigm, such as introducing spatially-aware cross-view attention~\cite{spatialdetr} and adding 3D positional embeddings on top of image features~\cite{petr}. BEVFormer~\cite{bevformer} introduces dense grid-based BEV queries and each query corresponds to a pillar that contains a set of 3D reference points. Spatial cross-attention is applied to object queries and sparse image features to obtain spatial information, and temporal self-attention is applied to object queries and past BEV queries to fuse temporal information.


\begin{figure*}[t]
	\centering
	\includegraphics[width=\textwidth]{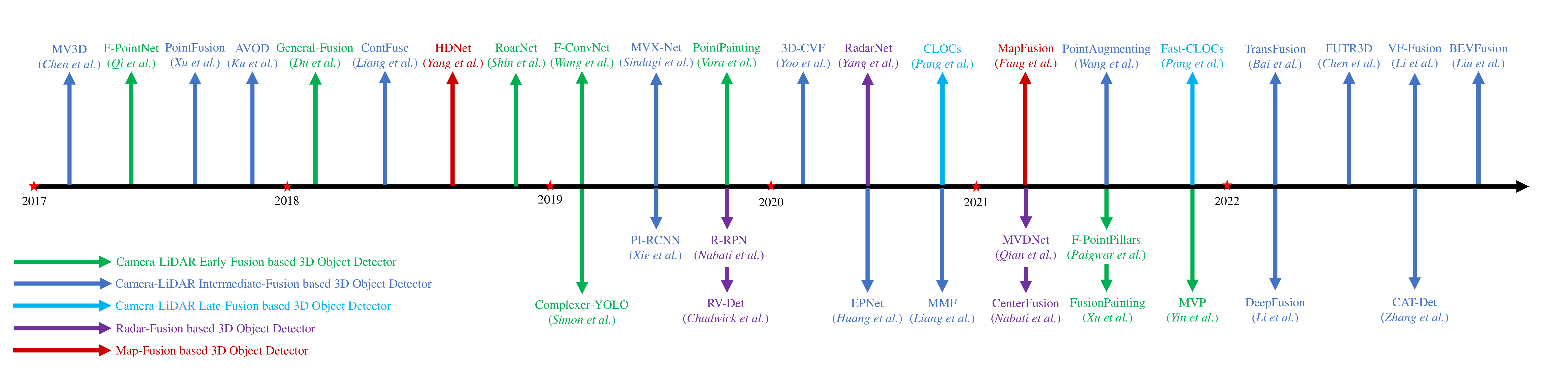}
	\caption{Chronological overview of the multi-modal 3D object detection methods.}
	\label{fig14-fusion-det}
	\vspace{-2mm}
\end{figure*}

\begin{table*}[t!]
	\caption{A taxonomy of multi-sensor fusion-based detection methods based on fused stages, representations, and operators.}
	\centering
	\begin{tabular}{|c|c|c|c|c|c|c|c|c|}
		\hline
		\multirow{2}*{Method} & \multicolumn{5}{c|}{Fusion Stage} & \multicolumn{2}{c|}{Fusion Representation} & \multirow{2}*{\tabincell{c}{Fusion \\ Operator}} \\
		\cline{2-8}
		&  \tabincell{c}{Input} &  \tabincell{c}{Backbone} & \tabincell{c}{Proposal} &  \tabincell{c}{RoI} & \tabincell{c}{Output} & Camera rep. & LiDAR rep. & \\
		\hline
		\hline
		F-PointNet~\cite{fpointnet18} & \checkmark & & & & & frustum & point cloud & region selection\\
		\hline
		F-ConvNet~\cite{F-ConvNet} & \checkmark & & & & & frustum & point cloud & region selection\\
		\hline
		RoarNet~\cite{roarnet} & \checkmark & & & & & 2D boxes \& poses & point cloud & region selection\\
		\hline
		PointPainting~\cite{pointpainting} & \checkmark & & & & & 2D segmentation & point cloud & point-wise append\\
		\hline
		MVX-Net~\cite{mvx-net} & & \checkmark & & & & image features & voxels & concatenation \& MLP \\
		\hline
		ContFuse~\cite{contfuse} & & \checkmark & & & & image features & BEV features & continuous convolution\\
		\hline
		PointFusion~\cite{pointfusion} & & \checkmark & & & & image features & point features & concatenation \& MLP\\
		\hline
		EPNet~\cite{epnet} & & \checkmark & & & & image features & point features & point-wise attention \\
		\hline
		MMF~\cite{mmf} & & \checkmark & \checkmark & \checkmark & & image features & BEV features & continuous convolution\\
		\hline
		3D-CVF~\cite{3d-cvf} & & \checkmark & \checkmark & \checkmark & & image features & BEV features & gated attention\\
		\hline
		MV3D~\cite{mv3d} & & & \checkmark & \checkmark & & image features & multi-view features & concatenation \& MLP\\
		\hline
		AVOD~\cite{avod} & & & \checkmark & \checkmark & & image features & BEV features & concatenation \& MLP\\
		\hline
		CLOCs~\cite{clocs} & & & & & \checkmark & 2D boxes & 3D boxes & box consistency\\
		\hline
	\end{tabular}
	\label{tab:fusion_tax}
	\vspace{-2mm}
\end{table*}

\section{Multi-Modal 3D Object Detection} \label{sec:multisensor}

In this section, we introduce the multi-modal 3D object detection approaches that fuse multiple sensory inputs. According to the sensor types, the approaches can be divided into three categories: LiDAR-camera, radar, and map fusion-based methods. In Section~\ref{sec:camera-lidar}, we review and analyze the multi-modal detection approaches with LiDAR-camera fusion, including the early-fusion based, the intermediate-fusion based, and the late-fusion based methods. In Section~\ref{sec:radar}, we investigate the multi-modal detection approaches with radar signals. In Section~\ref{sec:map}, we introduce the multi-modal 3D detection approaches with high-definition maps. A chronological overview of the multi-modal 3D object detection approaches is shown in Figure~\ref{fig14-fusion-det}.

\subsection{Multi-modal detection with LiDAR-camera fusion} \label{sec:camera-lidar}

\noindent\textbf{Problem and Challenge.} Camera and LiDAR are two complementary sensor types for 3D object detection. Cameras provide color information from which rich semantic features can be extracted, while LiDAR sensors specialize in 3D localization and provide rich information about 3D structures. Many endeavors have been made to fuse the information from cameras and LiDARs for accurate 3D object detection. Since LiDAR-based detection methods perform much better than camera-based methods, the state-of-the-art approaches are mainly based on LiDAR-based 3D object detectors and try to incorporate image information into different stages of a LiDAR detection pipeline. In view of the complexity of LiDAR-based and camera-based detection systems, combining the two modalities together inevitably brings additional computational overhead and inference time latency. Therefore, how to efficiently fuse the multi-modal information remains an open challenge. A taxonomy of multi-modal 3D object detection methods is in Table~\ref{tab:fusion_tax}.

\subsubsection{Early-fusion based 3D object detection}

\begin{figure*}[t]
	\centering
	\includegraphics[width=0.9\textwidth]{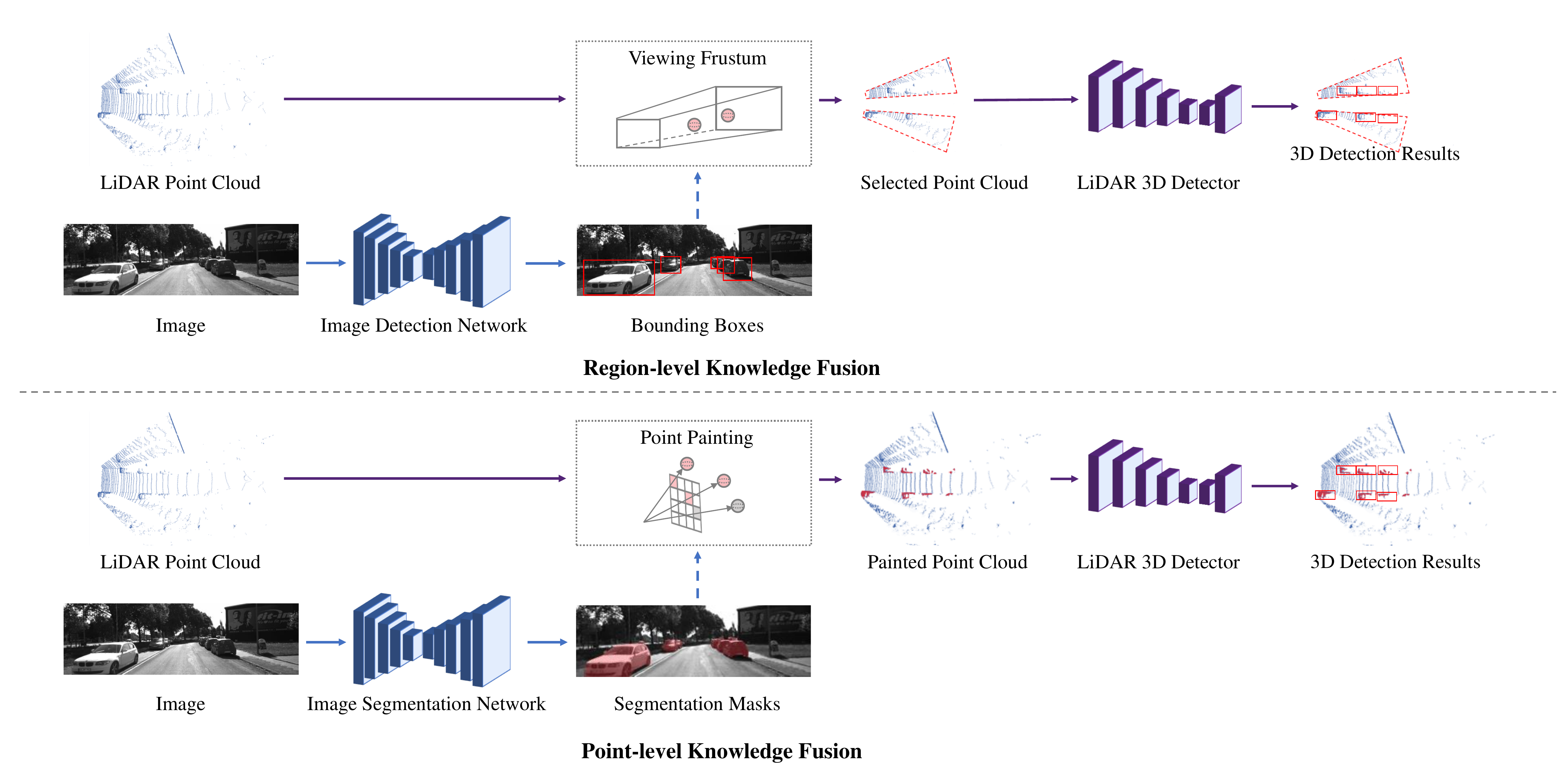}
	\caption{An illustration of early-fusion based 3D object detection methods.} 
	\label{fig15-early-fusion}
	\vspace{-2mm}
\end{figure*}

Early-fusion based methods aim to incorporate the knowledge from images into point cloud before they are fed into a LiDAR-based detection pipeline. Hence the early-fusion frameworks are generally built in a sequential manner: 2D detection or segmentation networks are firstly employed to extract knowledge from images, and then the image knowledge is passed to point cloud, and finally the enhanced point cloud is fed to a LiDAR-based 3D object detector. Based on the fusion types, the early-fusion methods can be divided into two categories: region-level knowledge fusion and point-level knowledge fusion. An illustration of the early-fusion based approaches is shown in Figure~\ref{fig15-early-fusion}.

\noindent\textbf{Region-level knowledge fusion.} Region-level fusion methods aim to leverage knowledge from images to narrow down the object candidate regions in 3D point cloud. Specifically, an image is first passed through a 2D object detector to generate 2D bounding boxes, and then the 2D boxes are extruded into 3D viewing frustums. The 3D viewing frustums are applied on LiDAR point cloud to reduce the searching space. Finally, only the selected point cloud regions are fed into a LiDAR detector for 3D object detection. F-PointNet~\cite{fpointnet18} first proposes this fusion mechanism, and many endeavors have been made to improve the fusion framework. \cite{F-ConvNet} divides a viewing frustum into grid cells and applies a convolutional network on the grid cells for 3D detection; \cite{roarnet} proposes a novel geometric agreement search; \cite{frustum-pointpillars} exploits the pillar representation; \cite{du-multimodal} introduces a model fitting algorithm to find the object point cloud inside each frustum.

\begin{figure*}[t]
	\centering
	\includegraphics[width=0.9\textwidth]{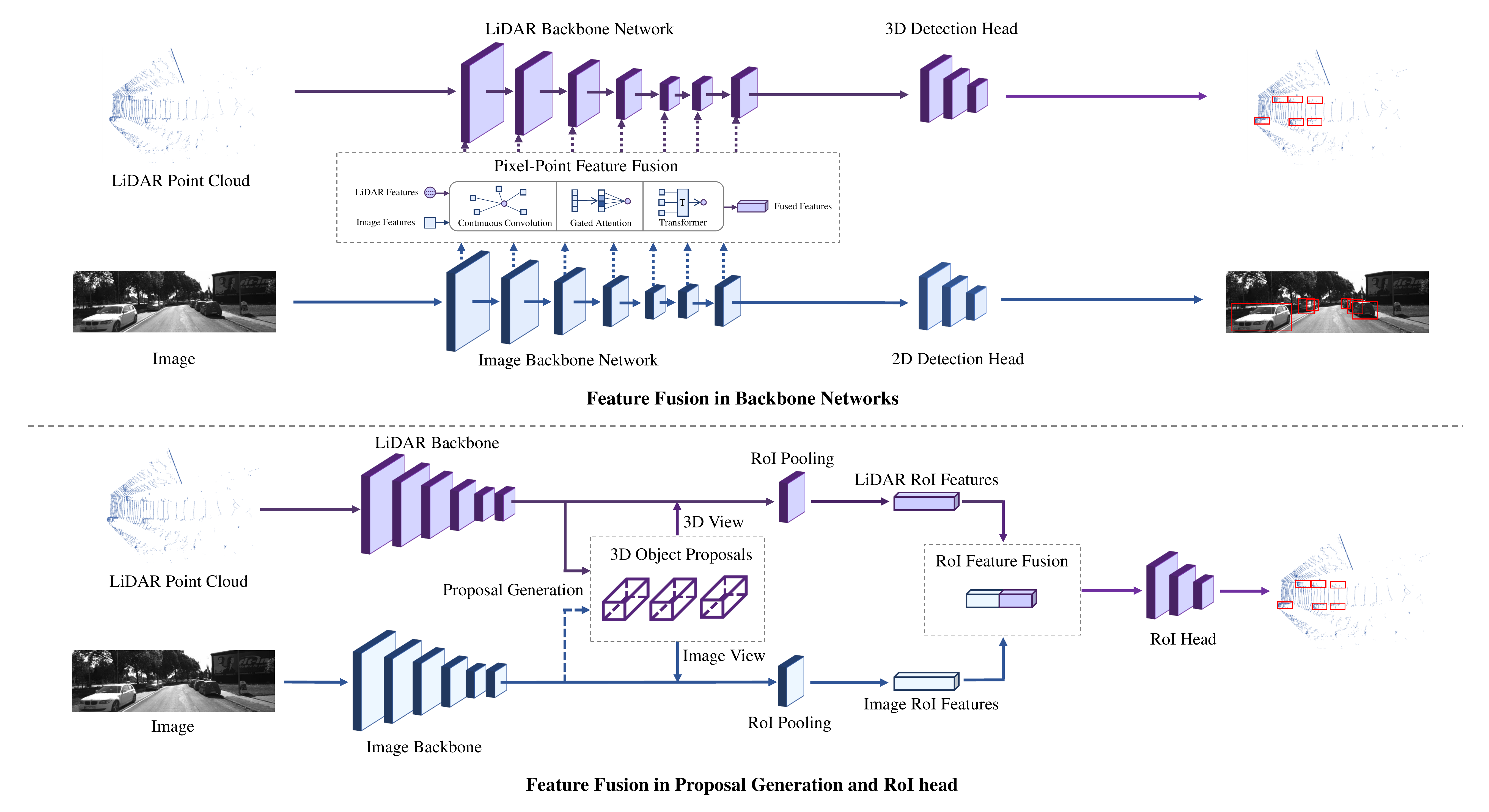}
	\caption{An illustration of intermediate-fusion based 3D object detection methods.} 
	\label{fig16-inter-fusion}
	\vspace{-3mm}
\end{figure*}

\begin{figure*}[t]
	\centering
	\includegraphics[width=0.9\textwidth]{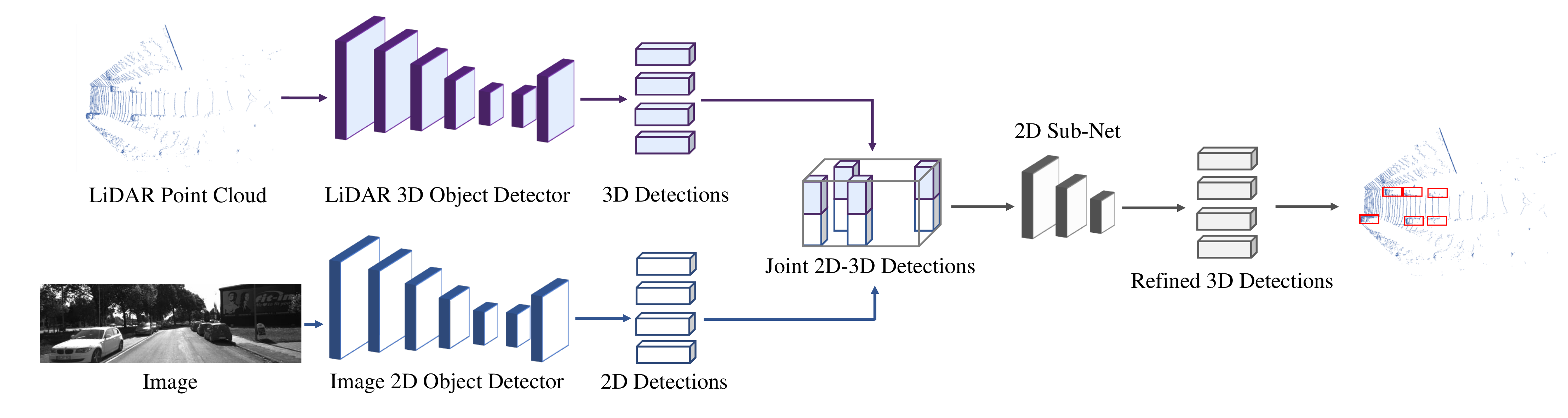}
	\caption{An illustration of late-fusion based 3D object detection methods.} 
	\label{fig17-late-fusion}
	\vspace{-2mm}
\end{figure*}

\noindent\textbf{Point-level knowledge fusion.} Point-level fusion methods aim to augment input point cloud with image features. The augmented point cloud is then fed into a LiDAR detector to attain a better detection result. PointPainting~\cite{pointpainting} is a seminal work that leverages image-based semantic segmentation to augment point clouds. Specifically, an image is passed through a segmentation network to obtain pixel-wise semantic labels, and then the semantic labels are attached to the 3D points by point-to-pixel projection. Finally, the points with semantic labels are fed into a LiDAR-based 3D object detector. This design paradigm has been followed by a lot of papers~\cite{fusionpainting, complexer-yolo, fusion-lasernet}. Apart from semantic segmentation, there also exist some works trying to exploit other information from images, \textit{e.g.} depth image completion~\cite{mvp}.

\noindent\textbf{Analysis: potentials and challenges of the early-fusion methods.} The early-fusion based methods focus on augmenting point clouds with image information before they are passed through a LiDAR 3D object detection pipeline. Most methods are compatible with a wide range of LiDAR-based 3D object detectors and can serve as a quite effective pre-processing step to boost detection performance. Nevertheless, the early-fusion methods generally perform multi-modal fusion and 3D object detection in a sequential manner, which brings additional inference latency. Given the fact that the fusion step generally requires a complicated 2D object detection or semantic segmentation network, the time cost brought by multi-modal fusion is normally non-negligible. Hence, how to perform multi-modal fusion efficiently at the early stage has become a critical challenge. 

\subsubsection{Intermediate-fusion based 3D object detection}

Intermediate-fusion based methods try to fuse image and LiDAR features at the intermediate stages of a LiDAR-based 3D object detector, \textit{e.g.} in backbone networks, at the proposal generation stage, or at the RoI refinement stage. These methods can also be classified according to the fusion stages. An illustration of intermediate-fusion based approaches is shown in Figure~\ref{fig16-inter-fusion}.

\noindent\textbf{Fusion in backbone networks.} Many endeavors have been made to progressively fuse image and LiDAR features in the backbone networks. In those methods, point-to-pixel correspondences are firstly established by LiDAR-to-camera transform, and then with the point-to-pixel correspondences, features from a LiDAR backbone can be fused with features from an image backbone through different fusion operators. The multi-modal fusion can be conducted in the intermediate layers of a grid-based detection backbone, with novel fusion operators such as continuous convolutions~\cite{continuous-conv, contfuse, mmf}, hybrid voxel feature encoding~\cite{mvx-net}, and Transformer~\cite{deepfusion, cat-det}. The multi-modal fusion can also be conducted only at the output feature maps of backbone networks, with fusion modules and operators including gated attention~\cite{3d-cvf}, unified object queries~\cite{futr3d}, BEV pooling~\cite{bevfusion}, learnable alignments~\cite{autoalign}, point-to-ray fusion~\cite{vff}, Transformer~\cite{transfusion}, and other techniques~\cite{fusion-seg-voxelnet, boost-multimodal, pointaugmenting}. In addition to the fusion in grid-based backbones, there also exist some papers incorporating image information into the point-based detection backbones~\cite{pointfusion, epnet, pi-rcnn, fusion-iccv21w, cross-modal-wacv}.

\noindent\textbf{Fusion in proposal generation and RoI head.} There exists a category of works that conduct multi-modal feature fusion at the proposal generation and RoI refinement stage. In those methods, 3D object proposals are first generated from a LiDAR detector, and then the 3D proposals are projected into multiple views, \textit{i.e.} the image view and bird's-eye view, to crop features from the image and LiDAR backbone respectively. Finally, the cropped image and LiDAR features are fused in an RoI head to predict parameters for each 3D object. MV3D~\cite{mv3d} and AVOD~\cite{avod} are pioneering works leveraging multi-view aggregation for multi-modal detection. Other papers~\cite{futr3d, transfusion} use the Transformer~\cite{transformer} decoder as the RoI head for multi-modal feature fusion. 

\noindent\textbf{Analysis: potentials and challenges of the intermediate-fusion methods.} The intermediate methods encourage deeper integration of multi-modal representations and yield 3D boxes of higher quality. Nevertheless, camera and LiDAR features are intrinsically heterogeneous and come from different viewpoints, so there still exist some problems on the fusion mechanisms and view alignments. Hence, how to fuse the heterogeneous data effectively and how to deal with the feature aggregation from multiple views remain a challenge to the research community. 

\subsubsection{Late-fusion based 3D object detection}

\noindent\textbf{Fusion at the box level.} Late-fusion based approaches operate on the outputs, \textit{i.e.} 3D and 2D bounding boxes, from a LiDAR-based 3D object detector and an image-based 2D object detector respectively. An illustration of late-fusion based approaches is shown in Figure~\ref{fig17-late-fusion}. In those methods, object detection with camera and LiDAR sensor can be conducted in parallel, and the output 2D and 3D boxes are fused to yield more accurate 3D detection results. CLOCs~\cite{clocs} introduces a sparse tensor that contains paired 2D-3D boxes and learns the final object confidence scores from this sparse tensor. \cite{fast-clocs} improves~\cite{clocs} by introducing a light-weight 3D detector-cued image detector.

\noindent\textbf{Analysis: potentials and challenges of the late-fusion methods.} The late-fusion based approaches focus on the instance-level aggregation and perform multi-modal fusion only on the outputs of different modalities, which avoids complicated interactions on the intermediate features or on the input point cloud. Hence these methods are much more efficient compared to other approaches. However, without resorting to deep features from camera and LiDAR sensors, these methods fail to integrate rich semantic information of different modalities, which limits the potential of this category of methods.


\subsection{Multi-modal detection with radar signals} \label{sec:radar}

\noindent\textbf{Problem and Challenge.} Radar is an important sensory type in driving systems. In contrast to LiDAR sensors, radar has four irreplaceable advantages in real-world applications: Radar is much cheaper than LiDAR sensors; Radar is less vulnerable to extreme weather conditions; Radar has a larger detection range; Radar provides additional velocity measurements. Nevertheless, compared to LiDAR sensors that generate dense point clouds, radar only provides sparse and noisy measurements. Hence, how to effectively handle the radar signals remains a critical challenge. 

\noindent\textbf{Radar-LiDAR fusion.} Many papers try to fuse the two modalities by introducing new fusion mechanisms to enable message passing between the radar and LiDAR signals, including voxel-based fusion~\cite{radarnet}, attention-based fusion~\cite{radar-qian}, introducing a range-azimuth-doppler tensor~\cite{radar-major}, leveraging graph neural networks~\cite{radar-meyer}, exploiting dynamic occupancy maps~\cite{sparse-pointnet}, and introducing 4D radar data~\cite{4Dradar}.

\noindent\textbf{Radar-camera fusion.} Radar-camera fusion is quite similar to LiDAR-camera fusion, as both radar and LiDAR data are 3D point representations. Most radar-camera approaches~\cite{radar-chadwick, radar-nabati-first, radar-nabati} adapt the existing LiDAR-based detection architectures to handle sparse radar points and adopt similar fusion strategies as LiDAR-camera based methods.

\subsection{Multi-modal detection with high-definition maps} \label{sec:map}

\noindent\textbf{Problem and Challenge.} High-definition maps (HD maps) contain detailed road information such as road shape, road marking, traffic signs, barriers, \textit{etc.} HD maps provide rich semantic information on surrounding environments and can be leveraged as a strong prior to assist 3D object detection. How to effectively incorporate map information into a 3D object detection framework has become an open challenge to the research community.

\begin{figure*}[t]
	\centering
	\includegraphics[width=\textwidth]{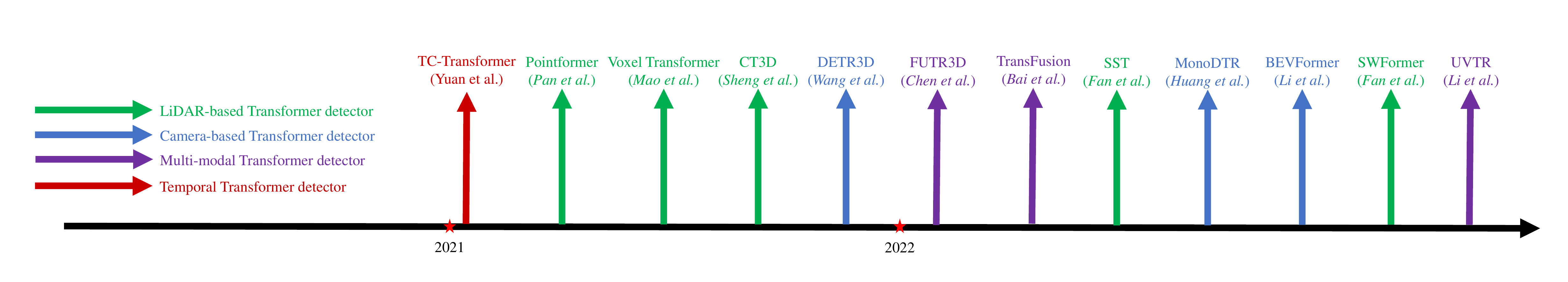}
	\caption{Chronological overview of Transformer-based 3D object detectors.}
	\label{fig18-transformer-roadmap}
	\vspace{-2mm}
\end{figure*}

\noindent\textbf{Multi-modal detection with map information.} High-definition maps can be readily transformed into a bird's-eye view representation and fused with rasterized BEV point clouds or feature maps. The fusion can be conducted by simply concatenating the channels of a rasterized point cloud and an HD map from the bird's-eye view~\cite{hdnet}, feeding LiDAR point cloud and HD map into separate backbones and fusing the output feature maps of the two modalities~\cite{hdmap-fang}, or simply filtering out those predictions that do not fall into the relevant map regions~\cite{nuscenes20}. Other map types have also been explored, \textit{e.g.} visibility map~\cite{hdmap-hu}, vectorized map~\cite{pippnp}.

\section{Transformer-based 3D Object Detection} \label{sec:transfomer}

In this section, we introduce the Transformer-based 3D object detection methods. Transformers~\cite{transformer} have shown prominent performance in many computer vision tasks, and many endeavors have been made to adapt Transformers to 3D object detection. In Section~\ref{sec:transformer-arch}, we review the Transformers tailored for 3D object detection from an architectural perspective. In Section~\ref{sec:transformer-app}, we introduce the applications of Transformers in different 3D object detectors. 

\subsection{Transformer architectures for 3D object detection} \label{sec:transformer-arch}

\noindent\textbf{Problem and Challenge.} While most 3D object detectors are based on convolutional architectures, recently Transformer-based 3D detectors have shown great potential and dominated 3D object detection leaderboards. Compared to convolutional networks, the query-key-value design in Transformers enables more flexible interactions between different representations and the self-attention mechanism results in a larger receptive field than convolutions. However, fully-connected self-attention has quadratic time and space complexity \textit{w.r.t.} the number of inputs, training Transformers can easily fall into sub-optimal results when the data size is small. Hence, it's critical to define proper query-key-value triplets and design specialized attention mechanisms for Transformer-based 3D object detectors.

\noindent\textbf{Transformer architectures.} The development of Transformer architectures in 3D object detection has experienced three stages: (1) Inspired by vanilla Transformer~\cite{transformer}, new Transformer modules with special attention mechanisms are proposed to obtain more powerful features in 3D object detection. (2) Inspired by DETR~\cite{detr}, query-based Transformer encoder-decoder designs are introduced to 3D object detectors. (3) Inspired by ViT~\cite{vit}, patch-based inputs and architectures similar to Vision Transformers are introduced in 3D object detection. 

In the first stage, many papers try to introduce novel Transformer modules into conventional 3D detection pipelines. In these papers, the choices of query, key, and value are quite flexible and new attention mechanisms are proposed. Pointformer~\cite{pointformer} introduces Transformer modules to point backbones. It takes point features and coordinates as queries and applies self-attention to a group of point clouds. Voxel Transformer~\cite{votr} replaces convolutional voxel backbones with Transformer modules, where sparse and submanifold voxel attention are proposed and applied to voxels. CT3D~\cite{ct3d} proposes a novel Transformer-based detection head, where proposal-to-point attention and channel-wise attention are introduced.  

In the second stage, many papers propose DETR-like architectures for 3D object detection. They leverage a set of object queries and use those queries to interact with different features to predict 3D boxes. DETR3D~\cite{detr3d} introduces object queries and generates a 3D reference point for each query. They use reference points to aggregate multi-view image features as keys and values, and apply cross-attention between object queries and image features. Finally, each query can decode a 3D bounding box for detection. Many following works have adopted the design of object queries and reference points. BEVFormer~\cite{bevformer} generates dense queries from BEV grids. TransFusion~\cite{transfusion} produces object queries from initial detections and applies cross-attention to LiDAR and image features in a Transformer decoder. UVTR~\cite{uvpr} fuses object queries with image and LiDAR voxels in a Transformer decoder. FUTR3D~\cite{futr3d} fuses object queries with features from different sensors in a unified way.

In the third stage, many papers try to apply the designs of Vision Transformers to 3D object detectors. Following~\cite{vit, swin}, they split inputs into patches and apply self-attention within each patch and across different patches. SST~\cite{sst} proposes a sparse Transformer, in which voxels in a local region are grouped into a patch and sparse regional attention is applied to the voxels in a patch, and then region shift is applied to change the grouping so new patches can be generated. SWFormer~\cite{swformer} improves~\cite{sst} with multi-scale feature fusion and voxel diffusion.

\subsection{Transformer applications in 3D object detection} \label{sec:transformer-app}

\noindent\textbf{Applications of Transformer-based 3D detectors.} Transformer architectures have been broadly adopted in various types of 3D object detectors. For point-based 3D object detectors, a point-based Transformer~\cite{pointformer} has been developed to replace the conventional PointNet backbone. For voxel-based 3D detectors, a lot of papers~\cite{votr, sst, swformer} propose novel voxel-based Transformers to replace the conventional convolutional backbone. For point-voxel based 3D object detectors, a new Transformer-based detection head~\cite{ct3d} has been proposed for better proposal refinement. For monocular 3D object detectors, Transformers can be used to fuse image and depth features~\cite{monodtr}. For multi-view 3D object detectors, Transformers are utilized to fuse multi-view image features for each query~\cite{bevformer, detr3d}. For multi-modal 3D object detectors, many papers~\cite{transfusion, uvpr, futr3d} leverage Transformer architectures and special cross-attention mechanisms to fuse features of different modalities. For temporal 3D object detectors, Temporal-Channel Transformer~\cite{seq-lidar-yuan} is proposed to model temporal relationships across LiDAR frames.

\begin{figure*}[t]
	\centering
	\includegraphics[width=\textwidth]{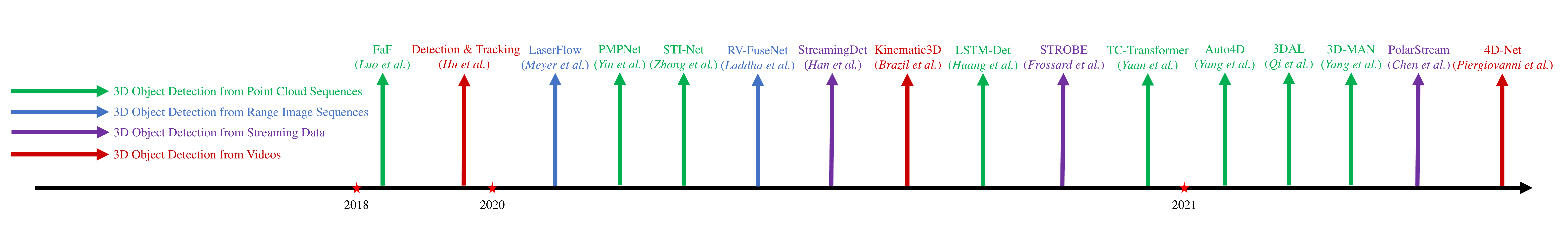}
	\caption{Chronological overview of the temporal 3D object detection methods.}
	\label{fig18-temporal-roadmap}
	\vspace{-2mm}
\end{figure*}

\section{Temporal 3D Object Detection} \label{sec:temporal}

In this section, we introduce the temporal 3D object detection methods. Based on the data types, these methods can be divided into three categories: detection from LiDAR sequences, detection from streaming inputs, and detection from videos. In Section~\ref{sec:seq-lidar}, we review the 3D object detection methods leveraging sequential LiDAR sweeps. In Section~\ref{sec:seq-stream}, we introduce the detection approaches with streaming data as input. In Section~\ref{sec:seq-video}, we investigate 3D detection from videos and multi-modal temporal data. A chronological overview of the temporal detection approaches is shown in Figure~\ref{fig18-temporal-roadmap} and a taxonomy is in Table~\ref{tab:temporal_tax}.

\begin{table}[t!]
	\caption{A taxonomy of temporal 3D object detection methods based on input representations.}
	\centering
	\begin{tabular}{|c|c|c|}
		\hline
		\multicolumn{2}{|c|}{Input}  & Methods \\
		\hline
		\hline
		\multirow{3}*{LiDAR} & multi-frame point clouds & \tabincell{c}{\cite{seq-lidar-huang, seq-lidar-yang, seq-lidar-zhang, seq-lidar-yin, seq-lidar-yuan} \\ \cite{offboard3d, auto4d}} \\
		\cline{2-3}
		&  multi-frame range images & \cite{laserflow, rv-fusenet} \\
		\cline{2-3}
		&  streaming inputs & \cite{seq-stream-han, seq-stream-frossard, seq-stream-chen} \\
		\hline
		Camera & videos & \cite{mono-video, Kinematic3D, 4d-net} \\
		\hline
	\end{tabular}
	\label{tab:temporal_tax}
	\vspace{-2mm}
\end{table}

\begin{figure}[t]
	\centering
	\includegraphics[width=0.48\textwidth]{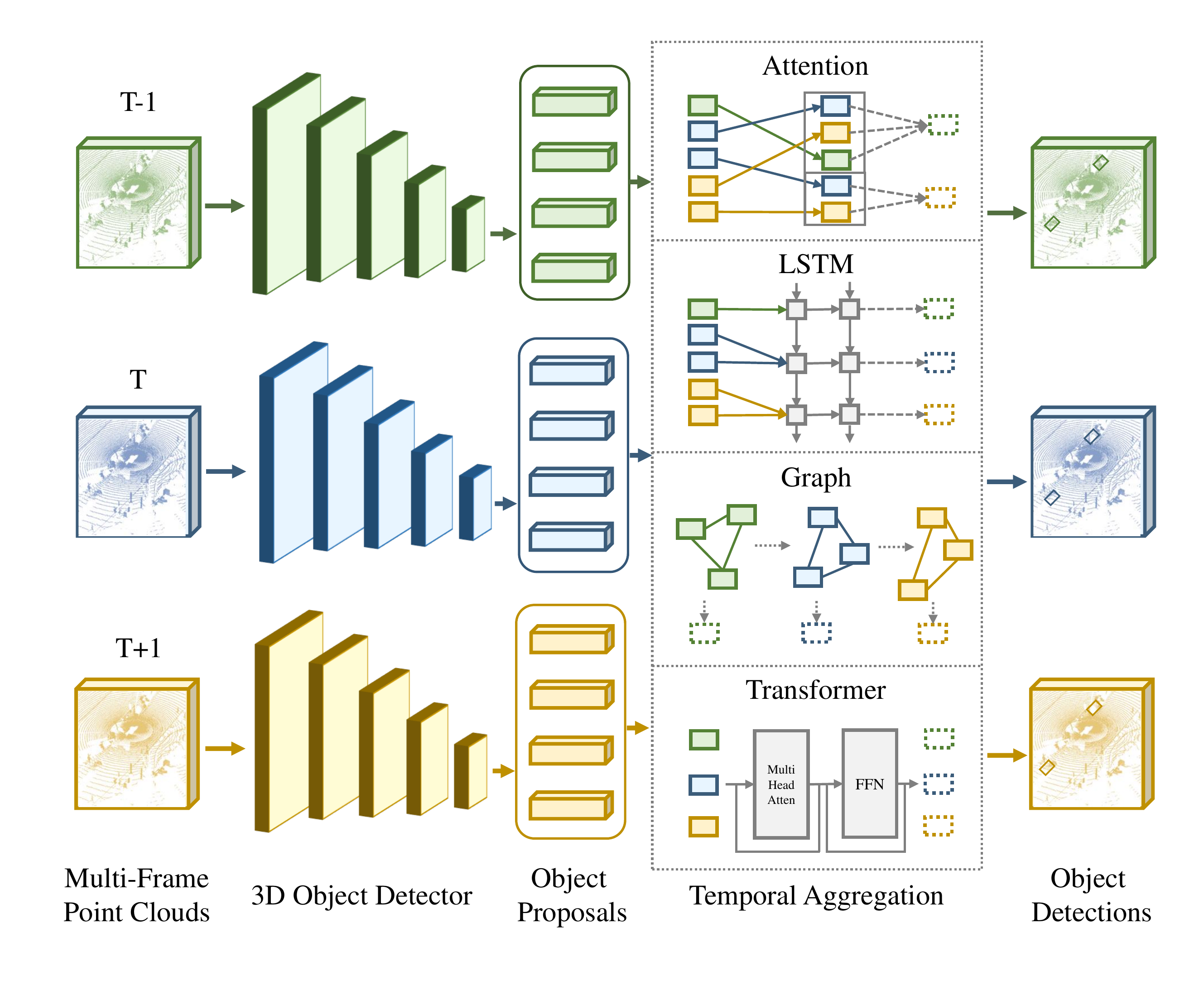}
	\caption{An illustration of detection from LiDAR sequences.} 
	\label{fig19-temporal-det}
	\vspace{-2mm}
\end{figure}

\subsection{3D object detection from LiDAR sequences} \label{sec:seq-lidar}

\noindent\textbf{Problem and Challenge.} While most methods focus on detection from a single-frame point cloud, there also exist many approaches leveraging multi-frame point clouds for more accurate 3D object detection. These methods are trying to tackle the temporal detection problem by fusing multi-frame features via various temporal modeling tools, and they can also obtain more complete 3D shapes by merging multi-frame object points into a single frame. Temporal 3D object detection has exhibited great success in offline 3D auto-labeling pipelines. However, in onboard applications, these methods still suffer from memory and latency issues, as processing multiple frames inevitably brings additional time and memory costs, which can become severe when models are running on embedded devices. An illustration of temporal 3D object detection from LiDAR sequences is shown in Figure~\ref{fig19-temporal-det}. 

\noindent\textbf{3D object detection from sequential sweeps.} Most detection approaches using multi-frame point clouds resort to proposal-level temporal information aggregation. Namely, 3D object proposals are first generated independently from each frame of point cloud through a shared detector, and then various temporal modules are applied on the object proposals and the respective RoI features to aggregate the information of objects across different frames. The adopted temporal aggregation modules include temporal attention~\cite{seq-lidar-yang}, ConvGRU~\cite{seq-lidar-yin}, graph network~\cite{seq-lidar-zhang}, LSTM~\cite{seq-lidar-huang}, and Transformer~\cite{seq-lidar-yuan}. Temporal 3D object detection is also applied in the 3D object auto-labeling pipelines~\cite{offboard3d, auto4d}. In addition to temporal detection from multi-frame point clouds, there are also some works~\cite{laserflow, rv-fusenet} leveraging sequential range images for 3D object detection. 

\begin{figure}[t]
	\centering
	\includegraphics[width=0.48\textwidth]{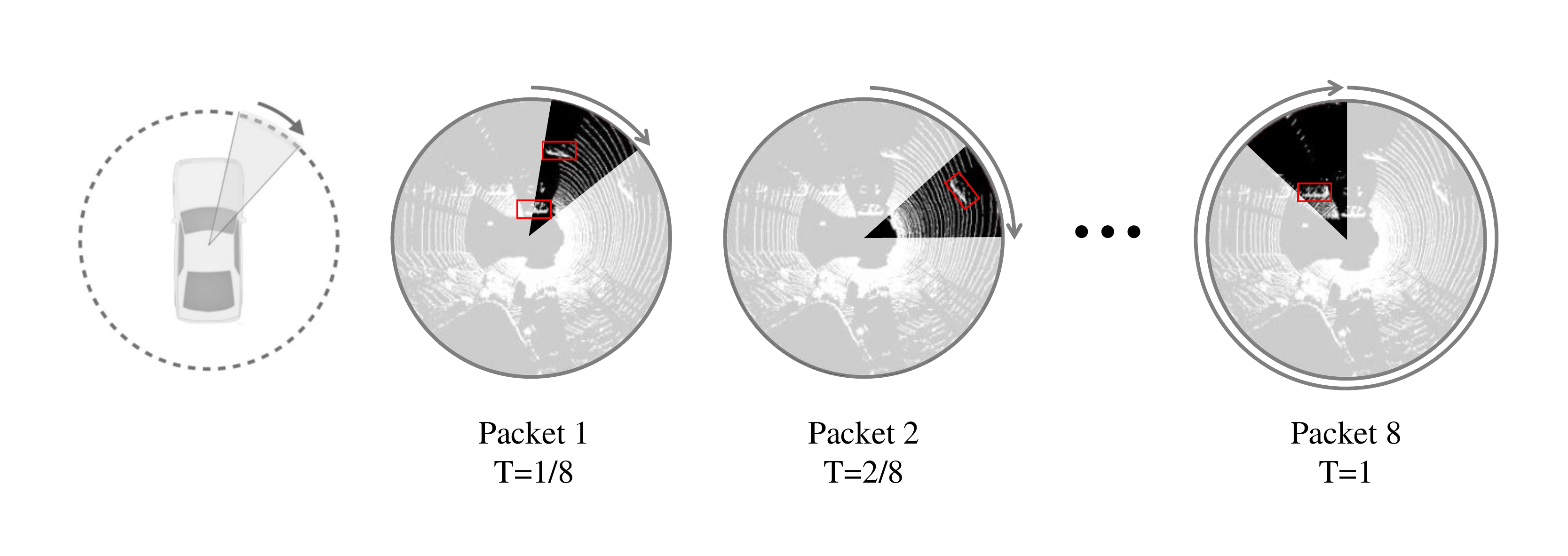}
	\caption{An illustration of streaming 3D object detection. Reference: ~\cite{seq-stream-frossard} and \cite{seq-stream-chen}.}
	\label{fig20-stream-det}
	\vspace{-2mm}
\end{figure}

\subsection{3D object detection from streaming data} \label{sec:seq-stream}

\noindent\textbf{Problem and Challenge.} Point clouds collected by rotating LiDARs are intrinsically a streaming data source in which LiDAR packets are sequentially recorded in a sweep. It typically takes $50$-$100$ ms for a rotating LiDAR sensor to generate a $360^{\circ}$ complete LiDAR sweep, which means that by the time a point cloud is produced, it no longer accurately reflects the scene at the exact time. This poses a challenge to autonomous driving applications which generally require minimal reaction times to guarantee driving safety. Many endeavors have been made to directly detect 3D objects from the streaming data. These methods generally detect 3D objects on the active LiDAR packets immediately without waiting for the full sweep to be built. Streaming 3D object detection is a more accurate and low-latency solution to vehicle perception compared to detection from full LiDAR sweeps. An illustration of 3D object detection from streaming data is shown in Figure~\ref{fig20-stream-det}. 

\noindent\textbf{Streaming 3D object detection.} Similar to temporal detection from multi-frame point clouds, streaming detection methods~\cite{seq-stream-han} can treat each LiDAR packet as an independent sample to detect 3D objects and apply temporal modules on the sequential packets to learn the inter-packets relationships. However, a LiDAR packet normally contains an incomplete point cloud and the information from a single packet is generally not sufficient for accurately detecting 3D objects. To this end, some papers try to provide more context information for detection in a single packet. The proposed techniques include a spatial memory bank~\cite{seq-stream-frossard} and a multi-scale context padding scheme~\cite{seq-stream-chen}.

\subsection{3D object detection from videos} \label{sec:seq-video}

\noindent\textbf{Problem and Challenge.} Video is an important data type and can be easily obtained in autonomous driving applications. Compared to single-image based 3D object detection, video-based 3D detection naturally benefits from the temporal relationships of sequential images. While numerous works focus on single-image based 3D object detection, only a few papers investigate the problem of 3D object detection from videos, which leaves an open challenge to the research community.  

\begin{figure}[t]
	\centering
	\includegraphics[width=0.48\textwidth]{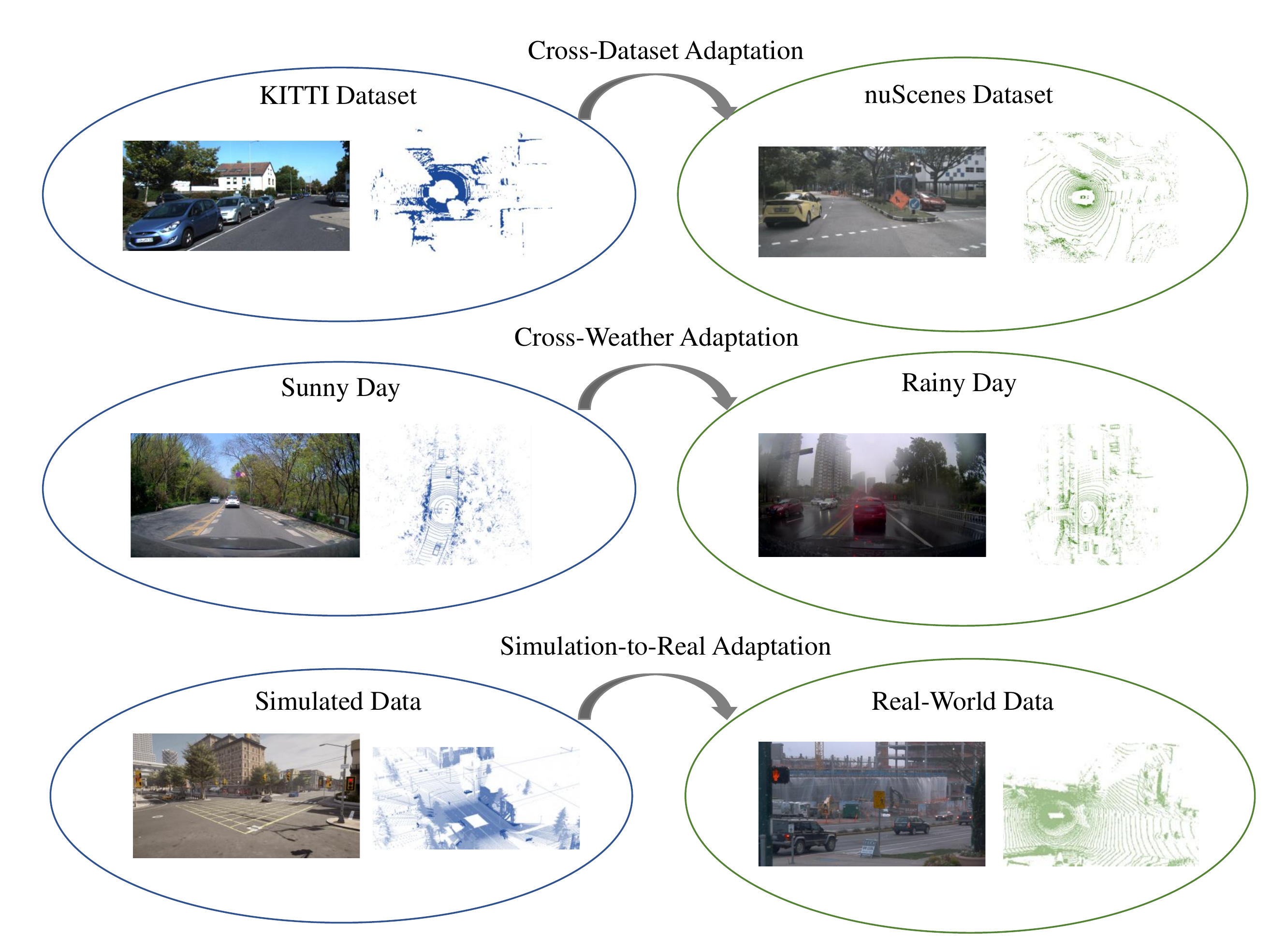}
	\caption{An illustration of domain gaps in 3D detection.}
	\label{fig21-domain-adaptation}
	\vspace{-2mm}
\end{figure} 

\noindent\textbf{Video-based 3D object detection.} Video-based detection approaches generally extend the image-based 3D object detectors by tracking and fusing the same objects across different frames. The proposed trackers include LSTM~\cite{mono-video} and the 3D Kalman filter~\cite{Kinematic3D}. In addition, there are some works~\cite{4d-net, LIFT} leveraging both videos and multi-frame point clouds for more accurate 3D object detection. Those methods propose 4D sensor-time fusion to learn features from both temporal and multi-modal data.
 
\section{Label-Efficient 3D Object Detection} \label{sec:xlreaning}

In this section, we introduce the methods of label-efficient 3D object detection. In previous sections, we generally assume the 3D detectors are trained under full supervision on a specific data domain and with a sufficient amount of annotations. However, in real-world applications, the 3D object detection methods inevitably face the problems of poor generalizability and lacking annotations. To address these issues, label-efficient techniques can be employed in 3D object detection, including domain adaptation (Section~\ref{sec:label-da}), weakly-supervised learning (Section~\ref{sec:label-weak}), semi-supervised learning (Section~\ref{sec:label-semi}), and self-supervised learning (Section~\ref{sec:label-self}) for 3D object detection. We will introduce those techniques in the following sections.

\begin{table}[t!]
	\caption{A taxonomy of domain adaptation methods for 3D object detection based on transferred domains and techniques.}
	\centering
	\begin{tabular}{|c|c|c|}
		\hline
		\tabincell{c}{Method} &  \tabincell{c}{Transferred \\ Domain} &  \tabincell{c}{Technique} \\
		\hline
		\hline
		Wang \textit{et al.}~\cite{trainingermany20} & cross-sensor & statistics normalization \\
		\hline
		SF-UDA$^{3D}$~\cite{sf-uda} & cross-sensor & self-training \\
		\hline
		ST3D~\cite{st3d} & cross-sensor & self-training \\
		\hline
		FAST3D~\cite{fast3d} & cross-sensor & self-training \\
		\hline
		SRDAN~\cite{srdan} & cross-sensor & domain alignments  \\
		\hline
		MLC-Net~\cite{mlc-net} & cross-sensor & domain alignments \\
		\hline
		3D-CoCo~\cite{3d-coco} & cross-sensor & domain alignments \\
  		\hline
		Rist \textit{et al.}~\cite{da-rist} & cross-sensor & multi-task learning \\
		\hline
		PIT~\cite{pit} & cross-sensor & image transform \\
		\hline
		SPG~\cite{spg} & cross-weather & semantic point generation \\
		\hline
		Saleh \textit{et al.}~\cite{da-saleh} & sim-to-real & Cycle GAN \\
		\hline
		DeBortoli \textit{et al.}~\cite{da-DeBortoli} & sim-to-real & adversarial training \\
		\hline
	\end{tabular}
	\label{tab:da_tax}
	\vspace{-2mm}
\end{table}

\subsection{Domain adaptation for 3D object detection} \label{sec:label-da}

\noindent\textbf{Problem and Challenge.} Domain gaps are ubiquitous in the data collection process. Different sensor settings and placements, different geographical locations, and different weathers will result in completely different data domains. In most conditions, 3D object detectors trained on a certain domain cannot perform well on other domains. Many techniques have been proposed to address the domain adaptation problem for 3D object detection, \textit{e.g.} leveraging consistency between source and target domains, and self-training on target domains. Nevertheless, most methods only focus on solving one specific domain transfer problem. Designing a domain adaptation approach that can be generally applied in any domain transfer tasks in 3d object detection will be a promising research direction. An illustration of the domain gaps in 3D object detection is shown in Figure~\ref{fig21-domain-adaptation} and a taxonomy of the domain adaptive methods is in Table~\ref{tab:da_tax}

\noindent\textbf{Cross-sensor domain adaptation.} Different datasets have different sensory settings, e.g. a 32-beam LiDAR sensor used in nuScenes~\cite{nuscenes20} versus a 64-beam LiDAR sensor in KITTI~\cite{kitti12conf}, and the data is also collected at different geographic locations, e.g. KITTI~\cite{kitti12conf} is collected in Germany while Waymo~\cite{waymo20} is collected in United States. These factors will lead to severe domain gaps between different datasets, and the detectors trained on a dataset generally exhibit quite poor performance when they are tested on other datasets. \cite{trainingermany20} is a notable work that observes the domain gaps between datasets, and they introduce a statistic normalization approach to handle the gaps. Many following works leverage self-training to resolve the domain adaptation problem. In those methods, a detector pre-trained on the source dataset will produce pseudo labels for the target dataset, and then the detector is re-trained on the target dataset with pseudo labels. These methods make improvements mainly on obtaining pseudo labels of higher quality, \textit{e.g.} \cite{sf-uda} proposes a scale-and-detect strategy, \cite{st3d} introduces a memory bank, \cite{fast3d} leverages the scene flow information, and \cite{da-you} exploits playbacks to enhance the quality of pseudo labels. In addition to the self-training approaches, there also exist some papers building alignments between source and target domains. The domain alignments can be established through a scale-aware and range-aware alignment strategy~\cite{srdan}, multi-level consistency~\cite{mlc-net}, and a contrastive co-training scheme~\cite{3d-coco}.

In addition to the domain gaps among datasets, different sensors also produce data of distinct characteristics. A 32-beam LiDAR produces much sparser point clouds compared to a 64-beam LiDAR, and images obtained from different cameras also have diverse sizes and intrinsics. \cite{da-rist} introduces a multi-task learning scheme to tackle the domain gaps between different LiDAR sensors, and \cite{pit} proposes the position-invariant transform to address the domain gaps between different cameras.

\noindent\textbf{Cross-weather domain adaptation.} Weather conditions have a huge impact on the quality of collected data. On rainy days, raindrops will change the surface property of objects so that fewer LiDAR beams can be reflected and detected, so point clouds collected on rainy days are much sparser than those obtained under dry weather. Besides fewer reflections, rain also causes false positive reflections from raindrops in mid-air. \cite{spg} addresses the cross-weather domain adaptation problem with a novel semantic point generation scheme.

\noindent\textbf{Sim-to-real domain adaptation.} Simulated data has been broadly adopted in 3D object detection, as the collected real-world data cannot cover all driving scenarios. However, the synthetic data has quite different characteristics from the real-world data, which gives rise to a sim-to-real adaptation problem. Many approaches are proposed to resolve this problem, including GAN~\cite{cycle-gan} based training~\cite{da-saleh} and introducing an adversarial discriminator~\cite{da-DeBortoli} to distinguish real and synthetic data.

\subsection{Weakly-supervised 3D object detection} \label{sec:label-weak}

\begin{figure}[t]
	\centering
	\includegraphics[width=0.48\textwidth]{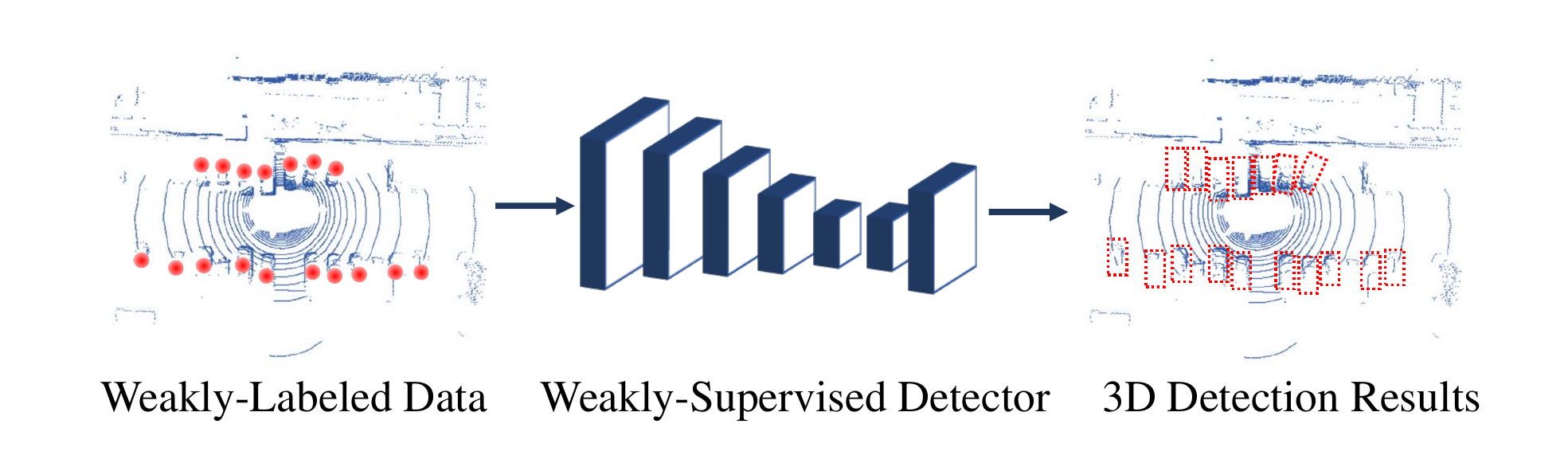}
	\caption{An illustration of weakly-supervised 3D detection.}
	\label{fig22-weak-learning}
	\vspace{-3mm}
\end{figure}

\noindent\textbf{Problem and Challenge.} Existing 3D object detection methods highly rely on training with vast amounts of manually labeled 3D bounding boxes, but annotating those 3D boxes is quite laborious and expensive. Weakly-supervised learning can be a promising solution to this problem, in which weak supervisory signals, \textit{e.g.} less expensive 2D annotations, are exploited to train the 3D object detection models. Weakly-supervised 3D object detection requires fewer human efforts for data annotation, but there still exists a non-negligible performance gap between the weakly-supervised and the fully-supervised methods. An illustration of weakly-supervised 3D object detection is shown in Figure~\ref{fig22-weak-learning}.

\noindent\textbf{Weakly-supervised 3D object detection.} Weakly-supervised approaches leverage weak supervision instead of fully annotated 3D bounding boxes to train 3D object detectors. The weak supervisions include 2D image bounding boxes~\cite{weak-wei, weak-peng}, a pre-trained image detector~\cite{weak-qin}, BEV object centers and vehicle instances~\cite{weak-meng, weak-meng-journal}. Those methods generally design novel learning mechanisms to skip the 3D box supervision and learn to detect 3D objects by mining useful information from weak signals.

\subsection{Semi-supervised 3D object detection} \label{sec:label-semi}

\begin{figure}[t]
	\centering
	\includegraphics[width=0.48\textwidth]{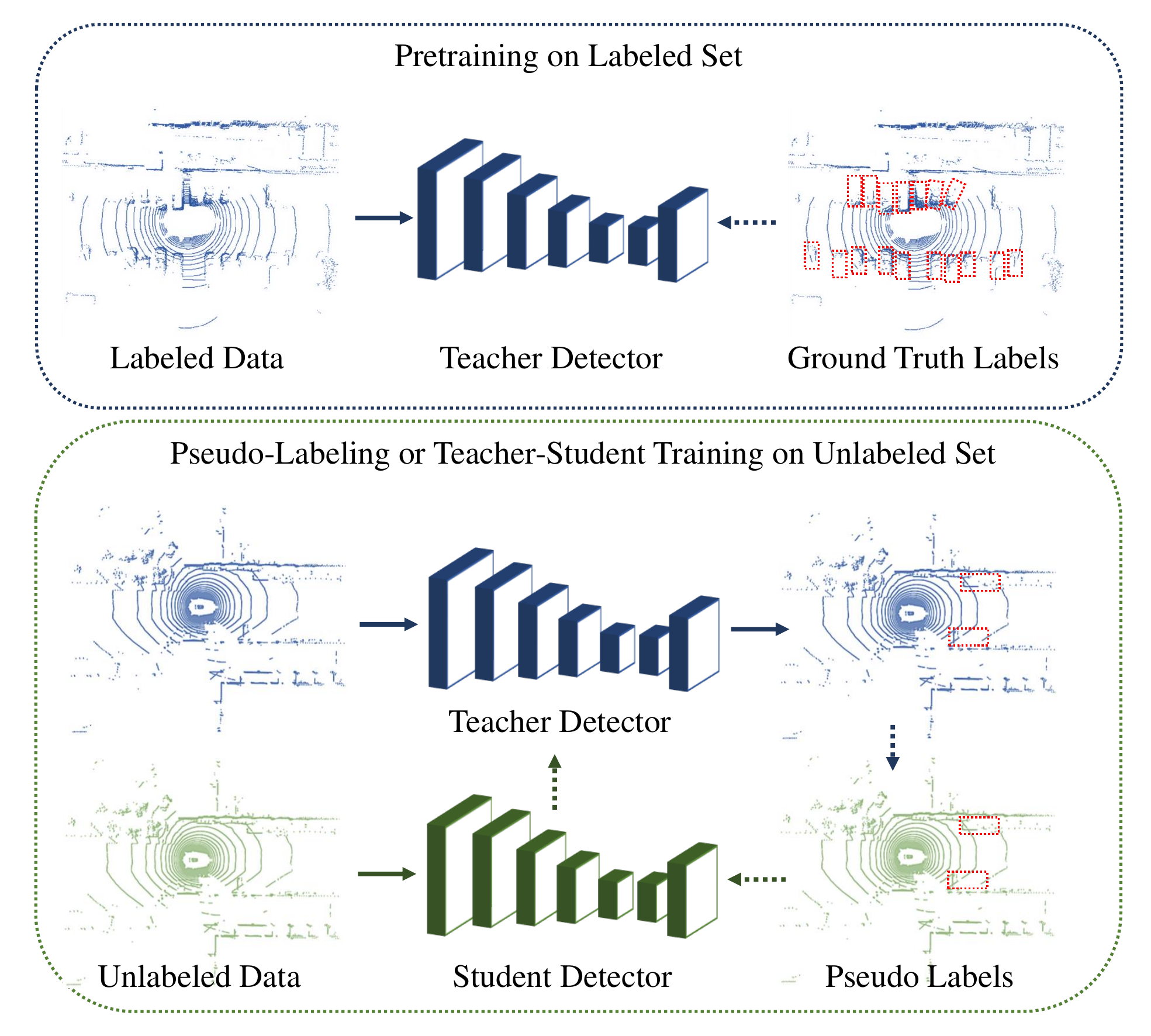}
	\caption{An illustration of semi-supervised 3D detection.}
	\label{fig23-semi-learning}
	\vspace{-3mm}
\end{figure}

\noindent\textbf{Problem and Challenge.} In real-world applications, data annotation requires much more human effort than data collection. Typically a data acquisition vehicle can collect more than $100$k frames of point clouds in a day, while a skilled human annotator can only annotate $100$-$1$k frames per day. This will inevitably lead to a rapid accumulation of a large amount of unlabeled data. Hence how to mine useful information from large-scale unlabeled data has become a critical challenge to both the research community and the industry. Semi-supervised learning, which exploits a small amount of labeled data and a huge amount of unlabeled data to jointly train a stronger model, is a promising direction. Combining 3D object detection with semi-supervised learning can boost detection performance. An illustration of semi-supervised 3D object detection is shown in Figure~\ref{fig23-semi-learning}.

\begin{figure}[t]
	\centering
	\includegraphics[width=0.48\textwidth]{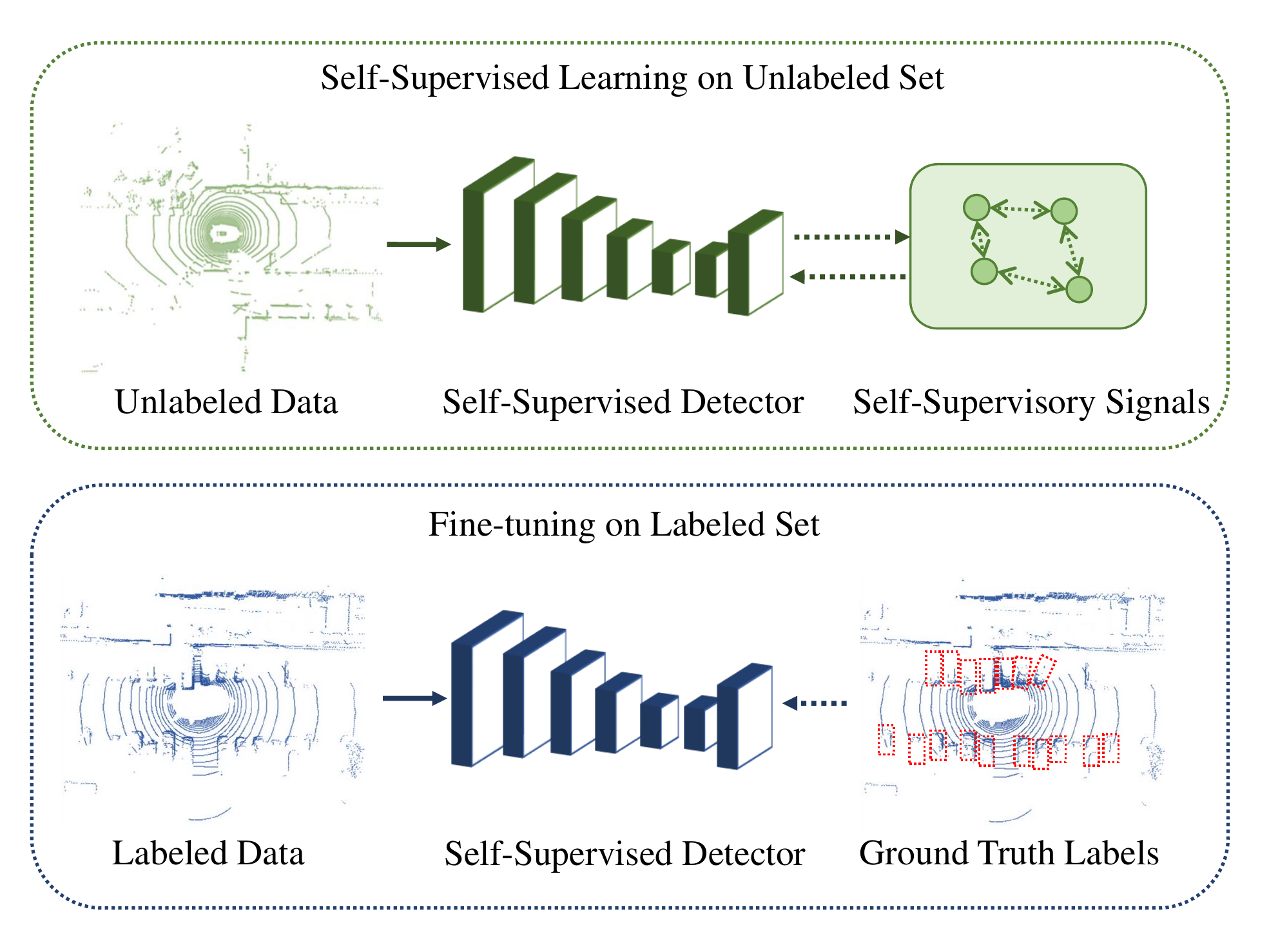}
	\caption{An illustration of self-supervised 3D detection.}
	\label{fig25-self-learning}
	\vspace{-3mm}
\end{figure}

\noindent\textbf{Semi-supervised 3D object detection.} There are mainly two categories of approaches in semi-supervised 3D object detection: pseudo-labeling and teacher-student learning. The pseudo labeling approaches~\cite{pseudo-labeling, 3dioumatch} first train a 3D object detector with the labeled data, and then use the 3D detector to produce pseudo labels for the unlabeled data. Finally, the 3D object detector is re-trained with the pseudo labels on the unlabeled domain. The teacher-student methods~\cite{se-ssd} adapt the Mean Teacher~\cite{mean-teacher} training paradigm to 3D object detection. Specifically, a teacher detector is first trained on the labeled domain, and then the teacher detector guides the training of a student detector on the unlabeled domain by encouraging the output consistencies between the two detection models.   

\subsection{Self-supervised 3D object detection} \label{sec:label-self}

\noindent\textbf{Problem and Challenge.} Self-supervised pre-training has become a powerful tool when there exists a large amount of unlabeled data and limited labeled data. In self-supervised learning, models are first pre-trained on large-scale unlabeled data and then fine-tuned on the labeled set to obtain a better performance. In autonomous driving scenarios, self-supervised pre-training for 3D object detection has not been widely explored. Existing methods are trying to adapt the self-supervised methods, \textit{e.g.} contrastive learning, to the 3D object detection problem, but the rich semantic information in multi-modal data has not been well exploited. How to effectively handle the raw point clouds and images to pre-train an effective 3D object detector remains an open challenge. An illustration of self-supervised 3D object detection is in Figure~\ref{fig25-self-learning}.

\noindent\textbf{Self-supervised 3D object detection.} Self-supervised methods generally apply the contrastive learning techniques~\cite{moco, moco-v2} to 3D object detection. Specifically, an input point cloud is first transformed into two views with augmentations, and then contrastive learning is employed to encourage the feature consistencies of the same 3D locations across the two views. Finally, the 3D detector pre-trained with contrastive learning is further fine-tuned on the labeled set to attain better performance. PointContrast~\cite{pointcontrast} first introduces the contrastive learning paradigm in 3D object detection, and the following papers improve this paradigm by leveraging the depth information~\cite{DepthContrast} and clustering~\cite{gcc-3d}. In addition to self-supervised learning for point cloud detectors, there are also some works trying to exploit both point clouds and images for self-supervised 3D detection, \textit{e.g.} \cite{simipu} proposes an intra-modal and inter-modal contrastive learning scheme on the multi-modal inputs.

\section{3D Object Detection in Driving Systems} \label{sec:system}
In this section, we introduce some critical problems of 3D object detection in driving systems. In Section~\ref{sec:system-end-to-end}, we review and analyze the approaches in which 3D object detection is trained together with other tasks, \textit{e.g.} tracking, trajectory prediction, motion planning, localization, in an end-to-end manner. In Section~\ref{sec:system-simulation}, we introduce the simulation systems designed for 3D object detection and autonomous driving. In Section~\ref{sec:system-robust}, we investigate the research topics on the robustness of 3D object detectors and safety-aware 3D object detection. In Section~\ref{sec:system-cooperative}, we review the approaches of collaborative 3D object detection.

\subsection{End-to-end learning for autonomous driving} \label{sec:system-end-to-end}

\begin{figure*}[t]
	\centering
	\includegraphics[width=\textwidth]{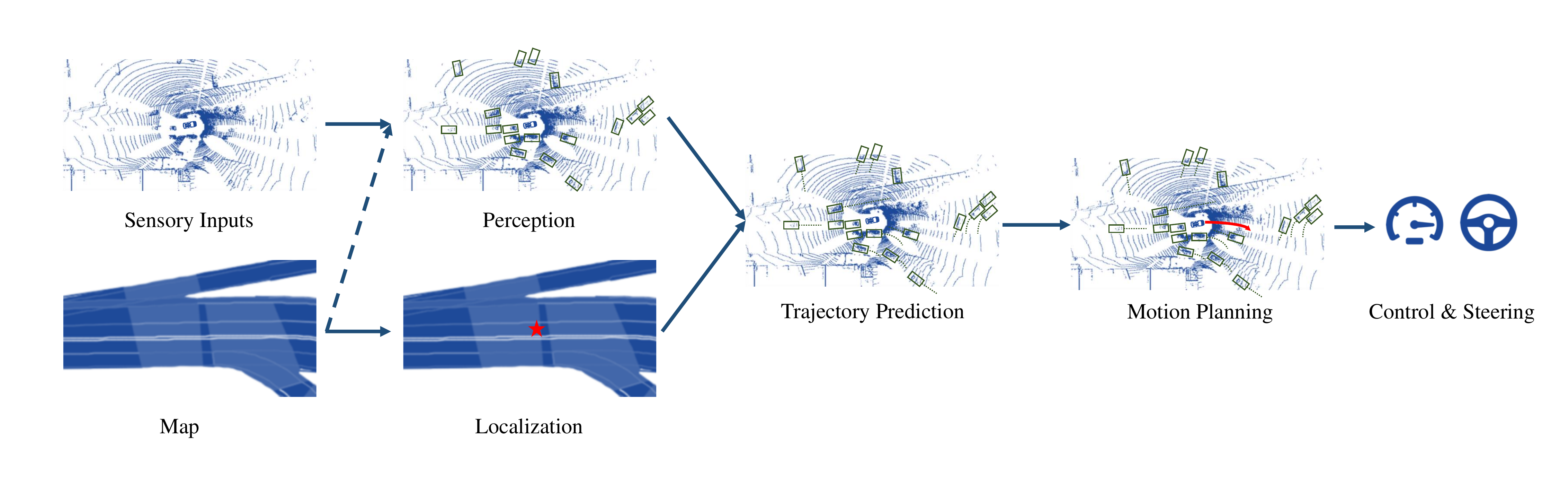}
	\caption{An illustration of the autonomous driving pipeline. Point cloud and map samples are from \cite{intentnet}.} 
	\label{fig25-end-to-end}
	\vspace{-2mm}
\end{figure*}

\noindent\textbf{Problem and Challenge.} 3D object detection is a critical component of perception systems, and the performance of 3D object detectors will have a profound influence on downstream tasks like tracking, prediction, and planning. Hence from the systematic perspective, jointly training 3D object detection models with other perception tasks as well as the downstream tasks will be a better solution to autonomous driving. An open challenge is how to involve all driving tasks in a unified framework and jointly train these tasks in an end-to-end manner. An illustration of end-to-end autonomous driving is shown in Figure~\ref{fig25-end-to-end}.

\noindent\textbf{Joint perception and prediction.} There are many works learning to perceive and track 3D objects and then predict their future trajectories in an end-to-end manner. FaF~\cite{fnf18} is a seminal work that proposes to jointly reason about 3D object detection, tracking, and trajectory prediction with a single 3D convolutional network. This design paradigm is followed by a lot of papers with improvements, \textit{e.g.} \cite{intentnet} leverages the map information, \cite{interaction-transformer} introduces an interactive Transformer, \cite{stinet} designs a spatial-temporal-interactive network, \cite{motionnet} proposes a spatio-temporal pyramid network, \cite{pnpnet} conducts all the tasks in a loop, \cite{end-to-end-phillip} involves the localization task into the system.

\noindent\textbf{Joint perception, prediction, and planning.} Many endeavors have been made to involve perception, prediction, and planning in a unified framework. Compared to the joint perception and prediction approaches, the whole system can benefit from the planner's feedback by adding motion planning to the end-to-end pipeline. Many techniques have been proposed to improve this framework, \textit{e.g.} \cite{perceive-predict-plan} introduces a semantic occupancy map to produce interpretable intermediate representations, \cite{perceive-attend-drive} incorporates spatial attention into the framework, \cite{dsdnet} proposes a deep structured network, \cite{mp3} proposes a map-free approach, \cite{lookout} produces a diverse set of future trajectories.

\noindent\textbf{End-to-end learning for autonomous driving.} Some methods try to build a completely end-to-end autonomous driving system, in which an autonomous vehicle takes sensory inputs and sequentially performs perception, prediction, planning, and motion control in a loop, and finally produces steering and speed signals for driving. \cite{end-to-end-bojarski} first introduces the idea and implements the image-based end-to-end driving system with a convolutional neural network. \cite{end-to-end-modal} proposes an end-to-end architecture with multi-modal inputs. \cite{end-to-end-imitation} and \cite{learning-to-drive-in-a-day} propose to learn end-to-end driving systems with conditional imitation learning and deep reinforcement learning respectively.

\subsection{Simulation for 3D object detection} \label{sec:system-simulation}

\noindent\textbf{Problem and Challenge.} 3D object detection models generally require a large amount of data for training. While the data can be collected in real-world scenarios, the real-world data generally suffers from a long-tail distribution. For example, the scenarios of traffic accidents or extreme weather are seldom recorded but are quite important for training a robust 3D object detector. Simulation is a promising solution to address the long tail data distribution problem, as we can create synthetic data for those rare but critical scenarios. An open challenge for simulation is how to create more realistic synthetic data.

\noindent\textbf{Visual simulation.} Many endeavors have been made to generate photo-realistic synthetic images in driving scenarios. The ideas of those methods include leveraging a graphics engine~\cite{sim-abu, synthia}, exploiting texture-mapped surfels~\cite{surfelgan}, leveraging real-world data~\cite{geosim}, and learning a controllable neural simulator~\cite{drivegan}.

\noindent\textbf{LiDAR simulation.} In addition to generating synthetic images, many approaches try to generate LiDAR point clouds by simulation. Some methods~\cite{sim-fang, sim-nakashima, lidar-aug} propose novel point cloud rendering mechanisms by simulating the real-world effects. Some approaches~\cite{lidarsim} leverage real-world instances to reconstruct 3D scenes. Other papers focus on simulation for safety-critical scenarios~\cite{advsim} or under adverse weather conditions~\cite{sim-hahner}.

\noindent\textbf{Driving simulation.} Many papers try to build an interactive driving simulation platform where a virtual vehicle can perceive and interact with the virtual environments and finally plan the maneuvers. CARLA~\cite{carla} is a pioneering open-source simulator for autonomous driving. Other papers utilize a graphics engine~\cite{airsim}, leverage real-world data~\cite{nuplan}, or develop a data-driven method~\cite{sim-amini} for driving simulation. There are also some works simulating the traffic flows~\cite{scenegen, trafficsim} or testing the safety of vehicles by simulation~\cite{testing-the-safety}.

\subsection{Robustness for 3D object detection} \label{sec:system-robust}

\noindent\textbf{Problem and Challenge.} Learning-based 3D object detectors are generally vulnerable to adversarial attacks. Adding perturbations or objects to the sensory inputs in an adversarial manner can fool the perception models and lead to misdetections. An open challenge of robust 3D object detection is to develop practical adversarial attack and defense algorithms that can be easy to implement and can be applied to most detection models. 

\noindent\textbf{Adversarial attacks on the LiDAR sensors.} Many endeavors have been made to attack the LiDAR sensors and fool the LiDAR-based perception models with adversarial machine learning. Cao \textit{et al.}~\cite{robust-cao-adv} attack the LiDAR sensor and spoof obstacles close to the front of a victim autonomous vehicle. To achieve this goal, they introduce a novel algorithm to strategically control the spoofed attack to fool the LiDAR-based 3D object detection model. Wicker \textit{et al.}~\cite{robust-wicker} study the problem of adversarial attacks on the point-based detection models. They propose an iterative saliency occlusion approach to generate adversarial point cloud examples by dropping critical points. Tu \textit{et al.}~\cite{robust-tu-phy} propose a method to generate physically realizable adversarial examples that can be placed on a vehicle and make this vehicle invisible to the LiDAR-based 3D object detectors. Sun \textit{et al.}~\cite{robust-sun} study the general vulnerability of current LiDAR-based 3D object detection models and identify the ignored occlusion patterns in LiDAR point clouds that make vehicles vulnerable to spoofing attacks. They further propose a black-box spoofing attack method that can fool all target detection models. Zhu \textit{et al.}~\cite{robust-zhu} propose to use arbitrary objects to attack LiDAR-based 3D object detection models. Towards this goal, they introduce a method to identify the adversarial locations in a 3D scene, so that arbitrary objects placed at these locations can fool the LiDAR perception systems. Li \textit{et al.}~\cite{robust-li} exploit the fact that LiDAR point clouds collected from a moving vehicle need calibration based on the moving trajectories, so they propose to spoof the vehicle's trajectory with adversarial perturbations, which can distort the LiDAR sweeps and fool the 3D object detectors. Tu \textit{et al.}~\cite{robust-tu-adv} perform adversarial attacks on the LiDAR perception models under the setting of multi-agent collaborative perception. Specifically, they fool the perception model of an agent by sending an adversarial message from the attacker in the multi-agent communication system.

\noindent\textbf{Adversarial attacks on the multi-modal sensory inputs.} In addition to attacking the LiDAR-based perception models, there exist some works trying to perform adversarial attacks on both cameras and LiDAR sensors simultaneously. Cao \textit{et al.}~\cite{robust-cao-inv} propose to generate a physically-realizable and adversarial 3D object that is invisible to both the camera and LiDAR sensor. The adversarial object is generated through optimization and can be leveraged to attack the multi-sensor fusion-based 3D object detection models. Tu \textit{et al.}~\cite{robust-tu-exp} perform adversarial attacks on multi-modal perception models by introducing an adversarial textured mesh that can be placed on a vehicle and make this vehicle invisible to the multi-modal perception models. Specifically, the adversarial mesh is first rendered into both LiDAR points and image pixels in a differentiable manner, and then the multi-modal inputs are passed through a fusion-based detector. Finally, an adversarial loss is employed to adjust the mesh parameters.

\subsection{Collaborative 3D object detection} \label{sec:system-cooperative}

\noindent\textbf{Problem and Challenge.} Existing 3D detection approaches are mainly based on a single ego-vehicle. However, detecting 3D objects with a single vehicle inevitably meets two challenges: occlusion and sparsity of the far-away objects. To this end, some papers resort to detection under the multi-agent collaborative setting, where an ego-vehicle can communicate with other agents, \textit{e.g.} vehicles or infrastructures, and exploit the information from other agents to improve the perception accuracy. A challenge of collaborative perception is how to properly balance the accuracy improvements and the communication bandwidth requirements. 


\noindent\textbf{Collaborative 3D object detection.} Collaborative detection approaches fuse the information from multiple agents to boost the performance of a 3D object detector. The fused information can be raw sensory inputs from other agents~\cite{cooper, emp}, which cost little communication bandwidth and is quite efficient for detection, and it can also be compressed feature maps~\cite{f-cooper, v2vnet, coop-vadivelu, coop-li}, which cost non-negligible communication bandwidth but generally lead to better detection performance. There are also some papers studying when to communicate with other agents~\cite{when2com} and which agent to communicate~\cite{who2com}. 

\section{Analysis and Outlooks} \label{sec:outlook}

In this section, we conduct a systematic comparison and analysis of the 3D object detection approaches and prospect the future research directions of 3D object detection for autonomous driving. In Section~\ref{sec:analysis}, we conduct a comprehensive analysis of the detection performances and the inference speeds of various 3D object detection methods, \textit{i.e.} LiDAR-based, camera-based, multi-modal approaches, on multiple datasets, from which we further summarize the research trends over the years. In Section~\ref{sec:future}, we propose future research directions in this area.

\subsection{Research trends} \label{sec:analysis}

We comprehensively collect the statistics of various types of 3D object detection methods in recent years. The statistics include performances and inference time of the 3D object detectors on the most broadly-adopted KITTI~\cite{kitti12conf}, nuScenes~\cite{nuscenes20}, and Waymo~\cite{waymo20} dataset. Table~\ref{tab:performance}, Table~\ref{tab:performance_contd} and Table~\ref{tab:performance_contd_contd} show the statistical data. By analyzing these data, we obtain some intriguing findings on the research trends of 3D object detection.

\subsubsection{Trends of dataset selection}

Before 2018, most methods were evaluated on the KITTI dataset, and the evaluation metric they adopted is 2D average precision ($AP_{2D}$), where they project the 3D bounding boxes into the image plane and compare them with the ground truth 2D boxes. From 2018 until now, more and more papers have adopted the 3D or BEV average precision ($AP_{3D}$ or $AP_{BEV}$), which is a more direct metric to measure 3D detection quality. For the LiDAR-based methods, the detection performances on KITTI quickly get converged over the years, \textit{e.g.} $AP_{3D}$ of easy cases increases from $71.40\%$~\cite{ipod} to $90.90\%$~\cite{pv-rcnn++}, and even $AP_{3D}$ of hard cases reaches $79.14\%$ \cite{votr}. Therefore, since 2019, more and more LiDAR-based approaches have turned to larger and more diverse datasets, such as the nuScenes dataset and the Waymo Open dataset. Large-scale datasets also provide more useful data types, \textit{e.g.} raw range images provided by Waymo facilitate the development of range-based methods. For the camera-based detection methods, $AP_{3D}$ of monocular detection on KITTI increases from $1.32\%$~\cite{oft-net} to $23.22\%$~\cite{DD3D}, leaving huge room for improvement. Until now, only a few monocular methods have been evaluated on the Waymo dataset. For the multi-modal detection approaches, the methods before 2019 are mostly tested on the KITTI dataset, and after that most papers resort to the nuScenes dataset, as it provides more multi-modal data.

\begin{table*}[ht]
	\caption{A comprehensive performance analysis of various categories of 3D object detection methods across different datasets. We report the inference time (ms) originally reported in the papers, and report AP$|_{R_{40}}$ (\%) for 3D car detection ($*$ denotes AP$|_{R_{40}}$ for BEV car detection) on the KITTI test benchmark, mAP (\%) and NDS scores on the nuScenes test set, Level 1 (L1) mAP and Level 2 (L2) mAP on the Waymo validation set. We group the methods based on the sensor types and the input representations, and sort them by the year of publication.}
	\centering
	\begin{tabular}{l|c|c|c|c|c|c|c|c|c|c|c}
		\hline
		\multirow{2}*{Method} & \multirow{2}*{Sensor} & \multirow{2}*{Representation} & \multirow{2}*{Year} &  \multirow{2}*{\tabincell{c}{Inference \\ Time (ms)}} & \multicolumn{3}{c|}{KITTI Car} & \multicolumn{2}{c|}{nuScenes} & \multicolumn{2}{c}{Waymo Vehicle} \\
		\cline{6-12}
		& & & & & Easy & Mod. & Hard & mAP & NDS & \tabincell{c}{L1} & \tabincell{c}{L2} \\
		\hline
		\hline
		IPOD~\cite{ipod} & LiDAR & Point & 2018 & - & 71.40 & 53.46 & 48.34 & - & - & - & - \\
		StarNet~\cite{starnet} & LiDAR & Point & 2019 & - & 81.63 & 73.99 & 67.07 & - & - & 53.7 & - \\
		PointRCNN~\cite{pointrcnn19} & LiDAR & Point & 2019 & 100 & 85.94 & 75.76 & 68.32 & - & - & - & - \\
		STD~\cite{std} & LiDAR & Point & 2019 & 80 & 86.61 & 77.63 & 76.06 & - & - & - & - \\
		3DSSD~\cite{3dssd} & LiDAR & Point & 2020 & 38 & 88.36 & 79.57 & 74.55 & 42.6 & 56.4 & - & - \\
		Point-GNN~\cite{pointgnn20} & LiDAR & Point & 2020 & 640 & 88.33 & 79.47 & 72.29 & - & - & - & - \\
		Pointformer~\cite{pointformer} & LiDAR & Point & 2021 & - & 87.13 & 77.06 & 69.25 & 53.6 & - & - & - \\
		\hline
		Vote3D~\cite{vote3d} & LiDAR & Voxel & 2015 & 500 & - & - & - & - & - & - & - \\
		Vote3Deep~\cite{vote3deep} & LiDAR & Voxel & 2017 & 1100  & - & - & - & - & - & - & - \\
		3D-FCN~\cite{3dfcn} & LiDAR & Voxel & 2017 & - & - & - & - & - & - & - & - \\
		VoxelNet~\cite{voxelnet18} & LiDAR & Voxel & 2018 & 220 & 77.47 & 65.11 & 57.73 & - & - & - & - \\
		SECOND~\cite{second} & LiDAR & Voxel & 2018 & 50 & 83.13 & 73.66 & 66.20 & - & - & - & - \\
		CBGS~\cite{cbgs} & LiDAR & Voxel & 2019 & - & - & - & - & 52.8 & 63.3 & - & - \\
		HVNet~\cite{hvnet} & LiDAR & Voxel & 2020 & 30 & - & - & - & - & - & - & - \\
		DOPS~\cite{dops} & LiDAR & Voxel & 2020 & - & - & - & - & - & - & 56.4 & - \\
		MVF~\cite{mvfwaymo} & LiDAR & Voxel & 2020 & - & - & - & - & - & - & 62.93 & - \\
		AFDet~\cite{afdet} & LiDAR & Voxel & 2020 & - & - & - & - & - & - & 63.69 & - \\
		SSN~\cite{ssn} & LiDAR & Voxel & 2020 & - & - & - & - & 46.3 & 56.9 & - & - \\
		CVC-Net~\cite{everyviewcounts} & LiDAR & Voxel & 2020 & - & - & - & - & 55.8 & 64.2 & 65.2 & - \\
		Wang \textit{et al.}~\cite{reconfigurablevoxels} & LiDAR & Voxel & 2020 & 50 & - & - & - & 48.5 & 59.0 & - & - \\
		SegVoxelNet~\cite{segvoxelnet} & LiDAR & Voxel & 2020 & 40 & 84.19 & 75.81 & 67.80 & - & - & - & - \\
		HotSpotNet~\cite{hotspotnet} & LiDAR & Voxel & 2020 & 40 & 87.60 & 78.31 & 73.34 & 59.3 & 66.0 & - & - \\
		Associate-3Ddet~\cite{associate3ddet} & LiDAR & Voxel & 2020 & 60 & 85.99 & 77.40 & 70.53 & - & - & - & - \\
		TANet~\cite{tanet} & LiDAR & Voxel & 2020 & - & 83.81 & 75.38 & 67.66 & - & - & - & - \\
		Part-A$^2$ Net~\cite{parta2} & LiDAR & Voxel & 2020 & 80 & 85.94 & 77.86 & 72.00 & - & - & - & - \\
		CenterPoint~\cite{centerpoint} & LiDAR & Voxel & 2021 & 70 & - & - & - & 58.0 & 65.5 & 76.7 & 68.8 \\
		Object DGCNN~\cite{objectdgcnn} & LiDAR & Voxel & 2021 & - & - & - & - & 58.7 & 66.1 & - & -\\
		CIA-SSD~\cite{ciassd} & LiDAR & Voxel & 2021 & 30 & 89.59 & 80.28 & 72.87 & - & - & - & - \\
		Voxel R-CNN~\cite{voxelrcnn} & LiDAR & Voxel & 2021 & 40 & 90.90 & 81.62 & 77.06 & - & - & 75.59 & 66.59 \\
		Voxel Transformer~\cite{votr} & LiDAR & Voxel & 2021 & 140 & 89.90 & 82.09 & 79.14 & - & - & 74.95 & 65.91 \\
  	SST~\cite{sst} & LiDAR & Voxel & 2022 & - & - & - & - & - & - & 74.2 &  65.5 \\
    	SWFormer~\cite{swformer} & LiDAR & Voxel & 2022 & - & - & - & - & - & - & 77.8 & 69.2 \\
		\hline
		PointPillars~\cite{pointpillars} & LiDAR & Pillar & 2019 & 16 & 79.05 & 74.99 & 68.30 & 40.1 & 55.0 & 56.62 & - \\
		Pillar-OD~\cite{pillar-od} & LiDAR & Pillar & 2020 & - & - & - & - & - & - & 69.8 & - \\
		PillarNet~\cite{pillarnet} & LiDAR & Pillar & 2022 & - & - & - & - & 66.0 & 71.4 & 83.23 & 76.09 \\
		\hline
		VeloFCN~\cite{velofcn} & LiDAR & BEV Image & 2016 & 1000 & - & - & - & - & - & -& - \\
		BirdNet~\cite{birdnet} & LiDAR & BEV Image & 2018 & - & 75.52$^*$ & 50.81$^*$ & 50.00$^*$ & - & - & - & - \\
		PIXOR~\cite{pixor} & LiDAR & BEV Image & 2018 & 35 & 81.70$^*$ & 77.05$^*$ & 72.95$^*$ & - & - & - & - \\
		HDNet~\cite{hdnet} & LiDAR & BEV Image & 2018 & 50 & 89.14$^*$ & 86.57$^*$ & 78.32$^*$ & - & - & - & - \\
		\hline
		PVCNN~\cite{pointvoxelcnn} & LiDAR & Point-Voxel & 2019 & 59 & - & - & - & - & - & - & - \\
		Fast Point R-CNN~\cite{fastpointrcnn} & LiDAR & Point-Voxel & 2019 & 65 & 84.28 & 75.73 & 67.39 & - & - & - & -\\
		PV-RCNN~\cite{pv-rcnn} & LiDAR & Point-Voxel & 2020 & - & 90.25 & 81.43 & 76.82 & - & - & 77.51 & 68.98 \\
		SA-SSD~\cite{sassd} & LiDAR & Point-Voxel & 2020 & 40 & 88.75 & 79.79 & 74.16 & - & - & - & - \\
		SPVNAS~\cite{spvnas} & LiDAR & Point-Voxel & 2020 & - & 87.8 & 78.4 & 74.8 & - & - & - & - \\
		InfoFocus~\cite{infofocus} & LiDAR & Point-Voxel & 2020 & - & - & - & - & 39.5 & - & - & - \\
		PVGNet~\cite{pvgnet} & LiDAR & Point-Voxel & 2021 & - & 89.94 & 81.81 & 77.09 & - & - & 74.0 & - \\
		HVPR~\cite{hvpr} & LiDAR & Point-Voxel & 2021 & 28 & 86.38 & 77.92 & 73.04 & - & - & - & - \\
		PV-RCNN++~\cite{pv-rcnn++} & LiDAR & Point-Voxel & 2021 & - & 90.14 & 81.88 & 77.15 & - & - & 79.25 & 70.61 \\
		CT3D~\cite{ct3d} & LiDAR & Point-Voxel & 2021 & - & 87.83 & 81.77 & 77.16 & - & - & 76.30 & 69.04  \\
		LiDAR R-CNN~\cite{lidar-rcnn} & LiDAR & Point-Voxel & 2021 & - & - & - & - & - & - & 76.0 & 68.3 \\
		Pyramid R-CNN~\cite{pyramid-rcnn} & LiDAR & Point-Voxel & 2021 & - & 88.39 & 82.08 & 77.49 & - & - & 76.30 & 67.23 \\
		\hline
		LaserNet~\cite{lasernet} & LiDAR & Range Image & 2019 & 30 & - & - & - & - & - & 52.11 & - \\
		RCD~\cite{rcd} & LiDAR & Range Image & 2020 & - & - & - & - & - & - & 69.59 & - \\
		RangeRCNN~\cite{rangercnn} & LiDAR & Range Image & 2020 & 45 & 88.47 & 81.33 & 77.09 & - & - & 75.43 & - \\
		RangeIoUDet~\cite{rangeioudet} & LiDAR & Range Image & 2021 & 22 & 88.60 & 79.80 & 76.76 & - & - & - & - \\
		PPC~\cite{tothepoint} & LiDAR & Range Image & 2021 & - & - & - & - & - & - & 65.2 & 56.7 \\
		RangeDet~\cite{rangedet} & LiDAR & Range Image & 2021 & - & - & - & - & - & - & 72.85 & - \\
		RSN~\cite{rsn} & LiDAR & Range Image & 2021 & 67.5 & - & - & - & - & - & 78.4 & 69.5 \\
		\hline
		Mono3D~\cite{mono3d} & Camera & Monocular & 2016 & - & - & - & - & - & - & - & - \\
		Deep3DBox~\cite{deep3dbox} & Camera & Monocular & 2017 & - & - & - & - & - & - & - & - \\
		Deep MANTA~\cite{deepmanta} & Camera & Monocular & 2017 & - & - & - & - & - & - & - & -\\
		SubCNN~\cite{subcnn} & Camera & Monocular & 2017 & - & - & - & - & - & - & - & - \\
		3D-RCNN~\cite{3drcnn} & Camera & Monocular & 2018 & - & - & - & - & - & - & - & - \\
		MultiFusion~\cite{multifusion} & Camera & Monocular & 2018 & - & 7.08 & 5.18 & 4.68 & - & - & - & - \\
		Mono3D++~\cite{mono3d++} & Camera & Monocular & 2019 & - & - & - & - & - & - & - & - \\
		DeepOptics~\cite{deep-optics} & Camera & Monocular & 2019 & - & - & - & - & - & - & - & - \\
		Weng \textit{et al.}~\cite{end-to-end-plidar} & Camera & Monocular & 2019 & - & - & - & - & - & - & - & - \\
		CenterNet~\cite{centernet} & Camera & Monocular & 2019 & - & - & - & - & 33.8 & 40.0 & - & - \\
		OFT-Net~\cite{oft-net} & Camera & Monocular & 2019 & 500 & 1.32 & 1.61 & 1.00 & - & - & - & - \\
		\hline
	\end{tabular}
	\label{tab:performance}
\end{table*}

\begin{table*}[ht]
	\caption{A comprehensive performance analysis of all branches of 3D object detection methods across different datasets. (Continued)}
	\centering
	\begin{tabular}{l|c|c|c|c|c|c|c|c|c|c|c}
		\hline
		\multirow{2}*{Method} & \multirow{2}*{Sensor} & \multirow{2}*{Representation} & \multirow{2}*{Year} &  \multirow{2}*{\tabincell{c}{Inference \\ Time (ms)}} & \multicolumn{3}{c|}{KITTI Car} & \multicolumn{2}{c|}{nuScenes} & \multicolumn{2}{c}{Waymo Vehicle} \\
		\cline{6-12}
		& & & & & Easy & Mod. & Hard & mAP & NDS & \tabincell{c}{L1} & \tabincell{c}{L2} \\
		\hline
		\hline
  	FQNet~\cite{FQNet} & Camera & Monocular & 2019 & 500 & 2.77 & 1.51 & 1.01 & - & - & - & - \\
		ROI-10D~\cite{ROI-10D} & Camera & Monocular & 2019 & 200 & 4.32 & 2.02 & 1.46 & - & - & - & - \\
		GS3D~\cite{GS3D} & Camera & Monocular & 2019 & - & 4.47 & 2.90 & 2.47 & - & - & - & - \\
		MonoFENet~\cite{MonoFENet} & Camera & Monocular & 2019 & - & 8.35 & 5.14 & 4.10 & - & - & - & - \\
		MonoGRNet~\cite{monogrnet} & Camera & Monocular & 2019 & 60 & 9.61 & 5.74 & 4.25 & - & - & - & - \\
		MonoDIS~\cite{monodis} & Camera & Monocular & 2019 & 100 & 10.37 & 7.94 & 6.40 & 30.4 & 38.4 & - & - \\
		MonoPSR~\cite{MonoPSR} & Camera & Monocular & 2019 & - & 10.76 & 7.25 & 5.85 & - & - & - & - \\
		M3D-RPN~\cite{m3d-rpn} & Camera & Monocular & 2019 & 160 & 14.76 & 9.71 & 7.42 & - & - & - & - \\
		AM3D~\cite{color-pslidar} & Camera & Monocular & 2019 & 400 & 16.50 & 10.74 & 9.52 & - & - & - & - \\
		SDFLabel~\cite{SDFLabel} & Camera & Monocular & 2020 & - & - & - & - & - & - & - & - \\
		MonoDR~\cite{MonoDR} & Camera & Monocular & 2020 & - & - & - & - & - & - & - & - \\
		Wang \textit{et al.}~\cite{ForeSeE} & Camera & Monocular & 2020 & - & - & - & - & - & - & - & - \\
		MoNet3D~\cite{MoNet3D} & Camera & Monocular & 2020 & - & - & - & - & - & - & - & - \\
		Cai \textit{et al.}~\cite{cai-mono} & Camera & Monocular & 2020 & - & 11.08 & 7.02 & 5.63 & - & - & - & - \\
		MonoPair~\cite{MonoPair} & Camera & Monocular & 2020 & 60 & 13.04 & 9.99 & 8.65 & - & - & - & - \\
		SMOKE~\cite{SMOKE} & Camera & Monocular & 2020 & 30 & 14.03 & 9.76 & 7.84 & - & - & - & - \\
		RTM3D~\cite{RTM3D} & Camera & Monocular & 2020 & - & 14.41 & 10.34 & 8.77 & - & - & - & - \\
		MoVi-3D ~\cite{MoVi-3D} & Camera & Monocular & 2020 & - & 15.19 & 10.90 & 9.26 & - & - & - & - \\
		UR3D~\cite{UR3D} & Camera & Monocular & 2020 & - & 15.58 & 8.61 & 6.00 & - & - & - & - \\
		D$^4$LCN~\cite{D4LCN} & Camera & Monocular & 2020 & 200 & 16.65 & 11.72 & 9.51 & - & - & - & - \\
		Ye \textit{et al.}~\cite{da-3ddet} & Camera & Monocular & 2020 & 400 & 16.77 & 12.72 & 9.17 & - & - & - & - \\
		Kinematic3D~\cite{Kinematic3D} & Camera & Monocular & 2020 & 120 & 19.07 & 12.72 & 9.17 & - & - & - & - \\
		CaDDN~\cite{CaDDN} & Camera & Monocular & 2021 & - & 19.17 & 13.41 & 11.46 & - & - & - & - \\
		PatchNet~\cite{PatchNet} & Camera & Monocular & 2021 & 400 & 15.68 & 11.12 & 10.17 & - & - & 0.39 & 0.38 \\
		MonoDLE~\cite{MonoDLE} & Camera & Monocular & 2021 & - & 17.23 & 12.26 & 10.29 & - & - & - & - \\
		M3DSSD~\cite{M3DSSD} & Camera & Monocular & 2021 & - & 17.51 & 11.46 & 8.98 & - & - & - & - \\
		Kumar \textit{et al.}~\cite{Mono-NMS} & Camera & Monocular & 2021 & - & 18.10 & 12.32 & 9.65 & - & - & - & - \\
		MonoRCNN~\cite{MonoRCNN} & Camera & Monocular & 2021 & 70 & 18.36 & 12.65 & 10.03 & - & - & - & - \\
		MonoRUn~\cite{MonoRUn} & Camera & Monocular & 2021 & - & 19.65 & 12.30 & 10.58 & - & - & - & - \\
		DDMP~\cite{DDMP} & Camera & Monocular & 2021 & - & 19.71 & 12.78 & 9.80 & - & - & - & - \\
		MonoFlex~\cite{MonoFlex} & Camera & Monocular & 2021 & - & 19.94 & 13.89 & 12.07 & - & - & - & - \\
		GUP Net~\cite{GUPNet} & Camera & Monocular & 2021 & - & 20.11 & 14.20 & 11.77 & - & - & - & - \\
		PCT~\cite{PCT} & Camera & Monocular & 2021 & - & 21.00 & 13.37 & 11.31 & - & - & 0.89 & 0.66 \\
		MonoEF~\cite{exfree-mono} & Camera & Monocular & 2021 & - & 21.29 & 13.87 & 11.71 & - & - & - & - \\
		Liu \textit{et al.}~\cite{liu-mono} & Camera & Monocular & 2021 & - & 21.65 & 13.25 & 9.91 & - & - & - & -\\
		DD3D~\cite{DD3D} & Camera & Monocular & 2021 & - & 23.22 & 16.34 & 14.20 & 41.8 & 47.7 & - & - \\
		FCOS3D~\cite{FCOS3D} & Camera & Monocular & 2021 & - & - & - & - & 35.8 & 42.8 & - & - \\
		PGD~\cite{PGD} & Camera & Monocular & 2022 & 28 & - & - & - & 38.6 & 44.8 & - & - \\
		MonoDTR~\cite{monodtr} & Camera & Monocular & 2022 & - & 21.99 & 15.39 & 12.73 & - & - & - & - \\
		\hline
		3DOP~\cite{3dop-conf} & Camera & Stereo & 2015 & - & - & - & - & - & - & - & - \\
		TLNet~\cite{TL-Net} & Camera & Stereo & 2019 & - & 7.64 & 4.37 & 3.74 & - & - & - & - \\
		Stereo R-CNN~\cite{stereo-rcnn} & Camera & Stereo & 2019 & 420 & 47.58 & 30.23 & 23.72 & - & - & - & - \\
		Pseudo-LiDAR~\cite{pseudolidar19} & Camera & Stereo & 2019 & - & 54.53 & 34.05 & 28.25 & - & - & - & - \\
		IDA-3D~\cite{IDA-3D} & Camera & Stereo & 2020 & - & - & - & - & - & - & - & - \\
		OC-Stereo~\cite{oc-stereo} & Camera & Stereo & 2020 & 350 & 55.15 & 37.60 & 30.25 & - & - & - & - \\
		ZoomNet~\cite{zoomnet} & Camera & Stereo & 2020 & 300 & 55.98 & 38.64 & 30.97 & - & - & - & - \\
		Disp R-CNN~\cite{disp-rcnn} & Camera & Stereo & 2020 & 420 & 58.53 & 37.91 & 31.93 & - & - & - & - \\
		P-LiDAR++~\cite{Pseudo-LiDAR++} & Camera & Stereo & 2020 & 500 & 61.11 & 42.43 & 36.99 & - & - & - & - \\
		Qian \textit{et al.}~\cite{end-to-end-pseudo-lidar} & Camera & Stereo & 2020 & 400 & 64.8 & 43.9 & 38.1 & - & - & - & - \\
		DSGN~\cite{DSGN} & Camera & Stereo & 2020 & 670 & 73.50 & 52.18 & 45.14 & - & - & - & - \\
		CG-Stereo~\cite{CG-Stereo} & Camera & Stereo & 2020 & - & 74.39 & 53.58 & 46.50 & - & - & - & - \\
		CDN~\cite{wasserstein-stereo} & Camera & Stereo & 2020 & 600 & 74.52 & 54.22 & 46.36 & - & - & - & - \\
		PLUMENet~\cite{PLUMENet} & Camera & Stereo & 2021 & 150 & 82.97$^*$ & 66.27$^*$ & 56.70$^*$ & - & - & - & - \\
		LIGA-Stereo~\cite{LIGA-Stereo} & Camera & Stereo & 2021 & 350 & 81.39 & 64.66 & 57.22 & - & - & - & - \\
		\hline
		Rubino \textit{et al.}~\cite{mcam3d-rubino} & Camera & Multi-View & 2017 & - & - & - & - & - & - & - & - \\    
		ImVoxelNet~\cite{imvoxelnet} & Camera & Multi-View & 2021 & 400 & 17.15 & 10.97 & 9.15 & - & - & - & - \\
		DETR3D~\cite{detr3d} & Camera & Multi-View & 2021 & - & - & - & - & 41.2 & 47.9 & - & - \\
        PETR~\cite{petr} & Camera & Multi-View & 2022 & - & - & - & - & 44.5 & 50.4 & - & - \\
        BEVDet~\cite{bevdet} & Camera & Multi-View & 2022 & - & - & - & - & 42.2 & 52.9 & - & - \\
        BEVerse~\cite{beverse} & Camera & Multi-View & 2022 & - & - & - & - & 39.3 & 53.1 & - & - \\
        BEVFormer~\cite{bevformer} & Camera & Multi-View & 2022 & - & - & - & - & 48.1 & 56.9 & - & - \\
        BEVDepth~\cite{bevdepth} & Camera & Multi-View & 2022 & - & - & - & - & 52.0 & 60.9 & - & - \\
        BEVStereo~\cite{bevstereo} & Camera & Multi-View & 2022 & - & - & - & - & 52.5 & 61.0 & - & - \\
        SOLOFusion~\cite{solofusion} & Camera & Multi-View & 2022 & - & - & - & - & 54.0 & 61.9 & - & - \\
        \hline
		F-PointNet~\cite{fpointnet18} & Fusion & Early & 2018 & - & 81.20 & 70.39 & 62.19 & - & - & - & - \\
		RoarNet~\cite{roarnet} & Fusion & Early & 2019 & 100 & 83.95 & 75.79 & 67.88 & - & - & - & - \\
		F-ConvNet~\cite{F-ConvNet} & Fusion & Early & 2019 & - & 85.88 & 76.51 & 68.08 & - & - & - & - \\
		PointPainting~\cite{pointpainting} & Fusion & Early & 2020 & - & 82.11 & 71.70 & 67.08 & 46.4 & 58.1 & - & - \\
		FusionPainting~\cite{fusionpainting} & Fusion & Early & 2021 & - & - & - & - & 66.5 & 70.7 & - & - \\
		PointAugmenting~\cite{pointaugmenting} & Fusion & Early & 2021 & 542 & - & - & - & 66.8 & 71.0 & 67.41 & 62.70 \\
		MVP~\cite{mvp} & Fusion & Early & 2021 & - & - & - & - & 66.4 & 70.5 & - & -  \\
		\hline
		MV3D~\cite{mv3d} & Fusion & Intermediate & 2017 & 240 & 71.09 & 62.35 & 55.12 & - & - & - & - \\
		PointFusion~\cite{pointfusion} & Fusion & Intermediate & 2018 & - & - & - & - & - & - & - & - \\
		AVOD~\cite{avod} & Fusion & Intermediate & 2018 & 100 & 81.94 & 71.88 & 66.38 & - & - & - & - \\
		\hline
	\end{tabular}
	\label{tab:performance_contd}
\end{table*}

\begin{table*}[ht]
	\caption{A comprehensive performance analysis of all branches of 3D object detection methods across different datasets. (Continued)}
	\centering
	\begin{tabular}{l|c|c|c|c|c|c|c|c|c|c|c}
		\hline
		\multirow{2}*{Method} & \multirow{2}*{Sensor} & \multirow{2}*{Representation} & \multirow{2}*{Year} &  \multirow{2}*{\tabincell{c}{Inference \\ Time (ms)}} & \multicolumn{3}{c|}{KITTI Car} & \multicolumn{2}{c|}{nuScenes} & \multicolumn{2}{c}{Waymo Vehicle} \\
		\cline{6-12}
		& & & & & Easy & Mod. & Hard & mAP & NDS & \tabincell{c}{L1} & \tabincell{c}{L2} \\
		\hline
		\hline
		ContFuse~\cite{contfuse} & Fusion & Intermediate & 2018 & 60 & 82.54 & 66.22 & 64.04 & - & - & - & - \\
  	MVX-Net~\cite{mvx-net} & Fusion & Intermediate & 2019 & - & 83.2 & 72.7 & 65.2 & - & - & - & - \\
		MMF~\cite{mmf} & Fusion & Intermediate & 2019 & 80 & 86.81 & 76.75 & 68.41 & - & - & - & - \\
		3D-CVF~\cite{3d-cvf} & Fusion & Intermediate & 2020 & 75 & 89.20 & 80.05 & 73.11 & 52.7 & 62.3 & - & - \\
		EPNet~\cite{epnet} & Fusion & Intermediate & 2020 & - & 89.81 & 79.28 & 74.59 & - & - & - & - \\
		TransFusion~\cite{transfusion} & Fusion & Intermediate & 2022 & - & - & - & - & 68.9 & 71.7 & - & - \\
		BEVFusion~\cite{bevfusion2} & Fusion & Intermediate & 2022 & - & - & - & - & 69.2 & 71.8 & - & - \\
		UVTR~\cite{uvpr} & Fusion & Intermediate & 2022 & - & - & - & - & 67.1 & 71.1 & - & - \\
		\hline
		CLOCs~\cite{clocs} & Fusion & Late & 2020 & 150 & 88.94 & 80.67 & 77.15 & - & - & - & - \\
		Fast-CLOCs~\cite{fast-clocs} & Fusion & Late & 2022 & 125  & 89.11 & 80.34 & 76.98 & 63.1 & 68.7 & - & - \\
		\hline
	\end{tabular}
	\label{tab:performance_contd_contd}
\end{table*}

\subsubsection{Trends of inference time}

PointPillars~\cite{pointpillars} has achieved remarkable inference speed with only $16$ms latency, and its architecture has been adopted by many following works~\cite{centerpoint, pillar-od, pillarnet}. However, even with the emergence of more powerful hardware, the inference speed didn't exhibit a significant improvement over the years. This is mainly because most methods focus on performance improvement and pay less attention to efficient inference. Many papers have introduced new modules into the existing detection pipelines, which also brings additional time costs. For the pseudo-LiDAR based detection methods, the stereo-based methods, and most multi-modal methods, the inference time is generally more than $100$ ms, which cannot satisfy the real-time requirement and hampers the deployment in real-world applications.

\subsubsection{Trends of the LiDAR-based methods}

LiDAR-based 3D object detection has witnessed great advances in recent years. Among the LiDAR-based methods, the voxel-based and point-voxel based detection approaches attain superior performances, \textit{e.g.} \cite{votr} attains $82.09\%$ moderate $AP_{3D}$ and \cite{pv-rcnn} obtains $90.25\%$ easy $AP_{3D}$ on the KITTI dataset. The pillar-based detection methods are extremely fast, \textit{e.g.} \cite{pointpillars} runs at $60$ Hz, but the detection accuracy is generally worse than the voxel-based methods. The range-based and BEV-based approaches are also quite efficient, \textit{e.g.} \cite{pixor} and \cite{lasernet} only requires $30$ ms for one-pass inference. The point-based detectors can obtain a good performance, but their inference speeds are greatly influenced by the choices of sampling and operators.

For point-based 3D object detectors, moderate AP has been increasing from $53.46\%$~\cite{pointrcnn19} to $79.57\%$~\cite{pointgnn20} on the KITTI benchmark. The performance improvements are mainly owing to two factors: more robust point cloud samplers and more powerful point cloud operators. The development of point cloud samplers starts with Farthest Point Sampling (FPS)~\cite{pointrcnn19, std}, and many following point cloud detectors have been improving point cloud samplers based on FPS, including fusion-based FPS~\cite{3dssd}, target-based FPS~\cite{starnet}, FPS with coordinates refinement~\cite{pointformer}. A good point cloud sampler could produce candidate points that have better coverage of the whole scene, so it avoids missing detections when the point cloud is sparse, which helps improve the detection performance. Besides point cloud samplers, point cloud operators have also progressed rapidly, from the standard set abstraction~\cite{pointrcnn19, ipod, std, 3dssd} to graph operators~\cite{pointgnn20, starnet} and Transformers~\cite{pointformer}. Point cloud operators are crucial for extracting powerful feature representations from point clouds. Hence powerful point cloud operators can help detectors better obtain semantic information about 3D objects and improve performance.

For grid-based 3D object detectors, moderate AP has been increasing from $50.81\%$~\cite{birdnet} to $82.09\%$~\cite{votr} on the KITTI benchmark. The performance improvements are mainly driven by better backbone networks and detection heads. The development of backbone networks has experienced four stages: (1) 2D networks to process BEV images that are generated by point cloud projection~\cite{birdnet, pixor}, (2) 2D networks to process pillars that are generated by PointNet encoding~\cite{pointpillars}, (3) 3D sparse convolutional networks to process voxelized point clouds~\cite{voxelnet18}, (4) Transformer-based architectures~\cite{votr, sst, swformer}. The trend of backbone designs is to encode more 3D information from point clouds, which leads to more powerful BEV representations and better detection performance, but those early designs are still popular due to efficiency. Detection head designs have experienced the transition from anchor-based heads~\cite{second} to center-based heads~\cite{centerpoint}, and the object localization ability has been improved with the development of detection heads. Other head designs such as IoU rectification~\cite{ciassd} and sequential head~\cite{point2seq} can further boost performance.

For point-voxel based 3D object detectors, moderate AP has been increasing from $75.73\%$~\cite{fastpointrcnn} to $82.08\%$~\cite{pyramid-rcnn} on the KITTI benchmark. The performance improvements come from more power operators~\cite{pointvoxelcnn, sassd} and modules~\cite{pv-rcnn, pv-rcnn++, pyramid-rcnn, ct3d} that can effectively fuse point and voxel features. 

For range-based 3D object detectors, L1 mAP has been increasing from $52.11\%$~\cite{lasernet} to $78.4\%$~\cite{rsn} on the Waymo Open dataset. The performance improvements come from designs of specialized operators~\cite{rcd, rangedet, tothepoint} that can handle range images more effectively, as well as view transforms and multi-view aggregation~\cite{rangercnn, rangeioudet, rsn}.

\subsubsection{Trends of the camera-based methods}

Camera-based 3D object detection has shown rapid progress recently. Among the camera-based methods, the stereo-based detection methods generally outperform the monocular detection approaches by a large margin. For example, the state-of-the-art stereo-based method~\cite{LIGA-Stereo} attains $64.66\%$ moderate $AP_{3D}$, while the state-of-the-art monocular method~\cite{DD3D} only achieves $16.34\%$ moderate $AP_{3D}$. This is mainly because depth and disparity estimated from stereo images are much more accurate than those estimated from monocular images, and accurate depth estimation is the most important factor in camera-based 3D object detection. Multi-camera 3D object detection has been progressing fast with the emergence of BEV perception and Transformers. State-of-the-art method~\cite{solofusion} attains $54.0\%$ mAP and $61.9$ NDS on nuScenes, which has outperformed some prestigious LiDAR-based 3D object detectors~\cite{pointpillars}.

For monocular 3D object detectors, moderate AP has been increasing from $1.51\%$~\cite{FQNet} to $16.34\%$~\cite{DD3D} on the KITTI benchmark. The major challenge of monocular 3D object detection is how to obtain accurate 3D information from a single 2D image, as localization errors dominate detection errors. The performance improvements are driven by more accurate depth prediction, which can be achieved by better network architecture designs~\cite{m3d-rpn, FCOS3D, centernet, ROI-10D}, leveraging depth images~\cite{multifusion} or pseudo-LiDAR point clouds~\cite{pseudolidar19, Pseudo-LiDAR++}, introducing geometry constraints~\cite{deep3dbox, cai-mono, mono3d, MonoPair}, and 3D object reconstruction~\cite{MonoRUn, SDFLabel, 3dvp, 3drcnn}.

For stereo-based 3D object detectors, moderate AP has been increasing from $4.37\%$~\cite{TL-Net} to $64.66\%$~\cite{LIGA-Stereo} on the KITTI benchmark. The performance improvements mainly come from better network designs and data representations. Early works~\cite{stereo-rcnn} rely on stereo-based 2D detection networks to produce paired object bounding boxes and then predict object-centric stereo/depth information with a sub-network. However, those object-centric methods generally lack global disparity information which hampers accurate 3D detection in a scene. Later on, pseudo-LiDAR based approaches~\cite{pseudolidar19} generate disparity maps from stereo images and then transform disparity maps into 3D pseudo-LiDAR point clouds that are finally passed to a LiDAR detector to perform 3D detection. The transformation from 2D disparity maps to 3D point clouds is crucial and can significantly boost 3D detection performance. Many following papers are based on the pseudo-LiDAR paradigm and improve it with stronger stereo matching network~\cite{Pseudo-LiDAR++} and end-to-end training of stereo matching and LiDAR detection~\cite{end-to-end-pseudo-lidar}. Recent methods~\cite{PLUMENet, DSGN} transforms disparity maps into 3D volumes and apply grid-based detectors on the volumes, which results in better performance. 

For multi-view 3D object detection, mAP has been increasing from $41.2\%$~\cite{detr3d} to $54.0\%$~\cite{solofusion} on the nuScenes dataset. For BEV-based approaches, the performance improvements are mainly from better depth prediction~\cite{bevdepth, bevstereo}. More accurate depth information results in more accurate camera-to-BEV transformation so detection performance can be improved. For query-based methods, the performance improvements come from better designs of 3D object queries~\cite{bevformer}, more powerful image features~\cite{petr}, and new attention mechanisms~\cite{spatialdetr}. 

\subsubsection{Trends of the multi-modal methods}

The multi-modal methods generally exhibit a performance improvement over the single-modal baselines but at the cost of introducing additional inference time. For instance, the multi-modal detector~\cite{pointaugmenting} outperforms the LiDAR baseline~\cite{centerpoint} by $8.8\%$ mAP on nuScenes, but the inference time of~\cite{pointaugmenting} also increases to $542$ ms compared to the baseline $70$ ms. The problem can be more severe in the early-fusion based approaches, where the 2D networks and the 3D detection networks are connected in a sequential manner. Most multi-modal detection methods are designed and tested on the KITTI dataset, in which only a front-view image and the corresponding point cloud are utilized. Recently more and more methods are proposed and evaluated on the nuScenes dataset, in which multi-view images, point clouds, and high-definition maps are provided.

For early-fusion based methods, moderate AP increases from $70.39\%$~\cite{fpointnet18} to $76.51\%$~\cite{F-ConvNet} on the KITTI benchmark, and mAP increases from $46.4\%$~\cite{pointpainting} to $66.8\%$~\cite{pointaugmenting} on nuScenes dataset. There are two crucial factors that contribute to the performance increase: knowledge fusion and data augmentation. From the results, we can observe that point-level knowledge fusion~\cite{pointpainting, fusionpainting} is generally more effective than region-level fusion~\cite{fpointnet18, F-ConvNet}. This is because region-level knowledge fusion simply reduces the detection range, while point-level knowledge fusion can provide fine-grained semantic information which is more beneficial in 3D detection. Besides, consistent data augmentations between point clouds and images~\cite{pointaugmenting} can also significantly boost detection performance. 

For intermediate and late fusion based methods, moderate AP increases from $62.35\%$~\cite{mv3d} to $80.67\%$~\cite{clocs} on the KITTI benchmark, and mAP increases from $52.7\%$~\cite{3d-cvf} to $69.2\%$~\cite{bevfusion} on the nuScenes dataset. Most methods focus on three critical problems: where to fuse different data representations, how to fuse these representations, and how to build reliable alignments between points and image pixels. For the where-to-fuse problem, different approaches try to fuse image and LiDAR features at different places, \textit{e.g.} 3D backbone networks, BEV feature maps, RoI heads, and outputs. From the results we can observe that fusion at any place can boost detection performance over single-modality baselines, and fusion in the BEV space~\cite{bevfusion, bevfusion2, transfusion} is more popular recently for its performance and efficiency. For the how-to-fuse problem, the development of fusion operators has experienced simple concatenation~\cite{avod}, continuous convolutions~\cite{mmf, contfuse}, attention~\cite{3d-cvf, epnet}, and Transformers~\cite{transfusion, uvpr, futr3d}, and fusion with Transformers exhibit prominent performance on all benchmarks. For the point-to-pixel alignment problem, most papers reply on fixed extrinsics and intrinsics to construct point-to-pixel correspondences. However, due to occlusion and calibration errors, those correspondences can be noisy and misalignment will harm performance. Recent works~\cite{bevfusion} circumvent this problem by directly fusing camera and LiDAR BEV feature maps, which is more robust to noise.  

\subsubsection{Systematic comparisons}

Considering all the input sensors and modalities, LiDAR-based detection is the best solution to the 3D object detection problem, in terms of both speed and accuracy. For instance, \cite{centerpoint} achieves $80.28\%$ moderate $AP_{3D}$ and still runs at $30$ FPS on KITTI. Multi-modal detection is built upon LiDAR-based detection, and can obtain a better detection performance compared to the LiDAR baselines, becoming state-of-the-art in terms of accuracy. Camera-based 3D object detection is a much cheaper and quite efficient solution in contrast to LiDAR and multi-modal detection. Nevertheless, the camera-based methods generally have a worse detection performance due to inaccurate depth predictions from images. The state-of-the-art monocular~\cite{DD3D} and stereo~\cite{LIGA-Stereo} detection approach only obtain $16.34\%$ and $64.66\%$ moderate $AP_{3D}$ respectively on KITTI. Recent advances in multi-view 3D object detection are quite promising. The state-of-the-art~\cite{solofusion} achieves $54.0\%$ mAP on nuScenes, which could perform on par with some classic LiDAR detectors~\cite{second}. In conclusion, LiDAR-based and multi-modal detectors are the best solutions considering speed and accuracy as the dominant factors, while camera-based detectors can be the best choice considering cost as the most important factor, and multi-view 3D detectors are becoming promising and may outperform LiDAR detectors in the future.

\subsection{Future outlooks} \label{sec:future}

With all the reviewed literature and the analysis of research trends over the past years, we can now make some predictions on the future research directions of 3D object detection.

\subsubsection{Open-set 3D object detection}

Nearly all existing works are proposed and evaluated on close datasets, in which the data only covers limited driving scenarios and the annotations only include basic classes, \textit{e.g.} cars, pedestrians, cyclists. Although those datasets can be large and diverse, they are still not sufficient for real-world applications, in which critical scenarios like traffic accidents and rare classes like unknown obstacles are important but not covered by the existing datasets. Therefore, existing 3D object detectors that are trained on the close sets have a limited capacity of dealing with those critical scenarios and cannot identify the unknown categories. To overcome the above limitations, designing 3D object detectors that can learn from the open world and recognize a wide range of object categories will be a promising research direction. \cite{openset-3ddet} is a good start for open-set 3D object detection and hopefully more methods will be proposed to tackle this problem.

\subsubsection{Detection with stronger interpretability}

Deep learning based 3D object detection models generally lack interpretability. Namely, some important questions on how the networks can identify 3D objects in point clouds, how occlusion and noise of 3D objects can affect the model outputs, and how much context information is needed for detecting a 3D object, have not been properly answered due to the black-box property of deep neural networks. On the other hand, understanding the behaviors of 3D detectors and answering these questions are quite important if we want to perform 3D object detection in a more robust manner and avoid those unexpected cases brought by black-box detectors. Therefore, the methods that can understand and interpret the existing 3D object detection models will be appealing in future research.

\subsubsection{Efficient hardware design for 3D object detection}

Most existing works focus on designing algorithms to tackle the 3D object detection problem, and their models generally run on GPUs. Nevertheless, unlike image operators that are highly optimized for GPU devices, point clouds and voxels are sparse and irregular, and the commonly adopted 3D operators like set abstraction or 3D sparse convolutions are not well suited for GPUs. Hence those LiDAR object detectors cannot run as efficiently as the image detectors on the existing hardware devices. To handle this challenge, designing novel devices where the hardware architectures are optimized for 3D operators as well as the task of 3D object detection will be an important research direction and will be beneficial for real-world deployment. \cite{pointacc} is a pioneering hardware work to accelerate point cloud processing, and we believe more and more papers will come in this field. In addition, new sensors, \textit{e.g.} solid-state LiDARs, LiDARs with doppler, 4D radars, will also inspire the design of 3D object detectors. 

\subsubsection{Detection in end-to-end self-driving systems}

Most existing works treat 3D object detection as an independent task and try to maximize the detection metrics such as average precision. Nevertheless, 3D object detection is closely correlated with other perception tasks as well as downstream tasks such as prediction and planning, so simply pursuing high average precision for 3D object detection may not be optimal when considering the autonomous driving system as a whole. Therefore, conducting 3D object detection and other tasks in an end-to-end manner, and learning 3D detectors from the feedback of planners, will be the future research trends of 3D object detection.

\section{Conclusion}

In this paper, we comprehensively review and analyze various aspects of 3D object detection for autonomous driving. We start from the problem definition, datasets, and evaluation metrics for 3D object detection, and then we introduce various kinds of sensor-based 3D object detection approaches, including LiDAR-based, camera-based, and multi-modal 3D object detection methods. We further investigate 3D object detection leveraging temporal data, with label-efficient learning, as well as its applications in autonomous driving systems. Finally, we summarize the research trends in recent years and prospect the future research directions of 3D object detection.

\section{Acknowledgements}


This project is funded in part by the National Key R\&D Program of China Project 2022ZD0161100,  by the Centre for Perceptual and Interactive Intelligence (CPIl) Ltd under the Innovation and Technology Commission (ITC)'s InnoHK, by General Research Fund Project 14204021 and Research Impact Fund Project R5001-18 of Hong Kong RGC. Hongsheng Li and Xiaogang Wang are PIs of CPII under the InnoHK.
\clearpage
\bibliographystyle{spbasic}
\footnotesize
\bibliography{ref}

\end{document}